\pdfoutput=1
\documentclass[twoside,11pt]{article}

\usepackage{jmlr2e}

\usepackage{lastpage}
\jmlrheading{26}{2025}{1-\pageref{LastPage}}{1/23; Revised
12/24}{5/25}{23-0058}{Atticus Geiger, Duligur Ibeling, Amir Zur, Maheep Chaudhary, Sonakshi Chauhan, Jing Huang, Aryaman Arora, Zhengxuan Wu, Noah Goodman, Christopher Potts, Thomas Icard}
\ShortHeadings{Causal Abstraction for Mechanistic Interpretability}{Geiger, Ibeling, Zur, Chaudhary, Chauhan, Huang, Arora, Wu, Goodman, Potts, Icard}
\firstpageno{1}

\usepackage{tikz} %
\usepackage{mathtools} %
\usepackage{enumitem}
\usepackage{booktabs} %
\usepackage{color}
\usepackage{wasysym}
\usepackage{microtype}
\usepackage{enumitem}
\usepackage{amsfonts}
\usepackage{amsmath} 
\usepackage{listings}
\usepackage{xcolor}
\usepackage{qtree}
\usepackage{clrscode3e}
\usepackage{caption}
\usepackage{comment}
\usepackage{subcaption}
\usepackage{svg}
\usepackage{amssymb}
\usepackage{bbm}
\usetikzlibrary{shapes.geometric}
\usetikzlibrary{ arrows, calc,matrix,positioning}
\usetikzlibrary{positioning}
\usepackage{gensymb} %

\usepackage[linesnumbered,ruled,vlined]{algorithm2e}
\usepackage{subcaption}

\hypersetup{colorlinks, linkcolor=black, urlcolor=black, citecolor=black}

\begin{document}

\title{Causal Abstraction: \\A Theoretical Foundation for Mechanistic Interpretability}

\author{Atticus Geiger$^{* \diamondsuit}$,
      Duligur Ibeling$^{\spadesuit}$,
      Amir Zur$^{\diamondsuit}$,
      Maheep Chaudhary$^{\diamondsuit}$,\AND
      Sonakshi Chauhan$^{\diamondsuit}$,
      Jing Huang$^{\spadesuit}$,
      Aryaman Arora$^{\spadesuit}$, 
      Zhengxuan Wu$^{\spadesuit}$,\AND
      Noah Goodman$^{\spadesuit}$,
      Christopher Potts$^{\spadesuit}$,
      Thomas Icard$^{* \spadesuit}$\\ \AND
  $^{\diamondsuit}$Pr(Ai)$^{2}$R Group \;\; $^{\spadesuit}$Stanford University\\ \AND
  $^*$Corresponding authors: atticusg@gmail.com;  icard@stanford.edu
  }

\editor{Jin Tian}

\maketitle

\begin{abstract}%
Causal abstraction provides a theoretical foundation for mechanistic interpretability, the field concerned with providing intelligible algorithms that are faithful simplifications of the known, but opaque low-level details of black box AI models. Our contributions are (1) generalizing the theory of causal abstraction from mechanism replacement (i.e., hard and soft interventions) to arbitrary mechanism transformation (i.e., functionals from old mechanisms to new mechanisms), (2) providing a flexible, yet precise formalization for the core concepts of polysemantic neurons, the linear representation hypothesis, modular features, and graded faithfulness, and (3) unifying a variety of mechanistic interpretability methods in the common language of causal abstraction, namely, activation and path patching, causal mediation analysis, causal scrubbing, causal tracing, circuit analysis, concept erasure, sparse autoencoders, differential binary masking, distributed alignment search, and steering.
\end{abstract}

\newpage

\tableofcontents

\newcommand{\TODO}[1]{\textcolor{red}{#1}}

\newcommand{\allvars}{\mathbf{V}}
\newcommand{\allvarrs}{\mathbf{W}}
\newcommand{\inputvars}{\vars^{\mathbf{In}}}
\newcommand{\inputval}{\pset^{\mathbf{In}}}
\newcommand{\outputvars}{\vars^{\mathbf{Out}}}
\newcommand{\outputval}{\pset^{\mathbf{Out}}}

\newcommand{\tset}{\mathbf{v}}
\newcommand{\tsett}{\mathbf{w}}
\newcommand{\values}[1]{\mathsf{Val}_{#1}}
\newcommand{\signature}{\Sigma}

\newcommand{\onevar}{X}
\newcommand{\onevarr}{Y}
\newcommand{\onevarrr}{Z}
\newcommand{\onevarrrr}{W}

\newcommand{\oneval}{x}
\newcommand{\onevall}{y}
\newcommand{\onevalll}{z}
\newcommand{\onevallll}{w}

\newcommand{\vars}{\mathbf{X}}
\newcommand{\varrs}{\mathbf{Y}}
\newcommand{\varrrs}{\mathbf{Z}}
\newcommand{\varrrrs}{\mathbf{W}}

\newcommand{\pset}{\mathbf{x}}
\newcommand{\psett}{\mathbf{y}}
\newcommand{\psettt}{\mathbf{z}}
\newcommand{\psetttt}{\mathbf{w}}

\newcommand{\inneuron}{N}
\newcommand{\outneuron}{O}
\newcommand{\hiddenneuron}{H}
\newcommand{\hiddenneuronn}{G}

\newcommand{\inneuronval}{n}
\newcommand{\outneuronval}{o}
\newcommand{\hiddenneuronval}{h}
\newcommand{\hiddenneuronvall}{g}

\newcommand{\inneurons}{\mathbf{N}}
\newcommand{\outneurons}{\mathbf{O}}
\newcommand{\hiddenneurons}{\mathbf{H}}
\newcommand{\hiddenneuronss}{\mathbf{G}}
\newcommand{\hiddenneuronset}{\mathbf{h}}
\newcommand{\hiddenneuronsett}{\mathbf{g}}
\newcommand{\hiddenneuronintal}{\mathbb{H}}

\newcommand{\inneuronset}{\mathbf{n}}
\newcommand{\outneuronset}{\mathbf{o}}

\newcommand{\simfunc}{\mathsf{Sim}}
\newcommand{\statistic}{\mathbb{S}}

\newcommand{\base}{\mathbf{b}}
\newcommand{\source}{\mathbf{s}}

\newcommand{\rintinv}{\mathsf{RecIntInv}}
\newcommand{\rintinvset}{\Delta}
\newcommand{\dintinv}{\mathsf{DistIntInv}}
\newcommand{\intinv}{\mathsf{IntInv}}

\newcommand{\weights}{W}
\newcommand{\bias}{b}

\newcommand{\project}[2]{\mathsf{Proj}_{#2}(#1)}
\newcommand{\Project}[2]{\mathsf{Proj}_{#2}\bigg(#1\bigg)}
\newcommand{\inverseproject}[2]{\mathsf{Proj}^{-1}_{#2}(#1)}

\newcommand{\mechanisms}[2]{\{\mathcal{F}_{#1}\}_{#1 \in #2}}
\newcommand{\mechanismms}[2]{\{\mathcal{G}_{#1}\}_{#1 \in #2}}

\newcommand{\mechanism}[1]{\mathcal{F}_{#1}}
\newcommand{\mechanismm}[1]{\mathcal{G}_{#1}}

\newcommand{\model}{\mathcal{M}}
\newcommand{\M}{\mathcal{M}}
\newcommand{\sols}{\mathsf{Solve}}

\newcommand{\intfam}{\Delta}
\newcommand{\intpset}{\mathbf{i}}
\newcommand{\intpsett}{\mathbf{j}}
\newcommand{\intpsettt}{\mathbf{k}}
\newcommand{\intvars}{\mathbf{I}}
\newcommand{\intvarrs}{\mathbf{J}}

\newcommand{\softint}{\mathcal{I}}
\newcommand{\softintfull}[2]{\{\softint_{#1}\}_{#1 \in #2}}

\newcommand{\intal}{\mathbb{I}}
\newcommand{\intall}{\mathbb{J}}
\newcommand{\intalfull}[2]{\{\intal_{#1}\}_{#1 \in #2}}

\newcommand{\allints}{\mathsf{Hard}}
\newcommand{\allsoftints}{\mathsf{Soft}}
\newcommand{\allintals}{\mathsf{Func}}

\newcommand{\hardintals}{\Phi}
\newcommand{\intals}{\Psi}

\newcommand{\seq}{\alpha}
\newcommand{\seqq}{\beta}
\newcommand{\algprec}{\lessdot}

\newcommand{\setmap}{\tau}
\newcommand{\intmap}{\omega}

\newcommand{\cellpart}{\Pi}
\newcommand{\cellfunc}{\pi}

\newcommand{\types}{\mathcal{T}}
\newcommand{\type}{T}

\newcommand{\lowmodel}{\mathcal{L}}
\newcommand{\highmodel}{\mathcal{H}}

\newcommand{\marg}[2]{\epsilon_{#2}(#1)}
\newcommand{\valmerge}[2]{\Delta(#1)}
\newcommand{\varmerge}[2]{\cellpart(#1)}
\newcommand{\pred}{pred_{<}}

\newcommand{\distribution}{\mathbb{P}}

\newcommand{\dom}{\mathsf{Domain}}
\newcommand{\indicator}[1]{\mathbbm{1}[#1]}

\newcommand{\EndoVars}{\allvars}

\newcommand{\LowLevel}{\lowmodel}
\newcommand{\HighLevel}{\highmodel}

\newcommand{\EndoVarsLowLevel}{\allvars_{\LowLevel}}
\newcommand{\EndoVarsHighLevel}{\allvars_{\HighLevel}}

\begin{center}

\resizebox{\textwidth}{!}{
\renewcommand{\arraystretch}{1.4}
{\small
\begin{tabular}{l c r}
\toprule
     Symbol & Meaning & Location \\
     \midrule
      $\dom(f)$ &  the domain of a function $f$ & \\
      $\tset \mapsto f(\tset)$ &  the function $f$ & \\
      $\indicator{\varphi}$ & An indicator function that outputs $1$ if $\varphi$ is true, $0$ otherwise. & \\   
     $\allvars$ & A complete set of variables & Definition~\ref{def:signature}\\ 
     $\values{\vars}$ & A function mapping variables $\vars \subseteq \allvars$ to values & Definition~\ref{def:signature}\\ 
     $\signature$ & A signature (variables $\allvars$ and values $\values{}$) & Definition~\ref{def:signature}\\ 
     $\tset$ & A total setting $\tset\in \values{\allvars}$ & Definition~\ref{def:settings} \\
     $\onevar, \oneval$ & A variable $\onevar \in \allvars$ and a value $\oneval \in \values{\onevar}$ & Definition~\ref{def:settings} \\
     $\vars, \pset$ & A set of variables $\vars \subseteq \allvars$ and a partial setting $\pset\in \values{\vars}$  & Definition~\ref{def:settings} \\
     $\project{\psett}{\vars}$ & The restriction of a partial setting $\psett$ to variables $\vars \subseteq \varrs \subseteq \allvars$ &  Definition~\ref{def:projection}\\
     $\inverseproject{\pset}{\varrs}$ & The set of partial settings $\psett$ where $\project{\psett}{\vars} = \pset$ &  Definition~\ref{def:projection}\\
     $\mechanisms{\onevar}{\allvars}$ & Mechanisms for the variables in $\allvars$ & Definition~\ref{def:causalmodel}\\ 
     $\prec$ & An ordering of $\allvars$ by causal dependency as induced via $\mechanisms{\onevar}{\allvars}$ & Definition~\ref{def:causalmodel}\\ 
     $\model$ & A causal model (a signature $\signature$ and mechanisms $ \mechanisms{\onevar}{\allvars}$) & Definition~\ref{def:causalmodel}\\ 
     $\sols(\model)$ & The set of total settings that are solutions to a model $\mathcal{M}$ & Definition~\ref{def:causalmodel}\\ 
     $\intpset \in \allints$ & A hard intervention that fixes the variables $\intvars \subseteq \allvars$ & Definition~\ref{def:intervention}\\
     $\softintfull{\onevar}{\vars} \in \allsoftints$ & A soft intervention that replaces the mechanisms of $\vars\subseteq \allvars$  & Definition~\ref{def:softintervention}\\
     $\intalfull{\onevar}{\vars} \in \allintals$ & An interventional that edits the mechanisms of $\vars \subseteq \allvars$ & Definition~\ref{def:interventional}\\
     $\allintals_{\allvars}$ & The set of interventionals that edit all mechanisms & Definition~\ref{def:interventional}\\
     $\hardintals$  & The interventionals in $\allintals_{\allvars}$ equivalent to hard interventions.  & Definition~\ref{def:interventional}\\
     $\seq$  & A sequence of elements & Definition~\ref{def:interventional}\\
     $\intals$  & A set of interventionals & Definition~\ref{def:interventional}\\
     $\setmap$  & A map from total settings of $\model$ to total settings of $\model^*$ & Definition~\ref{def:exact}\\
     $\intmap$  & A map from interventionals in $\model$ to interventionals of $\model^*$ & Definition~\ref{def:exact}\\
     $\highmodel, \lowmodel$ & High-level and low-level causal models & Definition~\ref{def:cc}\\ 
     $\cellpart_{\onevar_{\highmodel}}$  & A partition cell of $\allvars_{\lowmodel}$ assigned to the variable $\onevar_{\highmodel} \in \allvars_{\highmodel}$ & Definition~\ref{def:alignment}\\
     $\cellfunc_{\onevar_{\highmodel}}$  & A function mapping from values of $\cellpart_{\onevar_{\highmodel}}$ to values of $\onevar_{\highmodel}$ & Definition~\ref{def:alignment}\\
      \bottomrule
\end{tabular}
}
}
\end{center}

\newpage

\newpage
\section{Introduction}

We take the fundamental aim of explainable artificial intelligence to be explaining \emph{why} a model makes the predictions it does. For many purposes the paradigm of explanation is causal explanation \citep{Woodward2003, pearl:2019}, elucidating counterfactual difference-making details of the mechanisms underlying model behavior. However, not just any causal explanation will be apt. There is an obvious sense in which we already know all the low-level causal facts about a deep learning model. After all, we can account for every aspect of model behavior in terms of real-valued vectors, activation functions, and weight tensors. The problem, of course, is that these low-level explanations are typically not \emph{transparent} to humans---they fail to instill an understanding of the high-level principles underlying model behavior \citep{Lipton18, Creel2020} that can guide human action \citep{Karimi2021, Karimi2023}.

In many contexts it can be quite straightforward to devise simple algorithms for tasks that operate on human-intelligible concepts. 
The crucial question is, under what conditions does such a transparent algorithm constitute a \emph{faithful interpretation} \citep{Jacovi2020} of the known, but opaque, low-level details of a black box model? This is the motivating question for the explainable AI subfield known as \emph{interpretability}. The question takes on particular significance for \emph{mechanistic interpretability},\footnote{\cite{Saphra2024} argue that the term `mechanistic interpretability' often signals not a single research program, but rather a cultural identity within the AI alignment community \citep{olah2020zoom,  elhage2021mathematical,  wang2023interpretability, Nandagrok}. However, they also note that there is a natural `narrow' technical reading of the term, where the concern is specifically with \emph{causal} analyses of models' internal mechanisms \citep{vig2020causal, geiger-etal:2020:blackbox,geiger2021causal,Geiger-etal:2023:DAS, finlayson-etal-2021-causal, Meng:2022, Stolfo23, Todd2024,prakash2024finetuning, mueller2024questrightmediatorhistory}.  We embrace that causality is core to mechanistic interpretability.} which, in contrast to behavioral interpretability, is precisely aimed at reverse engineering the \textit{internals} of a black box model in terms of a transparent algorithm.  %

Mechanistic interpretability research is quite analogous to the problem that cognitive scientists face in understanding how the human mind works. At one extreme, we can try to understand minds at a very low level, e.g., of biochemical processes in the brain. At the other extreme, we can focus just on the input-output facts of the system, roughly speaking, on  `observable behavior.' Analogously, for a deep learning model, we can focus either on low-level features (weight tensors, activation functions, etc.), or on what input-output function is computed. In both cases, however, it can be illuminating to investigate the mediating processes and mechanisms that transform input to output at a slightly higher-level of abstraction. This is what \cite{Marr} famously called the \emph{algorithmic} level of analysis. To the extent that these algorithmic-level hypotheses are transparent to the scientist, we may have a useful elucidation of the agent's inner workings. 

However, it is crucial that mechanistic interpretability methods avoid telling `just-so' stories that are completely divorced from the internal workings of the model. To clarify what this means exactly, we need a common language for explicating and comparing methodologies, and for precisifying core concepts. We submit that the theory of \emph{causal abstraction} provides this common language. 

In some ways, modern deep learning models are like the weather or an economy: they involve large numbers of densely connected `microvariables' with complex, non-linear dynamics. One way of reining in this complexity is to find ways of understanding these systems in terms of higher-level, more abstract variables (`macrovariables'). For instance, the many microvariables might be clustered together into more abstract macrovariables. A number of researchers have been exploring  theories of causal abstraction, providing a mathematical framework for causally analyzing a system at multiple levels of detail \citep{Chalupka,Rubinstein2017,BackersHalpern2019,beckers20a,Rischel, Massidda2023, massidda2024learning}. These methods tell us when a high-level causal model is a simplification of a (typically more fine-grained) low-level model. To date, causal abstraction has been used to analyze weather patterns \citep{chalupka2016}, human brains \citep{Dubois2020PersonalityBT,Dubois2020CausalMO}, physical systems \citep{Kekic2023}, batteries \citep{Zennaro23}, epidemics \citep{Dyer2023}, and deep learning models \citep{Chalupka:2015, geiger2021causal, hu22b, Geiger-etal:2023:DAS, Wu:Geiger:2023:BDAS}.

Macrovariables will not always correspond to sets of microvariables. Just as with neural network models of human cognition \citep{Smolensky1988a}, this is the typical situation in mechanistic interpretability, where high-level concepts are thought to be represented by modular `features' distributed across individual neural activations \citep{Harradon:2018,olah2020zoom, huang2024ravel}. For example, the linear subspaces of activation space learned from distributed alignment search \citep{Geiger-etal:2023:DAS} and the output dimensions of sparse autoencoders \citep{bricken2023monosemanticity, Cunningham:2023} are features that are distributed across overlapping sets of neural activations. 

Our first contribution is to extend the theory of causal abstraction to remove this limitation, building heavily on previous work. The core issue is that typical hard and soft interventions replace variable mechanisms entirely, so they are unable to isolate quantities distributed across overlapping sets of microvariables. To address this, we consider a very general type of intervention---what we call \textit{interventionals}---that maps from old mechanisms to new mechanisms. While this space of operations is generally unconstrained, we isolate special classes of interventionals that form \textit{intervention algebras}, satisfying two key modularity properties. Such classes can essentially be treated as hard interventions with respect to a new (`translated') variable space. We elucidate this situation, generalizing earlier work by \cite{Rubinstein2017} and \cite{BackersHalpern2019}.

Our second contribution is to show how causal abstraction provides a solid theoretical foundation for the field of mechanistic interpretability. We leverage our general presentation to provide flexible, yet mathematically precise, definitions for the core mechanistic interpretability concepts of polysemantic neurons, the linear representation hypothesis, modular features, and graded faithfulness. Furthermore, we unify a wide range of existing interpretability methodologies in a common language, including activation and path patching, causal mediation analysis, causal scrubbing, causal tracing, circuit analysis, concept erasure, sparse autoencoders, differential binary masking, distributed alignment search, and activation steering. We also connect the behavioral interpretability methods of LIME, integrated gradients, and causal effect estimation to the language of causal abstraction.

We are optimistic about productive interplay between theoretical work on causal abstraction and applied work on mechanistic interpretability. In stark contrast to weather, brains, or economies, we can measure and manipulate the microvariables of deep learning models with perfect precision and accuracy, and thus empirical claims about their structure can be held to the highest standard of rigorous falsification through experimentation.

\newpage

\section{Causality and Abstraction} \label{section:causalmodels}
This section presents a general theory of causal abstraction. Although we build on much existing work in the recent literature, our presentation is in some ways more general, and in other ways less so. Due to our focus on (deterministic) neural network models, we do not incorporate probability into the picture. At the same time, because the operations employed in the study of modern machine learned system go beyond `hard' and `soft' interventions that replace model mechanisms (see Def. \ref{def:intervention} and \ref{def:softintervention} below), we define a very general kind of intervention, the \textit{interventional}, which is a functional mapping from old mechanisms to new mechanisms (see Def. \ref{def:interventional}). In order to impose structure on this unconstrained class of model transformations, we establish some new results on classes of interventionals that form what we will call \emph{intervention algebras} (see especially Theorems \ref{thm:rep}, \ref{thm:repsoft}). 

Next, we explore key relations that can hold between causal models. 
We begin with \textit{exact transformations} \citep{Rubinstein2017}, which characterize when the mechanisms of one causal model are realized by the mechanisms of another (`causal consistency'). We generalize exact transformation from hard interventions to interventionals that form intervention algebras (see Def. \ref{def:exact}). A bijective translation (see Def. \ref{def:bijtrans}) is an exact transformation that retains all the details in the original model, staying at the same level of granularity. On the other hand, a constructive causal abstraction (see Def. \ref{def:cc}) is a `lossy' exact transformation that merges microvariables into macrovariables, while maintaining a precise and accurate description of the original model. Also, we (1) decompose constructive causal abstraction into three operations, namely marginalization, variable merge, and value merge (see Prop. \ref{prop:decompose}), %
and (2) provide a framework for approximate transformations (see Def. \ref{def:approx}). 

Lastly, we define a family of \emph{interchange intervention} operations, which are central to understanding mechanistic interpretability through the lens of causal abstraction. We begin with simple interchange interventions (see Def. \ref{def:interchange}), where a causal model with input and output variables has certain variables fixed to values they would have under different input conditions. We extend these to recursive interchange interventions (see Def. \ref{def:recinterchange}), which allow variables to be fixed based on the results of previous interchange interventions. Crucially, we also define distributed interchange interventions that target variables distributed across multiple causal variables and involve a bijective translation to and from a transformed variable space (see Def. \ref{def:distinterchange}). We conclude by explicating how to construct an alignment for interchange intervention analysis and how to use interchange intervention accuracy to quantify approximate abstractions.

\subsection{Deterministic Causal Models with Implicit Graphical Structure}
We start with some basic notation. 
\begin{remark}[Notation throughout the paper]\label{rem:notation}

Capital letters (e.g., $\onevar$) are used for variables and lower case letters (e.g., $\oneval$) are used for values. Bold faced letters (e.g. $\vars$ or $\pset$) are used for sets of variables and sets of values. %
When a variable (or set of variables) and a value (or set of values) have the same letter, the values correspond to the variables (e.g., $\oneval \in \values{\onevar}$ or $\pset \in \values{\vars}$).
\end{remark}

\begin{definition}[Signature]\label{def:signature} We use $\allvars$ to denote a fixed set of variables, each $\onevar \in \allvars$ coming with a non-empty range $\values{\onevar}$ of possible values. 
Together $\signature = (\allvars ,\values{})$ are called a \emph{signature}. 
\end{definition} 

\begin{definition}[Partial and Total Settings] \label{def:settings}

We assume $\values{\onevar} \cap \values{\onevarr} = \varnothing$ whenever $\onevar \neq \onevarr$, meaning no two variables can take on the same value.\footnote{To allow the same value to occur multiple times, we can simply take any causal model where variables share values, and then `tag' the shared values with variable names to make them unique.} This assumption allows representing the values of a set of variables $\vars \subseteq \allvars$ as a set $\pset$ of values, with exactly one value $\oneval \in \values{\onevar}$ in $\pset$ for each $\onevar \in \vars$. When we need to be specific about the choice of variable $\onevar$, we denote this element of $\values{\onevar}$ as $\oneval$.
We thus have $\pset \subseteq \bigcup_{\onevar\in \vars} \values{\onevar}$, and we refer to the values $\pset \in \values{\vars} $  as  \emph{partial settings}. 
In the special case where $\tset\in \values{\allvars}$, we call $\tset$ a \emph{total setting}. 
\end{definition}

Another useful construct in this connection is the \emph{projection} of a partial setting:
\begin{definition}[Projection]\label{def:projection}

Given a partial setting $\psett$ for a set of variables $\varrs \supseteq \vars$, we define $\project{\psett}{\vars}$ to be the restriction of $\psett$ to the variables in $\vars$. Given a partial setting $\pset$, we define the inverse:
\[ \inverseproject{\pset}{\varrs} = \{\psett \in \values{\varrs}: \project{\psett}{\vars} = \pset \}. \] 
\end{definition}

\begin{definition} \label{def:causalmodel}
A (deterministic) \emph{causal model} is a pair $\model = (\signature,\mechanisms{\onevar}{\allvars})$, such that $\signature$ is a signature and $\mechanisms{\onevar}{\allvars}$ is a set of \emph{mechanisms}, with $\mechanism{\onevar}: \values{\allvars} \rightarrow \values{\onevar}$ assigning a value to $\onevar$  as a function of the values of all the variables, including $\onevar$ itself. We write $\mechanism{\vars}$ for $\{\mechanism{\onevar}\}_{\onevar \in \vars}$. We will often break up the input argument to a mechanism into partial settings that form a total setting, e.g., $\mechanism{\onevar}(\psett, \psettt)$ for $\varrs \cup \varrrs = \allvars$.
\end{definition}

\begin{remark}[Inducing Graphical Structure]\label{rem:graph}
Observe that our definition of causal model makes no explicit reference to a graphical structure defining a causal ordering on the variables. While the mechanism for a variable takes in total settings, it might be that the output of a mechanism depends only on a subset of values. This induces a \textit{causal ordering} among the variables, such that $\onevarr \prec \onevar$---or $\onevarr$ is a \emph{parent} of $\onevar$---just in case there is a setting $\psettt$ of the variables $\varrrs = \allvars\setminus\{\onevarr\}$, and two settings $\onevall,\onevall'$ of $\onevarr$ such that $\mechanism{\onevar}(\psettt,\onevall) \neq \mechanism{\onevar}(\psettt,\onevall')$ (see, e.g., \citealt{Woodward2003}). The resulting order $\prec$ captures a notion of direct causation, which can be extended to indirect causation by taking its transitive closure $\prec^*$. Throughout the paper, we will define mechanisms to take in partial settings of parent variables, though technically they take in total settings and depend only on the parent variables.
\end{remark}

When $\prec^*$ is irreflexive, we say the causal model is \textit{acyclic}. Most of our examples of causal models will have this property. However, it is often also possible to give causal interpretations of cyclic models (see, e.g., \citealt{Bongers}). Indeed, the abstraction operations to be introduced generally create cycles among variables, even from initially acyclic models (see \citealt[\S 5.3]{Rubinstein2017} for an example). In Section~\ref{sec:future}, we provide an example where we abstract a causal model representing the bubble sort algorithm into a cyclic model where any sorted list is a solution satisfying the equations. 

\begin{remark}[Acyclic Model Notation]\label{rem:acyc}
Our example in Section~\ref{sec:example} will involve causal abstraction  between two finite, acyclic causal models. We will call variables that depend on no other variable \textit{input variables} ($\inputvars$), and variables on which no other variables depend \textit{output variables} ($ \outputvars$). The remaining variables are \textit{intermediate variables}. We can intervene on input variables $\inputvars$ to `prime' the model with a particular input.\footnote{In some parts of the literature what we are calling input variables are designated as `exogenous' variables.}  As such, the constant functions for input variables in our examples will be overwritten.
\end{remark}

As $\model$ can also be interpreted simply as a set of equations, we can define the set of solutions, which may be empty.
\begin{definition}[Solution Sets]\label{def:solutionsets}
Given $\model = (\allvars,\mechanisms{\onevar}{\allvars})$, the set of solutions, called $\sols(\model)$, is the set of all $\tset \in \values{\allvars}$ such that all the equations $\project{\tset}{\onevar}= \mechanism{\onevar}(\tset)$ are satisfied for each $\onevar \in \allvars$. When $\model$ is acyclic, there is a single solution and we use $\sols(\model)$ to refer interchangeably to a singleton set of solutions and its sole member, relying on context to disambiguate.
\end{definition}

We give a general definition of intervention on a model (see, e.g., \citealt{Spirtes,Woodward2003,Pearl2009}). For the following, assume we have fixed a signature $\Sigma$.

\begin{definition}[Intervention] \label{def:intervention} Define a \textit{hard intervention} to be a partial setting $\intpset \in  \values{\intvars}$ for finite $\intvars \subseteq \allvars$. %
Given a model $\model$ with signature $\signature$, define $\M_{\intpset}$ to be just like $\M$, except that we replace $\mechanism{\onevar}$ with the constant function $\tset \mapsto \project{\intpset}{\onevar}$ for each $\onevar \in \intvars$. 
Define $\allints_{\vars} = \values{\vars}$ to be the set of all hard interventions on $\vars$. 
\end{definition}

\begin{definition}[Soft Intervention] \label{def:softintervention}  We define a \textit{soft intervention} on some finite set of variables $\vars \subseteq \allvars$ to be a family of functions $\softintfull{\onevar}{\vars}$ where $\softint_{\onevar}:\values{\allvars} \to \values{\onevar}$ for each $\onevar \in \vars$. %
Given a model $\model$ with signature $\signature$, define $\M_{\softint}$ to be just like $\M$, except that we replace each function $\mechanism{\onevar}$ with the function $\softint_{\onevar}$. Define $\allsoftints_{\vars}$ to be the set of all soft interventions on $\vars$.
\end{definition}
Soft interventions generalize hard interventions. The hard intervention $\intpset$ is equivalent to a constant soft intervention $\softint = \{\tset \mapsto \oneval\}_{\oneval \in \intpset}$.

\begin{definition}[Interventional] \label{def:interventional}  We define an \textit{interventional} on some finite set of variables $\vars \subseteq \allvars$ to be a family of functions $\intal = \intalfull{\onevar}{\vars}$ where $ \intal_{\onevar}:\allsoftints_{\vars} \to \allsoftints_{\onevar}$  for each $\onevar \in \vars$. %
We define $\M_{\intal}$ to be just like $\M$, except that we replace each function $\mechanism{\onevar}$ with the function $\intal_\onevar\langle \mechanism{\vars}\rangle$. Define $\allintals_{\vars}$ to be the set of interventionals on $\vars$.
\end{definition}

Interventionals generalize soft interventions. The soft intervention $\softint$ is  a constant interventional---namely, the family of functions that, for each $X \in \mathbf{X}$, sends any element of $\allsoftints_{\vars}$ to $\softint_{\onevar}$. When only one variable is targeted, we say that the intervention(al) is \emph{atomic}. 

\begin{remark}[On Terminology] %
What we are calling hard interventions are sometimes called \emph{structural} interventions \citep{EberhardtScheines}. Our soft interventions are essentially the same as the soft interventions studied, e.g., in \cite{Tian2008,BareinboimCorrea}, although as mentioned, we set aside probabilistic aspects of causal models in this paper. The main difference between soft interventions and interventionals is that the latter `mechanism replacements' can depend on the previous mechanisms in the model. This type of dependence has also been studied, e.g., in the setting of so-called \emph{parametric} interventions \citep{EberhardtScheines}. 
\end{remark}

\begin{remark}[Interventionals as Unconstrained Model Transformations]
The set\\  $\allintals_{\allvars}$ contains the interventionals %
that targets \textit{all} variables $\allvars$. This set of interventionals is quite general, containing every function that maps from and to causal models with the same signature. 
\end{remark}

\begin{remark}[Composing Intervention(al)s]  Interventionals containing families of constant functions to constant functions are exactly the set of hard interventions. Similarly, interventionals containing families of constant functions are exactly the set of soft interventions. As such, we treat all three types of interventions as interventionals on all variables, i.e., $\bigcup_{\vars \subseteq \allvars} \allints_{\vars} \subseteq \bigcup_{\vars \subseteq \allvars} \allsoftints_{\vars}  \subseteq \bigcup_{\vars \subseteq \allvars} \allintals_{\vars} \subseteq \allintals_{\allvars}$. This allows us to understand the composition of interventions as function composition. We simplify notation by writing, e.g., $\oneval \,\circ\, \onevall$ for the composition of hard interventions (setting $\onevar$ to $\oneval$ and then $\onevarr$ to $\onevall$).\footnote{Note that we write composition in the opposite order from common notation for function composition, following the more intuitive order of intervention composition adopted, e.g., in \cite{Rubinstein2017}.}
\end{remark}
The unconstrained space of interventionals $\allintals_{\allvars}$ is unruly, and there is no guarantee that an interventional can be thought of as isolating a natural model component. We want to characterize spaces of interventionals that `act like hard interventions' insofar as they possess a basic algebraic structure. We elaborate on this in the next section. 

The following is an example of a causal model with hard interventions, soft interventions, and interventionals defined on it.

\begin{example}
Define a signature $\signature$ with variables $\allvars = \{\onevar, \onevarr\}$ with values $\values{\onevar} = \{0, \dots, 9\}$ and $\values{\onevarr} = \{\mathsf{True}, \mathsf{False}\}$. Define a model $\model$ with mechanisms $\mechanism{\onevar}(\tset) = 0$ (recall that input variables have no parents and are mapped to constants per Remark~\ref{rem:acyc}) and $\mechanism{\onevarr}(\tset) = [\project{\tset}{\onevar} > 5]$. The graphical structure of $\model$ has one directed edge from $\onevar$ to $\onevarr$ because $\onevar \prec \onevarr$. Per Remark~\ref{rem:graph}, we could define the mechanism for $\onevarr$ as $\mechanism{\onevarr}(\oneval) = [\oneval > 5]$ omitting the value that $\onevarr$ does not depend on, namely its own.

Define the hard intervention $\onevall = \{\mathsf{True}\} \in \allints_{\onevarr}$, the soft intervention $\softint = \tset \mapsto [\project{\tset}{\onevar} \leq 5] \in \allsoftints_{\onevarr}$, and the interventional $\intal = \mechanism{\onevarr} \mapsto (\tset \mapsto \lnot \mechanism{\onevarr}(\tset)) \in \allintals_{\onevarr}$. The model $\model_{\onevall}$ has mechanisms $\mechanism{\onevar}(\tset) = 0$ and $\mechanism{\onevarr}(\tset) = \mathsf{True}$ and a graphical structure with no edges. The models $\model_{\softint}$, $\model_{\softint\circ \softint}$, and $\model_{\intal}$ all have the mechanisms $\mechanism{\onevar}(\tset) = 0$ and $\mechanism{\onevarr}(\oneval) = [\oneval \leq 5]$ and the same graphical structure as $\model$. The model $\model_{\intal \circ \intal}$ is identical to $\model$.

Define the interventional $\intall = \{\mechanism{\onevar}, \mechanism{\onevarr}\} \mapsto \{ \tset \to 6 \times \indicator{\project{\tset}{\onevarr}},\tset \mapsto [\lnot \mechanism{\onevarr}(\tset)] \} \in \allintals_{\allvars}$. The model $\model_{\intall}$ has mechanisms $\mechanism{\onevar}(\onevall) = 6 \times \indicator{\onevall}$ and $\mechanism{\onevarr}(\oneval) = [\oneval \leq 5]$ 
with a cyclic graphical structure where $\onevar \prec \onevarr$ and $\onevarr \prec \onevar$. This model has no solutions. The model $\model_{\intall \circ \intall}$ has mechanisms $\mechanism{\onevar}(\onevall) = 6 \times \indicator{\onevall}$ and $\mechanism{\onevarr}(\oneval) = [\oneval > 5]$ with a cyclic graphical structure where $\onevar \prec \onevarr$ and $\onevarr \prec \onevar$. The solutions to this model are $\{6, \mathsf{True}\}$ and $\{0, \mathsf{False}\}$.
\end{example}

\subsection{Intervention Algebras} \label{sec:intalgebras}
 We are interested in the subsets of $\allintals_{\allvars}$ %
 that are well-behaved in the sense that they share an algebraic structure with hard interventions under the operation of function composition. The relevant algebraic structure is captured in the next definition.

\begin{definition} Let $\Lambda$ be a set and $\oplus$ be a binary operation on $\Lambda$. We define $(\Lambda, \oplus)$ to be an \emph{intervention algebra} if there exists a signature $ \signature = (\allvars, \values{})$ such that $(\hardintals,\circ) \simeq (\Lambda, \oplus)$---that is, these structures are isomorphic---where $\hardintals$ is the set of all constant functionals mapping to constant functions (i.e., hard interventions) for signature $\signature$, and $\circ$ is function composition.
\end{definition}
As a matter of fact, intervention algebras can be given an intuitive characterization based on two key properties of hard interventions. 
\begin{definition}[Commutativity and Left-Annihilativity]\label{keyproperties}
Hard interventions $\oneval, \onevall \in \hardintals$  under function composition have the following properties:
\begin{enumerate}[label=(\alph*)]
    \item If hard interventions target different variables, $\circ$ is commutative:\newline if $\onevar \neq \onevarr$ then $\oneval \circ \onevall = \onevall \circ \oneval$; \label{keyproperty1}
    \item If hard interventions target the same variable, $\circ$ is left-annihilative:\newline if $\onevar = \onevarr$, then $\oneval \circ \onevall = \onevall$; \label{keyproperty2}
\end{enumerate} Note equality signifies the compositions are the very same functions from models to models.
\end{definition} These two properties highlight an important sense in which hard interventions are modular: when intervening on two different variables, the order does not matter; and when intervening on the same variable twice, the second undoes any effect of the first intervention.

We can use commutativity and left-annihilativity to build an equivalence relation that we will show captures the fundamental algebraic structure of hard interventions.
\begin{definition}
Let $A$ be any set with equivalence relation $\sim$, and define $(A^*,\cdot)$ to be the free algebra generated by elements of $A$ under concatenation. Define $\approx$ to be the smallest congruence on $A^*$ extending the following: \begin{equation}\{\langle \oneval\cdot \onevall, \onevall\cdot \oneval \rangle : \oneval \not \sim \onevall\} \cup \{\langle \onevall\cdot \oneval, \oneval \rangle: \oneval \sim \onevall\},\label{eqn:basicrel}\end{equation} for all $\oneval,\onevall \in A$, where $\cdot$ is concatenation in $A^*$. 
\end{definition}
As it turns out, $\approx$ can be obtained constructively as the result of two operations that define a normal form for sequences of atomic hard interventions.
\begin{definition}[Normal form]
    Let $A$ be a set equipped with equivalence relation $\sim$. Fix an order $\algprec$ on $\sim$-equivalence classes.  Define $\mathsf{Collapse}: A^* \to A^*$ to take a sequence and remove every element that has an $\sim$-equivalent element that occurs to its right. Define $\mathsf{Sort}: A^* \to A^*$ to take a sequence and sort it according to $\algprec$. For any element $\seq \in A^*$, we call $\mathsf{Sort}(\mathsf{Collapse}(\seq))$ the \emph{normal form} of $\seq$. This normal form clearly exists and is unique.
\end{definition}
\begin{lemma} For $\seq,\seqq \in A^*$, we have $\seq \approx \seqq$ iff $\mathsf{Sort}(\mathsf{Collapse}(\seq)) = \mathsf{Sort}(\mathsf{Collapse}(\seqq))$, that is, iff $\seq$ and $\seqq$ have the same normal form. \label{lemma:equivs}
\end{lemma}
\begin{proof} Let us write $\seq \equiv \seqq$ when $\mathsf{Sort}(\mathsf{Collapse}(\seq)) = \mathsf{Sort}(\mathsf{Collapse}(\seqq))$. First note that $\equiv$ is a congruence extending the relation in (\ref{eqn:basicrel}), and hence $\approx \;\subseteq\; \equiv$. 

For the other direction, it suffices to observe that both  $\mathsf{Collapse}(\mathbf{c}) \approx \mathbf{c}$ and  $\mathsf{Sort}(\mathbf{c}) \approx \mathbf{c}$, for any $\mathbf{c} \in A^*$. This follows from the fact that $\approx$ is a congruence extending (\ref{eqn:basicrel}). Then if $\seq \equiv \seqq$, we have $\mathsf{Sort}(\mathsf{Collapse}(\seq)) \approx \mathsf{Sort}(\mathsf{Collapse}(\seqq))$ (by reflexivity of $\approx$), and moreover: \begin{eqnarray*}
    \seq & \approx & \mathsf{Sort}(\mathsf{Collapse}(\seq)) \\
    & \approx & \mathsf{Sort}(\mathsf{Collapse}(\seqq)) \\
    & \approx & \seqq,
\end{eqnarray*} and hence $\equiv \; \subseteq \; \approx$. Consequently, $\seq \approx \seqq$ iff $\seq$ and $\seqq$ have the same normal form. 
\end{proof}
The foregoing produces a representation theorem for intervention algebras. 
\begin{theorem}
The quotient $(A^*/\!\approx, \odot)$ of $(A^*,\cdot)$ under $\approx$, 
 with $\odot$ defined so\begin{eqnarray*}
       [\seq]_{\approx} \odot [\seqq]_{\approx} & = & [\seq \cdot \seqq]_\approx,
   \end{eqnarray*} is a intervention algebra. \label{thm:rep}
\end{theorem}
\begin{proof}
Define a signature $\signature$ where the variables $\allvars$ are the $\sim$-equivalence classes of $A$ and the values of each variable are the members of the respective $\sim$-equivalence class. We need to show that $(A^*/\!\approx, \odot)$ is isomorphic to $(\hardintals,\circ)$, the set of all (finite) hard interventions with function composition.

Begin by defining the map $\iota^*:A^* \rightarrow \hardintals$, where $\iota^* (\oneval_1\cdot {{ }_{\dots}} \cdot \oneval_n) = \oneval_1 \circ {{ }_{\dots}} \circ \oneval_n$. First observe: 
\begin{equation}
    \label{eq:collapseSort}
\iota^*(\seq) = \iota^*(\mathsf{Collapse}(\seq)) = \iota^*(\mathsf{Sort}(\mathsf{Collapse}(\seq)))
\end{equation} Eq. (\ref{eq:collapseSort}) follows from commutativity and left-annihilativity (Def. \ref{keyproperties}).

Moreover, again by commutativity and left-annihilativity, if $\onevar=\onevarr$ but $\onevar \notin \varrrs$, then $\oneval \circ \psettt \circ \onevall = \psettt \circ \onevall$, which is precisely what justifies the first equality. 

Finally, define $\iota:A^*/\! \approx\; \to \hardintals$ so that $\iota([\seq]_{\approx}) = \iota^*(\seq)$. This is well-defined by Eq. (\ref{eq:collapseSort}) and Lemma \ref{lemma:equivs}. It is surjective because $\iota^*$ is surjective. It is also injective: if $\mathsf{Sort}(\mathsf{Collapse}(\seq)) \neq \mathsf{Sort}(\mathsf{Collapse}(\seqq))$, then pick the $\algprec$-least $\sim$-equivalence class of $A$---that is, the $\algprec$-least variable $\onevar$---such that $\mathsf{Sort}(\mathsf{Collapse}(\seq))$ and $\mathsf{Sort}(\mathsf{Collapse}(\seqq))$ disagree on $\onevar$. Without loss, we can assume $\iota^*(\mathsf{Sort}(\mathsf{Collapse}(\seq)))$ assigns $\onevar$ to some value $\oneval$, but  $\iota^*(\mathsf{Sort}(\mathsf{Collapse}(\seqq)))$ does not assign $\onevar$ to $\oneval$ (either because it does not assign $\onevar$ to any value or because it assigns $\onevar$ to a different value). In any case, if $[\seq]_{\approx} \neq [\seqq]_{\approx}$, then $\iota([\seq]_{\approx}) = \iota^*(\mathsf{Sort}(\mathsf{Collapse}(\seq)))\neq  \iota^*(\mathsf{Sort}(\mathsf{Collapse}(\seqq))) = \iota([\seqq]_{\approx})$.

That $\iota$ is an isomorphism follows from the sequence of equalities below:
\begin{align*}
 \iota([\seq]_{\approx} \odot [\seqq]_{\approx}) & =  \iota([\seq \cdot \seqq]_{\approx}) & \text{By the definition of } \odot \\
  & =  \iota(\mathsf{Sort}(\mathsf{Collapse}(\seq\cdot\seqq))) & \text{By Lemma \ref{lemma:equivs}} \\
  & =  \iota^*(\mathsf{Sort}(\mathsf{Collapse}(\seq\cdot\seqq))) & \text{By the definition of } \iota \\
  & =  \iota^*(\seq\cdot \seqq) & \text{By Equation } (\ref{eq:collapseSort})\\
  & =  \iota^*(\seq) \circ \iota^*(\seqq) & \text{By the definition of } \iota^* \\
  & =  \iota^*(\mathsf{Sort}(\mathsf{Collapse}(\seq))) \circ \iota^*(\mathsf{Sort}(\mathsf{Collapse}(\seqq))) & \text{By Equation } (\ref{eq:collapseSort}) \\
  & =  \iota(\mathsf{Sort}(\mathsf{Collapse}(\seq))) \circ \iota(\mathsf{Sort}(\mathsf{Collapse}(\seqq))) & \text{By the definition of } \iota \\
  & =  \iota([\seq]_{\approx}) \circ \iota([\seqq]_{\approx}) & \text{By Lemma \ref{lemma:equivs}}
\end{align*} This concludes the proof of Theorem \ref{thm:rep}. \end{proof}

Sets of atomic soft interventions also form intervention algebras:
\begin{theorem} Suppose $\intals_0$ is a set of atomic soft interventions with signature $\signature$. Let $\intals$ be the closure of $\intals_0$ under function composition. Then $(\intals,\circ)$ is an intervention algebra. \label{thm:repsoft}
\end{theorem}
\begin{proof} Just as in the proof of Theorem \ref{thm:rep} we can consider the free algebra generated by $A = \intals_0$, quotiented under $\approx$, to obtain $(A/\!\approx, \odot)$. The proof that this algebra is isomorphic to a set of hard interventions follows exactly as in the proof of Theorem \ref{thm:rep}, relying on the fact that soft interventions also satisfy the key properties \ref{keyproperty1} and \ref{keyproperty2} in Definition~\ref{keyproperties}: 
\begin{enumerate}[label=(\roman*)]
    \item If soft interventions target different variables, $\circ$ is commutative: if $\onevar \neq \onevarr$ then $\softint_{\onevar} \circ \softint_{\onevarr} = \softint_{\onevarr}  \circ \softint_{\onevar} $; 
    \item If interventions target the same variable, $\circ$ is left-annihilative: if $\onevar = \onevarr$, then $\softint_{\onevar}  \circ \softint_{\onevarr}  = \softint_{\onevarr} $. \label{keyprop2}
\end{enumerate} 
Consequently, we know that there exists a signature $\signature^*$ such that hard interventions on $\signature$ are isomorphic to the soft interventions $\intals$ with respect to function composition. Specifically, where $\intals^{\onevar}_0\subseteq \intals_0$ is the set of atomic soft interventions that target $\onevar$, the variables of $\signature^*$ are $\allvars^* = \{\onevar^*:  \intals^{\onevar}_0 \not = \emptyset \}$ and, for each $\onevar^* \in \allvars^*$, the values are $\values{\onevar^*} = \intals^{\onevar}_0$.
\end{proof}

\begin{remark} While both hard and soft interventions give rise to intervention algebras, the more general class of interventionals satisfies weaker algebraic constraints. For instance, it is easy to see that left-annihilativity (see \ref{keyproperty2} and \ref{keyprop2} above) often fails.  
Consider a signature $\signature$ with a single variable $\onevar$ that takes on binary values $\{0,1\}$. Define the interventional $\intal \langle \mechanism{\onevar}\rangle = \oneval \mapsto 1 - \mechanism{\onevar}(\oneval)$. Observe that $\intal \circ \intal \not = \intal$ because $\intal \circ \intal$ is the identity function, and so left-annihilativity fails. 
\end{remark}

While interventionals do not form intervention algebras in general, particular classes of interventionals can form intervention algebras.
\begin{example}
Consider a signature $\signature$ with variables $\{\onevar_1, \dots, \onevar_n\}$ that take on integer values. Define $\intals$ to contain interventionals that fix the $p$th digit of a number to be the digit $q$, where $\%$ is modulus and $//$ is integer division:
\[ \intal \langle \mechanism{\onevar_k} \rangle  = \tset \mapsto \mechanism{\onevar_k}(\tset) - \big((\mechanism{\onevar_k}(\tset) // 10^{p}) \mathbin{\%} 10\big) \cdot 10^p  + q \cdot 10^{p}\]
This class of interventionals is isomorphic to hard interventions on a signature $\signature^*$ with variables $\{\onevarr^{p}_0, \onevarr^{p}_1, \dots, \onevarr^{p}_n: p \in \{0,1,\dots\}\}$ that take on values $\{0, 1, \dots, 9\}$. The interventional fixing the $p$th digit of $\onevar_k$ to be the digit $q$ corresponds to the hard intervention fixing $\onevarr^{p}_k$ to the value $q$.
\end{example}

\begin{remark}[Assignment mutations] As a side remark, it is worth observing that another setting where intervention algebras appear is in programming language semantics and the semantics of, e.g., first-order predicate logic. Let $D$ be some domain of values and $\mathsf{Var}$ a set of variables. An \emph{assignment} is a function $g:\mathsf{Var}\rightarrow D$. Let $g^{x}_d$ be the \emph{mutation} of $g$, defined so that $g^{x}_d(y) = g(y)$ when $y \neq x$, and $g^\oneval_d(y) = d$ when $y=x$. Then a mutation $(\,\cdot\,)^\oneval_d$ can be understood as a function from the set of assignments to the set of assignments. Where $\mathfrak{M}$ is the set of all mutations, it is then easy to show that the pair $(\mathfrak{M},\circ)$ forms an intervention algebra. Furthermore, every intervention algebra can be obtained this way (up to isomorphism), for a suitable choice of $D$ and $\mathsf{Var}$. \end{remark}

\begin{definition}[Ordering on Intervention Algebras]
Let $(\Lambda,\oplus)$ be a intervention algebra. We define an ordering $\leq$ on elements of $\Lambda$ as follows: \begin{eqnarray*} \lambda \leq \lambda' & \mbox{ iff } & \lambda' \oplus \lambda = \lambda'.
\end{eqnarray*}
So, in particular, if the intervention algebra contains hard interventions---if it is of the form $(\hardintals,\circ)$---then this amounts to the ordering from \cite{Rubinstein2017}:
\begin{eqnarray*} \pset \leq \psett & \mbox{ iff } & \vars \subseteq \varrs \mbox{ and }\pset = \project{\psett}{\vars} \\
& \mbox{ iff } & \pset \subseteq \psett.
\end{eqnarray*} Note that $\leq$ is defined as in a semi-lattice, except that $\oplus$ (or $\circ$) is not commutative in general. So the order matters.
    
\end{definition}

\subsection{Exact Transformation with Interventionals}
Researchers have been interested in the question of when two models---potentially defined on different signatures---are compatible with one another in the sense that they could both accurately describe the same target causal phenomena. The next definition presents a deterministic variant of the notion of `exact transformation' from \cite{Rubinstein2017} (see also Def. 3.5 in \citealt{BackersHalpern2019}), generalized to interventionals. The other notions we study in this paper---namely, bijective translation and constructive abstraction---are special cases of exact transformation. 

\begin{definition}\label{def:exact}
Let $\model, \model^*$ be causal models and let $(\intals, \circ)$ and $(\intals^*, \circ)$ be two intervention algebras where $\intals$ and $\intals^*$ are interventionals on $\model$ and $\model^*$, respectively. Furthermore, let $\setmap: \values{\allvars} \to \values{\allvars^*}$ and $\intmap: \intals \to \intals^*$ be two partial surjective functions where $\intmap$ is $\leq$-preserving; that is, $\omega(\mathbb{I}) \leq \omega(\mathbb{I}')$ whenever $\mathbb{I} \leq \mathbb{I}'$.

Then $\model^*$ is an \emph{exact transformation} of $\model$ under $(\setmap, \intmap)$ if for all $\intal \in \dom(\intmap)$, the following diagram commutes:
\begin{center}
\begin{tikzpicture}
\node (n1) at (0,0) {$\intal$};
\node (n2) at (4,0) {$\intmap(\intal)$};
\path (n1) edge[->] (n2);
\node at (2,.25) {${\intmap}$};
\node (n3) at (0,-1.5) {$\sols\big(\model_{\intal}\big)$};
\node (n4) at (4,-1.5) {$\sols\big(\model^*_{\intmap(\intal)}\big) $};
\path (n3) edge[->] (n4);
\node at (2,-1.25) {$\setmap$};
\path (n1) edge[->] (n3);
\path (n2) edge[->] (n4);
\end{tikzpicture}
\end{center}
That is to say, the interventional $\intmap(\intal)$ on $\model^*$ results in the same total settings of $\allvars^*$ as the result of first determining a setting of $\allvars$ from $\intal$ and then applying the translation $\setmap$ to obtain a setting of $\allvars^*$. In a single equation:\footnote{
When evaluating whether $\setmap(S) = T$, as in, e.g.\ \eqref{eq:commute}, it is possible that not every element of $S$ is in the domain of $\setmap$, since $\setmap$ is partial.
Simply map such points to some distinguished element $\bot$.} 
\begin{eqnarray}\setmap\big(\sols(\model_{\intal})\big)  & = & \sols\big(\model^*_{\intmap(\intal)}\big).\label{eq:commute}\end{eqnarray}
\label{eq:exacttrans}
\end{definition} This definition captures the intuitive idea that $\model$ and $\model^*$ are consistent descriptions of the same causal situation. 

 \begin{remark} Definition~\ref{def:exact} is a variant of exact transformation in \cite{Rubinstein2017} with intervention algebras. However, their definition of exact transformation includes an existential quantifier over $\intmap$, stating that $\model^*$ is an exact transformation of $\model$ under $\setmap$ if an $\intmap$ exists that satisfies the commuting diagram. Our definition tells us when a particular pair $\setmap$ and $\intmap$ constitute an exact transformation. We believe this difference to be inessential.
 \end{remark}

\begin{remark}
 The composition of exact transformations is an exact transformation. That is, if $(\setmap_1,\intmap_1)$ and $(\setmap_2,\intmap_2)$ are exact transformation, then so is $(\setmap_1\circ \setmap_2,\intmap_1\circ \intmap_2)$, when these compositions are all defined. (See Lemma~5 from \citealt{Rubinstein2017}.)
\end{remark}

\subsubsection{Bijective Translation}\label{sec:bijtrans}
Exact transformations can be `lossy' in the sense that $\model$ may involve a more detailed, or finer-grained, description than $\model^*$. For example, the model $\model$ could encode the causal process of computer hardware operating on bits of memory while the model $\model^*$ encodes the fully precise, but less detailed, process of assembly code the hardware implements. In contrast, when $\setmap$ is a bijective function, there is an important sense in which $\model$ and $\model^*$ are just two equivalent (and inter-translatable) descriptions of the same causal setup. 

\begin{definition}[Bijective Translation]\label{def:bijtrans}
Fix signatures $\signature$ and $\signature^*$. Let $\model$ be a causal model with signature $\signature$ and mechanisms $\mechanism{\allvars}$. Let $\setmap: \values{\allvars} \to \values{\allvars^*}$ be a bijective map from total settings of $\signature$ to total settings of $\signature^*$. Define $\setmap(\model)$ to be the causal model with signature $\signature^*$ and mechanisms \[\mechanism{\onevar^*}(\tset^*) = \project{\setmap(\mechanism{\allvars}(\setmap^{-1}(\mathbf{\tset^*})))}{\onevar^*}\] for each variable $\onevar^* \in \allvars^*$.
We say that $\setmap(\model)$ is the bijective translation of $\model$ under $\setmap$. \label{def:translation}
\end{definition}

\begin{remark}[Bijective Translations Define a Canonical $\intmap$]
Let $\hardintals^*$ be the intervention algebra formed by hard interventions on $\signature^*$. We will now construct an intervention algebra $\intals$ consisting of interventionals on $\signature$ and define a function $\intmap: \intals \to \hardintals^*$.

For each $\intpset^* \in \allints_{\intvars^*}$ with $\intvars^* \subseteq \allvars^*$ , define the interventional using notation from Definition~\ref{def:interventional},
\[\intal\langle \mechanism{\allvars}\rangle \ \ \ = \ \ \  \tset \mapsto \setmap^{-1}\Bigg(\Project{\setmap\big(\mechanism{\allvars}(\tset)\big)}{\allvars^* \setminus \intvars^* } \cup \intpset^* \Bigg).\]
We add $\intal$ to $\intals$ and define $\intmap(\intal) = \intpset^*$. The interventional $\intal$ takes in a set of mechanisms $\mechanism{\allvars}$ for all of the variables and outputs a new set of mechanisms $\intal\langle \mechanism{\allvars}\rangle$ for all of the variables. To retrieve mechanisms for individual variables, a projection must be applied. By construction $(\intals, \circ)$ is isomorphic to $(\hardintals^*, \circ)$ with the $\leq$-order preserving (and reflecting) map $\intmap$, so $(\intals, \circ)$ is an intervention algebra.
\end{remark}

\begin{theorem}[Bijective Translations are Exact Transformations]\label{thm:bijexact}
 The bijective\\ translation $\model^* = \setmap(\model)$ is an exact transformation of $\model$ under $(\setmap, \intmap)$, relative to $\intals$ and\\ $\hardintals^*$ as constructed above.
\end{theorem}

\begin{proof}
Choose an arbitrary $\intal \in \intals$ with corresponding hard intervention $\intpset^* \in \hardintals^*$ where $\intmap(\intal) = \intpset^*$. Let $\mechanism{ }^*$ be the mechanisms of $\model^*$ and $\mechanismm{ }^*$ be the mechanisms of $\model^*_{\intpset^*}$.

Fix an arbitrary solution $\tset \in \sols(\model_{\intal})$. The following string of equalities shows that $\setmap(\tset)$ is in $\sols(\model^*_{\intpset^*})$ and therefore $\setmap(\sols(\model_{\intal})) \subseteq \sols(\model^*_{\intpset^*})$.
\begin{align*}
\footnotesize
 \setmap(\tset) &= \setmap(\intal\langle \mechanism{\allvars}\rangle(\tset)) & \text{\footnotesize By the definition of a solution} \\
  &= \setmap\Bigg(\setmap^{-1}\bigg(\Project{\setmap\big(\mechanism{\allvars}(\tset)\big)}{\allvars^* \setminus \intvars^*} \cup \intpset^*\bigg) \Bigg) & \text{\footnotesize By the definition of } \intal \\
  &= \Project{\setmap\big(\mechanism{\allvars}(\tset)\big)}{\allvars^* \setminus \intvars^* } \cup \intpset^* & \text{\footnotesize Inverses cancel} \\
  &= \Project{\mechanism{\allvars^*}^*(\setmap(\tset))}{\allvars^* \setminus \intvars^* } \cup \intpset^* & \text{\footnotesize $\model^*$ is a bijective translation of $\model$ under $\setmap$} \\
  &= \mechanismm{\allvars^*}^*(\setmap(\tset)) & \text{\footnotesize By the definition of hard interventions.} \\
\end{align*} 

 Fix an arbitrary solution $\tset^* \in \sols(\model^*_{\intpset^*})$. The following string of equalities show that $\setmap^{-1}(\tset^*)$ is in $\sols(\model_{\intal})$ and therefore $\setmap(\sols(\model_{\intal})) \supseteq \sols(\model^*_{\intpset^*})$.
\begin{align*}
 \setmap^{-1}(\tset^*) &= \setmap^{-1}(\mechanismm{\allvars^*}^*(\tset^*)) & \text{\footnotesize By the definition of a solution} \\
 &= \setmap^{-1}\bigg(\Project{\mechanism{\allvars^*}^*(\tset^*)}{\allvars^* \setminus \intvars} \cup \intpset^* \bigg) & \text{\footnotesize By the definition of hard interventions} \\
 &= \setmap^{-1}\bigg(\Project{\tau\big(\mechanism{\allvars}(\setmap^{-1}(\tset^*))\big)}{\allvars^* \setminus \intvars} \cup \intpset^*\bigg) & \text{\footnotesize$\model^*$ is a bij. trans. of $\model$ under $\setmap^{-1}$} \\
 &= \intal\langle \mechanism{\allvars}\rangle(\setmap^{-1}(\tset^*)) & \text{\footnotesize By the definition of $\intal$} \\
\end{align*} 
Thus, $\setmap(\sols(\model_{\intal})) = \sols(\model^*_{\intpset^*})$ for an arbitrary $\intal$, and we can conclude $\model^* = \setmap(\model)$ is an exact transformation of $\model$ under $(\setmap, \intmap)$.
\end{proof}

\begin{example}
If the variables of a causal model form a vector space, then a natural bijective translation is a rotation of a vector. Consider the following causal model $\model$ that computes boolean conjunction.

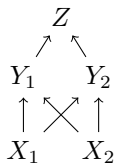
\begin{center}
\begin{tikzpicture}
\node[] (X1) {$\onevar_1$};
\node[ right of=X1] (X2) {$\onevar_2$};
\node[above of=X1] (Y1) {$\onevarr_1$};
\node[ above of=X2] (Y2) {$\onevarr_2$};
\node (Z) at (0.5, 1.8) {$\onevarrr$};

\draw[->] (X1) -> (Y1);
\draw[->] (X1) -> (Y2);
\draw[->] (X2) -> (Y1);
\draw[->] (X2) -> (Y2);
\draw[->] (Y1) -> (Z);
\draw[->] (Y2) -> (Z);
\end{tikzpicture}
\end{center}

The variables $\onevar_1$ and $\onevar_2$ take on binary values from $\{0,1\}$ and have constant mechanisms mapping to $0$. The variables $\onevarr_1$ and $\onevarr_2$ take on real-valued numbers and have mechanisms defined by a 20$\degree$ rotation matrix
\[\begin{bmatrix}\mechanism{\onevarr_1}(\oneval_1, \oneval_2) & \mechanism{\onevarr_2}(\oneval_1, \oneval_2)\end{bmatrix} = \begin{bmatrix} \oneval_1\\ \oneval_2\end{bmatrix}^\top \begin{bmatrix} \mathsf{cos}(20\degree) & -\mathsf{sin}(20\degree) \\ \mathsf{sin}(20\degree) & \mathsf{cos}(20\degree) \end{bmatrix} \]
The variable $\onevarrr$ takes on binary values from $\{0,1\}$ and has the mechanism \[\mechanism{\onevarrr}(\onevall_1, \onevall_2) = \indicator{(\mathsf{sin}(20\degree) + \mathsf{cos}(20\degree))\onevall_1 + (\mathsf{cos}(20\degree) - \mathsf{sin}(20\degree))\onevall_2 = 2}\] Note that $\onevarrr = \onevar_1 \land \onevar_2$, since $\mechanism{\onevarrr}$ un-rotates $\onevarr_1$ and $\onevarr_2$ before summing their values.

The variables $\onevarr_1$ and $\onevarr_2$ perfectly encode the values of the variables $\onevar_1$ and $\onevar_2$ using a coordinate system with axes that are tilted by 20$\degree$. We can view the model $\model$ through this tilted coordinate system using a bijective translation. Define the function
\[\tau\left(\begin{bmatrix}\oneval_1\\ \oneval_2\\ \onevall_1\\ \onevall_2 \\ \onevalll\end{bmatrix}\right) = \begin{bmatrix}\oneval_1\\ \oneval_2\\ \onevall_1\\ \onevall_2 \\ \onevalll\end{bmatrix}^\top \begin{bmatrix}\phantom{000}1\phantom{000} & \phantom{000}0\phantom{000} & 0 & 0 & \phantom{000}0\phantom{000}\\ 0 & 1 & 0 & 0 & 0\\ 0 & 0 & \mathsf{cos}(-20\degree) & -\mathsf{sin}(-20\degree) & 0  \\ 0 & 0 & \mathsf{sin}(-20\degree) & \mathsf{cos}(-20\degree) & 0 \\ 0 & 0 & 0 & 0 & 1 \end{bmatrix} \]
and consider the model $\tau(\model)$. This model will have altered mechanisms for $\onevarr_1$, $\onevarr_2$, and $\onevarrr$. In particular, these mechanisms are
\[\begin{bmatrix}\mechanism{\onevarr_1}(\oneval_1, \oneval_2) & \mechanism{\onevarr_2}(\oneval_1, \oneval_2)\end{bmatrix} = \begin{bmatrix} \oneval_1\\ \oneval_2\end{bmatrix}^\top \begin{bmatrix} 1 & 0 \\ 0 & 1 \end{bmatrix} \]
and $\mechanism{\onevarrr}(\onevall_1, \onevall_2) = \indicator{\onevall_1 + \onevall_2 = 2}$. 
\end{example}

\subsubsection{Constructive Causal Abstraction }\label{sec:constructive}
Suppose we have a `low-level model' $\lowmodel = (\signature_{\lowmodel},\mechanism{\lowmodel})$ built from `low-level variables' $\allvars_{\lowmodel}$ and a `high-level model' $\highmodel = (\signature_{\highmodel},\mechanism{\highmodel})$ built from `high-level variables' $\allvars_{\highmodel}$. 
What structural conditions must be in place for $\highmodel$ to be  a high-level \emph{abstraction} of the low-level model $\lowmodel$? 
At a minimum, this requires that the high-level interventions represent the low-level ones, in the sense of Def. \ref{def:exact}; $\highmodel$ should be an exact transformation of $\lowmodel$. What else must be the case?

A prominent further intuition about abstraction is that it may involve associating specific high-level variables with \emph{clusters} of low-level variables. That is, low-level variables are to be clustered together in `macrovariables' that abstract away from low-level details. To systematize this idea, we introduce alignment between a low-level and a high-level signature:
\begin{definition}[Alignment] \label{def:alignment} 

An \emph{alignment} between signatures $\signature_\lowmodel$ and $\signature_\highmodel$ is given by a pair $\langle \cellpart,\cellfunc\rangle$ of a partition $\cellpart=\{\cellpart_{\onevar_{\highmodel}}\}_{\onevar_{\highmodel} \in \allvars_{\highmodel} \cup \{\bot \}}$ and a family $\cellfunc= \{\cellfunc_{\onevar_{\highmodel}}\}_{\onevar_{\highmodel} \in \allvars_{\highmodel}}$ of maps, such that:
\begin{enumerate}
    \item The partition $\cellpart$ of $\allvars_{\lowmodel}$ consists of non-overlapping, non-empty cells $\cellpart_{\onevar_{\highmodel}} \subseteq \allvars_{\lowmodel}$ for each $\onevar_{\highmodel} \in \allvars_{\highmodel}$, in addition to a (possibly empty) cell $\cellpart_{\bot}$;
    \item There is a partial surjective map $\cellfunc_{\onevar_{\highmodel}}:\values{\cellpart_{\onevar_\highmodel}}\rightarrow\values{\onevar_{\highmodel}}$ for each $\onevar_{\highmodel} \in \allvars_{\highmodel}$. 
\end{enumerate}
\end{definition}
In words, the set $\cellpart_{\onevar_{\highmodel}}$ consists of those low-level variables that are `aligned' with the high-level variable $\onevar_{\highmodel}$, and $\cellfunc_{\onevar_{\highmodel}}$ tells us how a given setting of the low-level cluster $\cellpart_{\onevar_{\highmodel}}$ corresponds to a setting of the high-level variable $\onevar_{\highmodel}$. The remaining set $\cellpart_\bot$ consists of those low-level variables that are `forgotten', that is, not mapped to any high-level variable.

\begin{remark} An alignment $\langle \cellpart, \cellfunc\rangle$ induces a unique partial function $\intmap^{\cellfunc}$ that maps from low-level hard interventions to high-level hard interventions. We only define $\intmap^{\cellfunc}$ on low-level interventions that target full partition cells, excluding $\cellpart_{\bot}$. For $\pset_{\lowmodel} \in \values{\cellpart_{\vars_{\highmodel}}}$ where $\vars_{\highmodel} \subseteq \allvars_{\highmodel}$ and $\cellpart_{\vars_{\highmodel}} = \bigcup_{\onevar \in \vars_{\highmodel}} \cellpart_{\onevar}$, we define 
\begin{eqnarray} \intmap^{\cellfunc}(\pset_{\lowmodel}) & \overset{\textnormal{def}}{=} & \bigcup_{\onevar_{\highmodel} \in \allvars_{\highmodel}} \cellfunc_{\onevar_{\highmodel}} \big( \project{\pset_{\lowmodel}}{\cellpart_{\onevar_{\highmodel}}} \big). \label{eq:transl}
\end{eqnarray} 
As a special case of Eq. (\ref{eq:transl}), we obtain a unique partial function $\setmap^{\cellfunc}:\values{\allvars_{\lowmodel}} \rightarrow \values{\allvars_{\highmodel}}$ that maps from low-level total settings to high-level total settings. To wit,
for any $\tset_{\lowmodel} \in \values{\allvars_{\lowmodel}}$:
\begin{eqnarray} \setmap^{\cellfunc}(\tset_{\lowmodel}) & = & \bigcup_{\onevar_{\highmodel} \in \allvars_{\highmodel}} \cellfunc_{\onevar_{\highmodel}} \big( \project{\tset_{\lowmodel}}{\cellpart_{\onevar_{\highmodel}}} \big). \label{eq:settransl}
\end{eqnarray} 
Thus, the cell-wise maps $\cellfunc_{\onevar_{\highmodel}}$ canonically give us these partial functions $(\setmap^{\cellfunc}, \intmap^{\cellfunc})$. 
\end{remark}

\begin{definition}[Constructive Abstraction]\label{def:cc} 
We say that $\highmodel$ is a constructive abstraction of $\lowmodel$ under an alignment $\langle \cellpart, \cellfunc \rangle$ iff $\highmodel$ is an exact transformation of $\lowmodel$ under $(\setmap^{\cellfunc}, \intmap^{\cellfunc})$.
\end{definition}
See Section~\ref{sec:example} for an example of constructive causal abstraction.
\begin{remark} \label{remark:bijtrans}
Though the idea was implicit in much earlier work (going back at least to \citealt{Simon1961} and \citealt{Simon}), 
\cite{BackersHalpern2019} and \cite{beckers20a} explicitly introduced the notion of a constructive abstraction in the setting of probabilistic causal models.\footnote{\cite{BackersHalpern2019} require each $\cellfunc$ is total, while we allow partial functions for more flexibility.} As Examples 3.10 and 3.11 from \cite{BackersHalpern2019} show, there are exact transformations that are not also constructive abstractions, even when we restrict attention to hard interventions. 
\end{remark}

\begin{remark} In the field of program analysis, \textit{abstract interpretation} is a framework that can be understood as a special case of constructive causal abstraction where models are acyclic and high-level variables are aligned with individual low-level variables rather than sets of low-level variables \citep{CousotCousot77-1}. The functions $\setmap$ and $\setmap^{-1}$ are the abstraction and concretization operators that form a Galois connection, and the commuting diagram summarized in Equation~\ref{eq:commute} guarantees that abstract transfer functions are consistent with concrete transfer functions.
\end{remark}

\subsubsection{Decomposing Alignments Between Models} \label{section:decomposition}

Given the importance and prevalence of this relatively simple notion of abstraction, it is worth understanding the notion from different angles. Abstraction under the alignment $\langle \cellpart, \cellfunc \rangle$ can be decomposed via the following three fundamental operations on variables. \textit{Marginalization} removes a set of variables. \textit{Variable merge} collapses a partition of variables, i.e., each partition cell becoming a single variable. \textit{Value merge} collapses a partition of values for each variable, i.e., each partition cell becoming a single value. The first and third operations relate closely to concepts identified in the philosophy of science literature as being critical to addressing the problem of \emph{variable choice} \citep{Kinney,Woodward2021}. 

First, \emph{marginalization} removes a set of variables $\vars$.
As an alignment, the variables $\vars$ are placed into the cell $\cellpart_\bot$, while each other variable $\onevarr \notin \vars$ is left untouched; in a model that is an abstraction under this alignment, the parents and children of each marginalized variable are directly linked.
\begin{definition}[Marginalization]
Define the marginalization of $\vars \subset \allvars$ to be an alignment from the signature $\signature = (\allvars, \values{})$ to the high-level signature $\signature^* = (\allvars \setminus \vars, \values{})$: We set the partitions $\cellpart_{\onevarr} = \{\onevarr\}$ for $\onevarr \in \allvars \setminus \vars$ and $\cellpart_\bot = \vars$, while the functions are identity, $\cellfunc_\onevarr = \onevall \mapsto \onevall$ for $\onevarr \in \allvars \setminus \vars$. \label{def:marg}
\end{definition}

Marginalization is essentially a matter of \emph{ignoring} a subset $\vars$ of variables. 

Next, \emph{variable merge} collapses each cell of a partition into single variables. Variables are merged according to a partition $\{\cellpart_{\onevar^*}\}_{\onevar^* \in \allvars^*}$ with cells indexed by \textit{new} variables $\allvars^*$. In a model that is an abstraction under a variable merge, these new variables depend on the parents of their partition and determine the children of their partition.

\begin{definition}[Variable Merge] \label{def:variable-merge} 
Let $\{\cellpart_{\onevar^*}\}_{\onevar^* \in \allvars^*}$ be a partition of $\allvars$ indexed by new high-level variables $\allvars^*$ where $\values{\onevar^*}^* = \values{\cellpart_{\onevar^*}}$ for each $\onevar^* \in \allvars^*$.
Then the variable merge of $\allvars$ into $\allvars^*$ is an alignment to the new signature $\signature^* = (\allvars^*, \values{}^*)$ with partition $\cellpart = \{\cellpart_{\onevar^*}\}_{\onevar^* \in \allvars^*} \cup \{\cellpart_\bot\}$ where $\cellpart_\bot = \varnothing$ and functions $\cellfunc_{\onevar^*}(\psett) = \psett$ for each $\onevar^* \in \allvars^*$.
\end{definition}

Finally, \emph{value merge} alters the value space of each variable, potentially collapsing values:

\begin{definition}[Value Merge]\label{def:value-merge}
Choose some family $\delta = \{\delta_\onevar\}_{\onevar \in \allvars}$ of partial surjective functions $\delta_\onevar : \values{\onevar} \to \values{\onevar}^*$ mapping to new variable values. 
The value merge is an alignment to the new signature $\signature^* = (\allvars, \values{ }^*)$ 
with partition cells $\cellpart_\onevar = \{\onevar\}$ for $\onevar \in \allvars$, $\cellpart_\bot = \varnothing$, and functions $\cellfunc_\onevar = \delta_\onevar$ for $\onevar \in \allvars$.
\end{definition}

The notion of value merge  relates to an important concept in the philosophy of causation. The range of values $\values{\onevar}$ for a variable $\onevar$ can be more or less coarse-grained, and some levels of resolution seem to be better causal-explanatory targets. For instance, to use a famous example from \cite{Yablo}, if a bird is trained to peck any target that is a shade of red, then it would be misleading, if not incorrect, to say that appearance of crimson (a particular shade of red) causes the bird to peck. Roughly, the reason is that this suggests the wrong counterfactual contrasts: if the target were not crimson (but instead another shade of red, say, scarlet), the bird would still peck. Thus, for a given explanatory purpose, the level of grain in a model should guarantee that cited causes can be \emph{proportional} to their effects \citep{Yablo,Woodward2021}.

The three operations above are notable not only for conceptual reasons, but also because they suffice to decompose any alignment, as we will now explain. Composition of alignments is defined, as expected, via composition of their maps. Formally:
\begin{definition}
Let $\langle \cellpart, \cellfunc\rangle$ be an alignment from signature $(\allvars, \values{})$ to signature $(\allvars', \values{}')$
and $\langle \cellpart', \cellfunc'\rangle$ be an alignment from signature $(\allvars', \values{}')$ to signature $(\allvars'', \values{}'')$.
We define the \emph{composition} $\langle \Pi', \pi' \rangle \circ \langle \Pi, \pi\rangle$ as an alignment $\langle \Pi'',  \pi''\rangle$ from signature $(\allvars, \values{})$ to signature $(\allvars'', \values{}'')$ whose cells $\Pi''$ and maps $\pi''$ are given as follows:
\begin{equation*}
\Pi''_{X''} = \bigcup_{\substack{X' \in \Pi'_{X''}}} \Pi_{X'} , \quad
\Pi''_{\bot} = \bigcup_{\substack{X' \in \Pi'_{\bot} \cup \{\bot\}}} \Pi_{X'} , \quad
\pi''_{X''}(\mathbf{y}) = \pi'_{X''}\Big[ \bigcup_{X' \in \Pi'_{X''}} \pi_{X'}\big(\project{\mathbf{y}}{\Pi_{X'}}\big)\Big]
\end{equation*}
\end{definition}
We then have the following:
\begin{proposition}\label{prop:decompose}
\label{prop:transdecomp}
Let $\langle \cellpart, \cellfunc \rangle$ be an alignment. Then there is a marginalization, variable merge, and value merge whose composition is $\langle \cellpart, \cellfunc \rangle$.
\end{proposition}
\begin{proof}
It is straightforward to verify that the composition of alignments corresponding to the following three operations gives $\langle \cellpart, \cellfunc \rangle$.
First, marginalize the variables $\vars = \cellpart_\bot$. Then, variable merge with $\allvars^*$ as the index set for the partition $\cellpart$. Lastly, value merge with $\delta_\onevar = \cellfunc_\onevar$ for each $\onevar \in \allvars^*$.
\end{proof}

 Here, we give an example of an abstraction under a composition of marginalization, variable merge, and value merges. At each step we give a model that is a constructive abstraction of the previous model under the corresponding operation, with the final result being a high-level model that is a constructive abstraction of the initial model.

For succinctness, we will henceforth use marginalization, variable merge, and value merge as operations on models themselves. Thus, rather than saying, e.g., ``a constructive abstraction of the model under a value merge alignment,'' we simply say ``a value merged model,'' and so on.

\begin{example}
Consider a causal model $\model$ that computes the maximum of two positive numbers. 

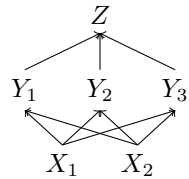
\centerline{
\begin{tikzpicture}
\tikzstyle{nnline}=[->]
\tikzstyle{intline}=[->, very thick, dashed]

\node[] (x1) at (0, 0) {$\onevar_1$} ;
\node[] (x2) at (1, 0) {$\onevar_2$} ;
\node[] (h1) at (-0.5, 1) {$\onevarr_{1}$} ;
\node[] (h2) at (0.5, 1) {$\onevarr_{2}$} ;
\node[] (h3) at (1.5, 1) {$\onevarr_{3}$} ;
\node[] (y) at (0.5, 2) {$\onevarrr$} ;

\draw[nnline] (x1.north) -- (h1.south);
\draw[nnline] (x1.north) -- (h2.south);
\draw[nnline] (x1.north) -- (h3.south);

\draw[nnline] (x2.north) -- (h1.south);
\draw[nnline] (x2.north) -- (h2.south);
\draw[nnline] (x2.north) -- (h3.south);

\draw[nnline] (h1.north) -- (y.south);
\draw[nnline] (h2.north) -- (y.south);
\draw[nnline] (h3.north) -- (y.south);
\end{tikzpicture}
}

The variables $\onevar_1$ and $\onevar_2$ take on positive real values and have constant functions mapping to $1$. The variables $\onevarr_1, \onevarr_2, \onevarr_3$ take on real-valued numbers and have the following mechanisms: \[
\begin{bmatrix}
    \mechanism{\onevarr_1}(\oneval_1, \oneval_2) \\
    \mechanism{\onevarr_2}(\oneval_1, \oneval_2) \\
    \mechanism{\onevarr_3}(\oneval_1, \oneval_2)
\end{bmatrix} = \text{ReLU}\left( \begin{bmatrix}
    \oneval_1 \\
    \oneval_2
\end{bmatrix}^\top \begin{bmatrix}
    \phantom{-}1 & -1 & 1 \\
    -1 & \phantom{-}1 & 1
\end{bmatrix} \right)
\]

The variable $\onevarrr$ takes on a real-number value and has the mechanism $\mechanism{\onevarrr}(\onevall_1, \onevall_2, \onevall_3) = \frac{1}{2} (\onevall_1 + \onevall_2 + \onevall_3)$. Observe that $\project{\sols(\model_{\oneval_1 \circ \oneval_2})}{\onevarrr} = \mathsf{max}(\oneval_1, \oneval_2)$; only one of $\onevarr_1$ and $\onevarr_2$ takes on a positive value, depending on whether $\onevar_1$ or  $\onevar_2$ is greater. 

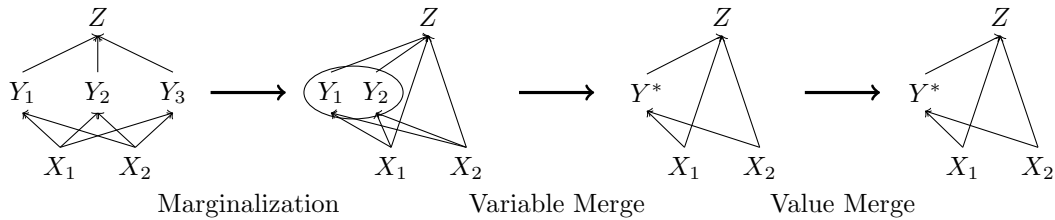
\centerline{
\begin{tikzpicture}
\tikzstyle{nnline}=[->]
\tikzstyle{intline}=[->, very thick, dashed]
\def\x{-3.5}
\node[] (x1) at (0  + \x, 0) {$\onevar_1$} ;
\node[] (x2) at (1 + \x, 0) {$\onevar_2$} ;
\node[] (h1) at (-0.5  + \x, 1) {$\onevarr_{1}$} ;
\node[] (h2) at (0.5  + \x, 1) {$\onevarr_{2}$} ;
\node[] (h3) at (1.5  + \x, 1) {$\onevarr_{3}$} ;
\node[] (y) at (0.5  + \x, 2) {$\onevarrr$} ;

\draw[nnline] (x1.north) -- (h1.south);
\draw[nnline] (x1.north) -- (h2.south);
\draw[nnline] (x1.north) -- (h3.south);

\draw[nnline] (x2.north) -- (h1.south);
\draw[nnline] (x2.north) -- (h2.south);
\draw[nnline] (x2.north) -- (h3.south);

\draw[nnline] (h1.north) -- (y.south);
\draw[nnline] (h2.north) -- (y.south);
\draw[nnline] (h3.north) -- (y.south);

 \draw[->, very thick] (-1.5, 1)--(-0.5, 1);
\node[] (label) at (-1, -0.5) {Marginalization};

 \def\x{0.9}

\node[] (x1) at (0 + \x, 0) {$\onevar_1$} ;
\node[] (x2) at (1 + \x, 0) {$\onevar_2$} ;
\node[] (h1) at (-0.8 + \x, 1) {$\onevarr_{1}$} ;
\node[] (h2) at (-0.2 + \x, 1) {$\onevarr_{2}$} ;
\node[] (y) at (0.5 + \x, 2) {$\onevarrr$} ;
\node[draw, ellipse, minimum width=37.5pt, minimum height=20pt] at (-0.5 + \x, 1) {};

\draw[nnline] (x1.north) -- (h1.south);
\draw[nnline] (x1.north) -- (h2.south);
\draw[nnline] (x1.north) -- (y.south);

\draw[nnline] (x2.north) -- (h1.south);
\draw[nnline] (x2.north) -- (h2.south);
\draw[nnline] (x2.north) -- (y.south);

\draw[nnline] (h1.north) -- (y.south);
\draw[nnline] (h2.north) -- (y.south);

 \def\x{1.8}

\draw[->, very thick] (2.6, 1)--(3.6, 1);
\node[] (label) at (3.1, -0.5) {Variable Merge};
\node[] (x1) at (3 + \x, 0) {$\onevar_1$} ;
\node[] (x2) at (4 + \x , 0) {$\onevar_2$} ;
\node[] (h1) at (2.5 + \x, 1) {$\onevarr^*$} ;
\node[] (y) at (3.5 + \x, 2) {$\onevarrr$} ;

\draw[nnline] (x1.north) -- (h1.south);
\draw[nnline] (x1.north) -- (y.south);

\draw[nnline] (x2.north) -- (h1.south);
\draw[nnline] (x2.north) -- (y.south);

\draw[nnline] (h1.north) -- (y.south);

\def\x{2.5}
 
 \draw[->, very thick] (6.4, 1)--(7.4, 1);
\node[] (label) at (6.9, -0.5) {Value Merge};

\node[] (x1) at (6 + \x, 0) {$\onevar_1$} ;
\node[] (x2) at (7 + \x, 0) {$\onevar_2$} ;
\node[] (h) at (5.5 + \x, 1) {$\onevarr^*$} ;
\node[] (y) at (6.5 + \x, 2) {$\onevarrr$} ;

\draw[nnline] (x1.north) -- (h.south);
\draw[nnline] (x1.north) -- (y.south);

\draw[nnline] (x2.north) -- (h.south);
\draw[nnline] (x2.north) -- (y.south);

\draw[nnline] (h.north) -- (y.south);
\end{tikzpicture}
}

The marginalization of $\model$ with $\cellpart_\bot = \{\onevarr_3\}$ removes the mechanism of $\onevarr_3$ and changes the mechanism of $\onevarrr$ to $\mechanism{\onevarrr}(\oneval_1, \oneval_2,\onevall_1, \onevall_2) = \frac{1}{2}(\onevall_1 + \onevall_2 + \oneval_1 + \oneval_2)$.

The variable merge of the marginalized model with $\cellpart_{\onevarr^*} = \{\onevarr_1, \onevarr_2\}$ constructs a new variable $\onevarr^*$ with mechanism 
\[\mechanism{\onevarr^*}(\oneval_1, \oneval_2) = \begin{bmatrix}
    \mechanism{\onevarr_1}(\oneval_1, \oneval_2) \\
    \mechanism{\onevarr_2}(\oneval_1, \oneval_2)
\end{bmatrix} = \text{ReLU} \left( \begin{bmatrix}
    \oneval_1 - \oneval_2 \\
    \oneval_2 - \oneval_1
\end{bmatrix} \right)\]
and alters the mechanism of $\onevarrr$ to $\mechanism{\onevarrr}( \oneval_1, \oneval_2,(\onevall^*_1, \onevall^*_2)) = \frac{1}{2}(\onevall^*_1 + \onevall^*_2 + \oneval_1 + \oneval_2)$.

Finally, we value merge the marginalized and variable merged model. Define $\delta_{\onevarr^*} : \mathbb{R}^2 \to \{0, 1\}$ where $\delta_{\onevarr^*}(\onevall_1, \onevall_2) = \indicator{\onevall_1 \geq 0}$, which creates a binary partition over the values of $\onevarr$. In the case that $\onevall_1 =\oneval_1 - \oneval_2 $, then $\delta_{\onevarr^*}(h) = 1$, meaning that $\oneval_1 \geq \oneval_2$; in the other possible case where $\onevall = 0$ and $\oneval_1 < \oneval_2$, then $\delta_{\onevarr^*}(h) = 0$. After value merge with $\delta$, the mechanism of $\onevarr^*$ is $\mechanism{\onevarr^*}(\oneval_1, \oneval_2) = \indicator{\oneval_1 \geq \oneval_2}$ and the mechanism of $\onevarrr$ is $\mechanism{\onevarrr}( \oneval_1, \oneval_2,\onevall^*) = \onevall^* \cdot \oneval_1 + (1 - \onevall^*)\oneval_2$.
\end{example}

\subsection{Approximate Transformation} \label{section:approximate}
Constructive causal abstraction and other exact transformations are all-or-nothing notions; the exact transformation relation either holds or it doesn't. This binary concept prevents us from having a graded notion of faithful interpretation that is more useful in practice. We define a notion of approximate abstraction \citep{beckers20a, Rischel, Zennaro23b} that can be flexibly adapted:
\begin{definition}[Approximate Transformation] \label{def:approx} Consider causal models $\model, \model^*$ and intervention algebras $(\intals, \circ), (\intals^*, \circ)$ with signatures $\signature$ and $\signature^*$, respectively. Furthermore, let  $\setmap: \values{\allvars} \to \values{\allvars^*}$ and $\intmap$ be surjective partial functions where $\omega: \intals \to \intals^*$ is $\leq$ order preserving. Finally, we also need:
\begin{enumerate}
    \item a (`distance') function $\simfunc$ that maps two total settings of $\model^*$ to a real number,
    \item a probability distribution $\distribution$ over $\dom(\intmap)$ used to describe which interventionals are expected,
    \item a real-valued statistic $\statistic$ for the random variable $\simfunc\big(\setmap(\sols(\model_{\intal}) ), \sols(\model^*_{\intmap(\intal)} )\big)$ where $ \intal \sim \distribution$.
\end{enumerate}
Taken together, we can construct a metric that quantifies the degree to which $\model^*$ is an \textit{approximate transformation} of $\model$ in a single number:

\begin{equation}
 \statistic_{\intal\sim \distribution}[\simfunc\big( \setmap(\sols(\model_{\intal}) ), \sols(\model^*_{\intmap(\intal)} )\big) ]\label{eq:approxtrans}
\end{equation}

This metric is a graded version of Equation~\ref{eq:commute} in the definition of exact transformation. If this number is above a particular cutoff $\eta$, we can say that $\model^*$ is an $\eta$-approximate abstraction of $\model$.
\end{definition}

\begin{remark}
When $\model$ is a model with inputs and outputs, we might consider probability distributions that only give mass to interventionals that assign all input variables. This ensures that the default value for input variables is not taken into account (See Remark~\ref{rem:acyc}).
\end{remark}

\begin{remark}
In a paper on approximate causal abstraction, \cite{beckers20a} propose $d_{\text{max}}$-$\alpha$-approximate abstraction, which takes the maximum distance over low-level interventions.
\end{remark}

\begin{example}
Let $\mathcal{M}$ be a causal model that computes the sum of two numbers with addend variables $\onevar_1$ and $\onevar_2$ that take on values $\{0, 1, 2, \dots\}$ and a sum variable $\onevarr$ that takes on values $\{0, 1, 2, \dots\}$. Let $\mathcal{M}^*$ be a causal model that only sums multiples of 10 with addend variables $\onevar_1^*$ and $\onevar_2^*$ that take on values $\{0, 10, 20, \dots\}$ and a sum variable $\onevarr^*$ that takes on values $\{0, 10, 20, \dots\}$. 

We can quantify the degree to which $\model^*$ approximates $\model$ where $\distribution$ is uniform over hard interventions targeting exactly $\onevar_1$ and $\onevar_2$, $\simfunc$ outputs the absolute difference between the values of $\onevarr^*$, and $\statistic$ is expected value. Let $(\intals, \circ)$ and $(\intals^*, \circ)$ be hard interventions and define maps $\setmap(\oneval_1, \oneval_2, \onevall) = \{\oneval_1 \% 10, \oneval_2 \% 10, \onevall \% 10\}$ and $\intmap(\psettt) = \{\project{\psettt}{\onevarrr} \% 10: \onevarrr \in \varrrs\}$. 

The degree to which $\model^*$ is an approximate transformation of $\model$ is 
\begin{align*}
\statistic_{\{\oneval_1, \oneval_2\}\sim \distribution}[\simfunc\big( \setmap(\sols(\model_{\{\oneval_1, \oneval_2\}}) ), \sols(\model^*_{\intmap(\{\oneval_1, \oneval_2\})} )\big) ] =\\
\statistic_{\{\oneval_1, \oneval_2\}\sim \distribution}[\simfunc\big( (\oneval_1 \% 10, \oneval_2 \% 10, (\oneval_1 + \oneval_2) \% 10), (\oneval_1 \% 10, \oneval_2 \% 10, \oneval_1 \% 10 + \oneval_2 \% 10)\big)] = \\
\mathbb{E}_{\{\oneval_1, \oneval_2\}\sim \distribution}[|(\oneval_1 + \oneval_2) \% 10) - (\oneval_1 \% 10 + \oneval_2 \% 10)|] =4.5
\end{align*}
    
\end{example}

\subsection{Interchange Interventions}\label{sec:interchange}
An \textit{interchange intervention} \citep{geiger-etal:2020:blackbox, geiger2021causal} is an operation on a causal model with input and output variables (e.g., acyclic models; recall Remark \ref{rem:acyc}).  Specifically, the causal model is provided a `base' input and an intervention is performed that fixes some variables to be the values they would have if different `source' inputs were provided. Such interventions will be central to grounding mechanistic interpretability in causal abstraction. 

\begin{definition}[Interchange Interventions]\label{def:interchange}
Let $\model$ be a causal model with input variables $\inputvars \subseteq \allvars$. Furthermore, consider source inputs $\source_1, \dots, \source_k \in \values{\inputvars}$ and disjoint sets of target variables $\vars_1, \dots, \vars_k \subseteq \allvars \setminus \inputvars$. Define the hard intervention
\[\intinv(\model,\langle \source_1, \dots ,\source_k \rangle, \langle \vars_1, \dots, \vars_k\rangle) \overset{\textnormal{def}}{=}
\bigcup_{1\leq j \leq k} \project{\sols(\model_{\source_j})}{\vars_j}.\]
\end{definition}
We take interchange interventions as a base case and generalize to \textit{recursive interchange interventions}, where variables are fixed to be the value they would have if a recursive interchange intervention (that itself may be defined in terms of other interchange interventions) were performed.
\begin{definition}[Recursive Interchange Interventions]\label{def:recinterchange}
Let $\model$ be a causal model with input variables $\inputvars \subseteq \allvars$. Define recursive interchange interventions of depth $0$ to simply be interchange interventions. Given $\source_1, \dots, \source_k \in \values{\inputvars}$, disjoint sets of target variables $\vars_1, \dots, \vars_k \subseteq \allvars^* \setminus \inputvars$, and interchange interventions $\intpset_1, \dots, \intpset_k$ of depth $m$, we define the recursive interchange interventions of depth $m+1$ to be
\[\rintinv^{m+1}(\model,\langle \source_1, \dots ,\source_k \rangle, \langle \vars_1, \dots, \vars_k\rangle, \langle\intpset_{1}, \dots, \intpset_{k} \rangle) \overset{\textnormal{def}}{=}
\bigcup_{1\leq j \leq k} \project{\sols(\model_{\source_j \circ \intpset_j})}{\vars_j}.\]
\end{definition}

\cite{Geiger-etal:2023:DAS} generalize interchange interventions to \textit{distributed interchange interventions} that target variables distributed across multiple causal variables. We define this operation using an interventional that applies a bijective translation, performs an interchange intervention in the new variable space, and then applies the inverse translation to get back to the original variable space. 
\begin{definition}[Distributed Interchange Interventions]\label{def:distinterchange}
Let $\model$ be a causal model with input variables $\inputvars \subseteq \allvars$ and let $\setmap:\values{\allvars} \to \values{\allvars^*}$ be a bijective translation that preserves inputs.\footnote{I.e., $\inputvars \subseteq \allvars^*$ and for all $\pset \in \values{\inputvars}$ and $\psett \in \values{\allvars \setminus \inputvars}$ there exists a $\psett^* \in \values{\allvars^* \setminus \inputvars}$ such that $\setmap(\pset \cup \psett ) = \pset \cup \psett^*$.} For source inputs $\source_1, \dots, \source_k \in \values{\inputvars}$, and disjoint target variables $\vars^*_1, \dots, \vars^*_k \subseteq \allvars^* \setminus \inputvars$, we define
\[ \dintinv(\model, \setmap, \langle \source_1, \dots ,\source_k \rangle, \langle \vars^*_1, \dots, \vars^*_k\rangle)\]
to be an interventional on all variables $\allvars$ that replaces each mechanism for $\onevar$ with the function
\[\tset \mapsto \Project{\tau^{-1} \Big( \sols\big(\tau(\model)_{\intinv(\model,\langle \source_1, \dots ,\source_k \rangle, \langle \vars^*_1, \dots, \vars^*_k\rangle)}\big) \Big)}{\onevar} .\]
\end{definition}

When conducting a causal abstraction analysis using interchange interventions, a partition $\{\cellpart_{\onevar}\}_{\onevar \in \allvars_{\highmodel} \cup \{\bot\}}$ defined over all high-level variables and mapping $\{\cellfunc_{\onevar}\}_{\onevar \in \inputvars_{\highmodel}}$ defined over high-level input variables are together sufficient to fully determine an alignment $\langle \Pi, \pi \rangle$.

\begin{remark}[Constructing an Alignment for Interchange Intervention Analysis]\label{def:interchangealign}
Consider  low-level causal model $\lowmodel$ and high-level causal model $\highmodel$ with input variables $\inputvars_{\lowmodel} \subseteq \allvars_{\lowmodel}$ and $\inputvars_{\highmodel} \subseteq \allvars_{\highmodel}$, respectively. Suppose we have a partially constructed alignment $(\{\cellpart_\onevar\}_{\onevar \in \allvars_{\highmodel} \cup \bot}, \{\cellfunc_\onevar\}_{\onevar \in \inputvars_{\highmodel}})$ with
\[
\onevar \in \inputvars_{\highmodel} \Leftrightarrow \cellpart_\onevar \subseteq \inputvars_{\lowmodel} \;\;\;\;\;\;\;\;\;\;\;
  \onevar \in \allvars_{\highmodel} \setminus \inputvars_{\highmodel} \Leftrightarrow \cellpart_\onevar \subseteq \allvars_{\lowmodel} \setminus  \inputvars_{\lowmodel} \]
We can induce the remaining alignment functions from the partitions and the input alignment functions, and the two causal models. For $\onevarr_{\highmodel} \in \allvars_{\highmodel} \setminus \inputvars_{\highmodel}$ and $\psettt_{\lowmodel} \in \values{\cellpart_{\onevarr_{\highmodel}}}$, if there exists $\pset_{\lowmodel} \in \values{\inputvars_{\lowmodel}}$ such that $\psettt_{\lowmodel} = \project{\sols(\lowmodel_{\pset_{\lowmodel}})}{\cellpart_{\onevarr_{\highmodel}}}$,
 then, with $\pset_{\highmodel} = \cellfunc_{\inputvars_{\highmodel}}(\pset_{\lowmodel})$, we define the alignment functions  
\[\cellfunc_{\onevarr_{\highmodel}}(\psettt_{\lowmodel}) =
\project{\sols(\highmodel_{\pset_{\highmodel}})}{\onevarr_{\highmodel}}
\]
and otherwise, we leave $\cellfunc_{\onevarr_{\highmodel}}$ undefined for $\psettt_{\lowmodel}$.
In words, map the low-level partial settings realized for a given input to the corresponding high-level values realized for the same input.

This is a subtle point, but, in general, this construction is not well defined because two different inputs can produce the same $\psettt_{\lowmodel}$ while producing different $\project{\sols(\highmodel_{\pset_{\highmodel}})}{\onevarr_{\highmodel}}$. If this were to happen, the causal abstraction relationship simply wouldn't hold for the alignment.
\end{remark}

Once an alignment $\langle \cellpart, \cellfunc \rangle$ is constructed, aligned interventions must be performed to experimentally verify the alignment is a witness to the high-level model being an abstraction of the low-level model.  Observe that $\cellfunc$ will only be defined for values of intermediate partition cells that are realized when some input is provided to the low-level model. This greatly constrains the space of low-level interventions to intermediate partitions that will correspond with high-level interventions. Specifically, we are only able to interpret low-level \textit{interchange interventions} as high-level interchange interventions. 

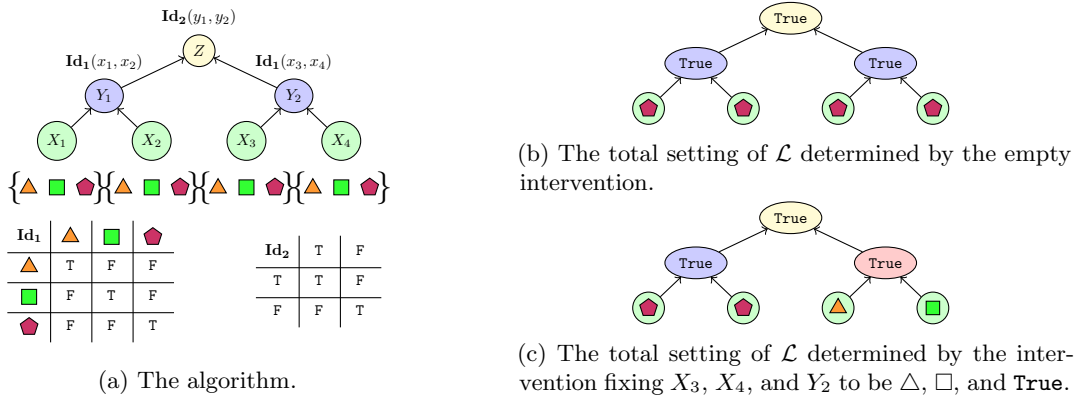
\begin{figure}[tp]
    \centering
    \begin{minipage}{0.48\textwidth}
\begin{subfigure}{\textwidth}
    \centering
    \begin{tikzpicture}[scale=0.6, every node/.style={scale=0.6}]
\def\scalex{2.1}
\def\scaley{2}
\def\startx{0}
\def\starty{-0.43}
\def\opac{1.0}

\foreach \xshift in {0,1,2,3}{
    \node[] () at (\xshift *\scalex + -0.45*\scalex+ \startx,-0.309*\scaley+ \starty) {\huge \{};
    \node[] () at (\xshift *\scalex + 0.45*\scalex+ \startx,-0.309*\scaley+ \starty) {\huge \}};
    \node[fill=orange!80, draw,regular polygon, regular polygon sides=3, minimum size=12pt,inner sep=0pt] () at (\xshift *\scalex + -0.3*\scalex+ \startx,-0.309*\scaley+ \starty) {};
    \node[fill=green!80, draw,regular polygon, regular polygon sides=4, minimum size=10pt] () at (\xshift *\scalex + 0*\scalex+ \startx,-0.3*\scaley+ \starty) {};
    \node[fill=purple!80, draw,regular polygon, regular polygon sides=5, minimum size=10pt] () at (\xshift *\scalex + 0.3*\scalex++ \startx,-0.3*\scaley+ \starty) {};
}

\def\starty{0}

\node[fill=green!20, draw, circle, minimum size=20pt] (A) at (0+ \startx,0+ \starty) {$\onevar_1$};

\node[fill=green!20, draw, circle, minimum size=20pt] (B) at (1*\scalex+ \startx,0+ \starty) {$\onevar_2$};

\node[fill=green!20, draw, circle, minimum size=20pt] (C) at (2*\scalex+ \startx,0+ \starty) {$\onevar_3$};

\node[fill=green!20, draw, circle, minimum size=20pt] (D) at (3*\scalex+ \startx,0+ \starty) {$\onevar_4$};

\node[fill=blue!20, draw, ellipse, minimum size=20pt] (E) at (0.5*\scalex+ \startx,0.5*\scaley+ \starty) {$\onevarr_1$};

\node[fill=blue!20, draw, ellipse, minimum size=20pt] (F) at (2.5*\scalex + \startx,0.5*\scaley+ \starty) {$\onevarr_2$};

\node[fill=yellow!20, draw, ellipse, minimum size=20pt] (G) at (1.5*\scalex + \startx,1*\scaley+ \starty) {$\onevarrr$};

\node () at (0.5*\scalex+ \startx,0.87*\scaley+ \starty) {$\mathbf{Id_1}(\oneval_1,\oneval_2)$};

\node () at (2.5*\scalex + \startx,0.87*\scaley+ \starty) {$\mathbf{Id_1}(\oneval_3,\oneval_4)$};

\node () at (1.5*\scalex + \startx,1.37*\scaley+ \starty) {$\mathbf{Id_2}(\onevall_1,\onevall_2)$};

\draw [->,opacity=\opac] (A) -- (E);
\draw [->,opacity=\opac] (B) -- (E);
\draw [->,opacity=\opac] (C) -- (F);
\draw [->,opacity=\opac] (D) -- (F);
\draw [->,opacity=\opac] (E) -- (G);
\draw [->,opacity=\opac] (F) -- (G);
\end{tikzpicture}

\begin{tikzpicture}[scale=0.6, every node/.style={scale=0.6}]
\newcommand{\true}{\texttt{T}}
\newcommand{\false}{\texttt{F}}

\def\opac{1.0}
    \matrix (mat) [matrix of nodes,
        row sep=-\pgflinewidth,
        column sep=-\pgflinewidth,
        nodes={rectangle,text width=2em,align=center},
        text depth=1ex,
        text height=2ex,
        nodes in empty cells] at (0,0)
    { 
    $\mathbf{Id_1}$\Huge &  & &  \\
     & \true  & \false & \false \\
     & \false & \true & \false \\
    & \false & \false & \true \\
    };
    \foreach \x in {1}
    {
      \draw 
        ([xshift=-.5\pgflinewidth,opacity=\opac]mat-\x-1.south west) --   
        ([xshift=-.5\pgflinewidth,opacity=\opac]mat-\x-4.south east);
      }
    \foreach \x in {1}
    {
      \draw 
        ([yshift=.5\pgflinewidth,opacity=\opac]mat-1-\x.north east) -- 
        ([yshift=.5\pgflinewidth,opacity=\opac]mat-4-\x.south east);
    } 

    \foreach \x in {2,...,3}
    {
      \draw[opacity=\opac*0.5]
        ([xshift=-.5\pgflinewidth]mat-\x-1.south west) --   
        ([xshift=-.5\pgflinewidth]mat-\x-4.south east);
      }
    \foreach \x in {2,...,3}
    {
      \draw[opacity=\opac*0.5]
        ([yshift=.5\pgflinewidth]mat-1-\x.north east) -- 
        ([yshift=.5\pgflinewidth]mat-4-\x.south east);
    } 

    \node[fill=orange!80, draw,regular polygon, regular polygon sides=3, minimum size=12pt,inner sep=0pt] () at (mat-1-2) {};
    \node[fill=green!80, draw,regular polygon, regular polygon sides=4, minimum size=10pt] () at (mat-1-3) {};
    \node[fill=purple!80, draw,regular polygon, regular polygon sides=5, minimum size=10pt] () at (mat-1-4) {};
    
        \node[fill=orange!80, draw,regular polygon, regular polygon sides=3, minimum size=12pt,inner sep=0pt] () at (mat-2-1) {};
    \node[fill=green!80, draw,regular polygon, regular polygon sides=4, minimum size=10pt] () at (mat-3-1) {};
    \node[fill=purple!80, draw,regular polygon, regular polygon sides=5, minimum size=10pt] () at (mat-4-1) {};

\matrix (mat) [matrix of nodes,
        row sep=-\pgflinewidth,
        column sep=-\pgflinewidth,
        nodes={rectangle,text width=2em,align=center},
        text depth=1ex,
        text height=2ex,
        nodes in empty cells] at (5.05,0)
    { 
     $\mathbf{Id_2}$ \Huge & \true & \false  \\
     \true & \true  & \false\\
     \false & \false & \true \\
    };

    \foreach \x in {1}
    {
      \draw 
        ([xshift=-.5\pgflinewidth,opacity=\opac]mat-\x-1.south west) --   
        ([xshift=-.5\pgflinewidth,opacity=\opac]mat-\x-3.south east);
      }
      
    \foreach \x in {1}
    {
      \draw 
        ([yshift=.5\pgflinewidth,opacity=\opac]mat-1-\x.north east) -- 
        ([yshift=.5\pgflinewidth,opacity=\opac]mat-3-\x.south east);
    } 

    \foreach \x in {2}
    {
      \draw[opacity=\opac*0.5]
        ([xshift=-.5\pgflinewidth]mat-\x-1.south west) --   
        ([xshift=-.5\pgflinewidth]mat-\x-3.south east);
      }
    \foreach \x in {2}
    {
      \draw[opacity=\opac*0.5]
        ([yshift=.5\pgflinewidth]mat-1-\x.north east) -- 
        ([yshift=.5\pgflinewidth]mat-3-\x.south east);
    }

\renewcommand{\true}{\texttt{True}}
\renewcommand{\false}{\texttt{False}}

    \end{tikzpicture}
    \caption{The algorithm.}
    \label{fig:low}
\end{subfigure}
\end{minipage}
\hfill
    \begin{minipage}{0.48\textwidth}
\begin{subfigure}{\textwidth}
    \centering
    \begin{tikzpicture}[scale=0.6, every node/.style={scale=0.6}]
\def\scalex{2.1}
\def\scaley{2}
\def\startx{0}
\def\starty{0}
\def\opac{1.0}
\node[fill=green!20, draw, circle, minimum size=20pt] (A) at (0+ \startx,0+ \starty) {};

\node[fill=green!20, draw, circle, minimum size=20pt] (B) at (1*\scalex+ \startx,0+ \starty) {};

\node[fill=green!20, draw, circle, minimum size=20pt] (C) at (2*\scalex+ \startx,0+ \starty) {};

\node[fill=green!20, draw, circle, minimum size=20pt] (D) at (3*\scalex+ \startx,0+ \starty) {};

\node[fill=purple!80, draw,regular polygon, regular polygon sides=5, minimum size=10pt] () at (0+ \startx,0+ \starty) {};

\node[fill=purple!80, draw,regular polygon, regular polygon sides=5, minimum size=10pt] () at (1*\scalex+ \startx,0+ \starty) {};

\node[fill=purple!80, draw,regular polygon, regular polygon sides=5, minimum size=10pt] () at (2*\scalex+ \startx,0+ \starty) {};

\node[fill=purple!80, draw,regular polygon, regular polygon sides=5, minimum size=10pt] () at (3*\scalex+ \startx,0+ \starty) {};

\node[fill=blue!20, draw, ellipse, minimum size=20pt] (E) at (0.5*\scalex+ \startx,0.5*\scaley+ \starty) {\texttt{True}};

\node[fill=blue!20, draw, ellipse, minimum size=20pt] (F) at (2.5*\scalex + \startx,0.5*\scaley+ \starty) {\texttt{True}};

\node[fill=yellow!20, draw, ellipse, minimum size=20pt] (G) at (1.5*\scalex + \startx,1*\scaley+ \starty) {\texttt{True}};

\draw [->,opacity=\opac] (A) -- (E);
\draw [->,opacity=\opac] (B) -- (E);
\draw [->,opacity=\opac] (C) -- (F);
\draw [->,opacity=\opac] (D) -- (F);
\draw [->,opacity=\opac] (E) -- (G);
\draw [->,opacity=\opac] (F) -- (G);
\end{tikzpicture}
    \caption{The total setting of $\lowmodel$ determined by the empty intervention.}
    \label{fig:lownointervention}
\end{subfigure}

\begin{subfigure}{\textwidth}
    \centering
    \begin{tikzpicture}[scale=0.6, every node/.style={scale=0.6}]
\def\scalex{2.1}
\def\scaley{2}
\def\startx{0}
\def\starty{0}
\def\opac{1.0}
\node[fill=green!20, draw, circle, minimum size=20pt] (A) at (0+ \startx,0+ \starty) {};

\node[fill=green!20, draw, circle, minimum size=20pt] (B) at (1*\scalex+ \startx,0+ \starty) {};

\node[fill=green!20, draw, circle, minimum size=20pt] (C) at (2*\scalex+ \startx,0+ \starty) {};

\node[fill=green!20, draw, circle, minimum size=20pt] (D) at (3*\scalex+ \startx,0+ \starty) {};

\node[fill=purple!80, draw,regular polygon, regular polygon sides=5, minimum size=10pt] () at (0+ \startx,0+ \starty) {};

\node[fill=purple!80, draw,regular polygon, regular polygon sides=5, minimum size=10pt] () at (1*\scalex+ \startx,0+ \starty) {};

\node[fill=orange!80, draw,regular polygon, regular polygon sides=3, minimum size=12pt,inner sep=0pt] () at (2*\scalex+ \startx,0+ \starty) {};

\node[fill=green!80, draw,regular polygon, regular polygon sides=4, minimum size=12pt,inner sep=0pt] () at (3*\scalex+ \startx,0+ \starty) {};

\node[fill=blue!20, draw, ellipse, minimum size=20pt] (E) at (0.5*\scalex+ \startx,0.5*\scaley+ \starty) {\texttt{True}};

\node[fill=red!20, draw, ellipse, minimum size=20pt] (F) at (2.5*\scalex + \startx,0.5*\scaley+ \starty) {\texttt{True}};

\node[fill=yellow!20, draw, ellipse, minimum size=20pt] (G) at (1.5*\scalex + \startx,1*\scaley+ \starty) {\texttt{True}};

\draw [->,opacity=\opac] (A) -- (E);
\draw [->,opacity=\opac] (B) -- (E);
\draw [->,opacity=\opac] (C) -- (F);
\draw [->,opacity=\opac] (D) -- (F);
\draw [->,opacity=\opac] (E) -- (G);
\draw [->,opacity=\opac] (F) -- (G);
\end{tikzpicture}
    \caption{The total setting of $\lowmodel$ determined by the intervention fixing $\onevar_3$, $\onevar_4$, and $\onevarr_2$ to be $\bigtriangleup$, $\square$, and $\texttt{True}$.}
    \label{fig:lowintervention}
\end{subfigure}
\end{minipage}
\caption{A tree-structured algorithm that perfectly solves the hierarchical equality task with a compositional solution.}
\end{figure}

\cite{geiger-etal-2021-iit} propose \textit{interchange intervention accuracy}, which is simply the proportion of interchange interventions where the low-level and high-level causal models have the same input--output behavior (see Section~\ref{sec:example} for an example).
\begin{definition}[Interchange Intervention Accuracy]\label{def:IIA}
Consider a low-level causal model $\lowmodel$ aligned to a high-level causal model $\highmodel$, an alignment $(\cellpart, \cellfunc)$ as defined in Definition~\ref{def:interchangealign}, and a probability distribution $\distribution$ over the domain of $\intmap^{\cellfunc}$. We define the interchange intervention accuracy as follows:
 \[\mathsf{IIA}(\highmodel, \lowmodel, (\cellpart, \cellfunc)) = \mathbb{E}_{\mathbf{i} \sim \distribution(\dom(\intmap^{\cellfunc}))}\Big[  \mathbbm{1}[\project{\setmap^{\cellfunc}(\lowmodel_{\mathbf{i}})}{\outputvars_{\highmodel}} = \project{\highmodel_{\intmap^{\cellfunc}(\mathbf{i})}}{ \outputvars_{\highmodel}}] \Big].\]
 Interchange intervention accuracy is equivalent to input-output accuracy if we further restrict $\intmap$ to be defined only on interchange interventions where base and sources are all the same single input. 
\end{definition}
This is a special case of approximate causal abstraction (Section~\ref{section:approximate}) where $\simfunc(\tset_{\lowmodel}, \tset_{\highmodel}) = \indicator{\project{\setmap^{\cellfunc}(\tset_{\lowmodel})}{\outputvars_{\highmodel}} = \project{\tset_{\highmodel}}{\outputvars_{\highmodel}}}$ and $\statistic$ is expected value.
This analysis can be extended to the case of distributed interchange interventions by simply applying a bijective translation to the low-level model before constructing the alignment to the high-level model.

\subsection{Example: Causal Abstraction in Mechanistic Interpretability}\label{sec:example}
With the theory laid out, we can now present an example of causal abstraction from the field of mechanistic interpretability. We begin by defining two basic examples of causal models that demonstrate a potential to model a diverse array of computational processes; the first causal model represents a tree-structured algorithm and the second is fully-connected feed-forward neural network. Both the network and the algorithm solve the same `hierarchical equality' task. 

A basic equality task is to determine whether a pair of objects is identical. A hierarchical equality task is to determine whether a pair of pairs of objects have identical relations. The input to the hierarchical task is two pairs of objects, and the output is $\mathsf{True}$ if both pairs are equal or both pairs are unequal, and $\mathsf{False}$ otherwise. For illustrative purposes, we define the domain of objects to consist of a triangle, square, and pentagon. For example, the input $(\pentagon, \pentagon, \bigtriangleup, \square)$ is assigned the output $\mathsf{False}$ and the inputs $(\pentagon, \pentagon, \bigtriangleup, \bigtriangleup)$ and $(\pentagon, \square, \bigtriangleup, \square)$ are both labeled $\mathsf{True}$.
 
 We chose hierarchical equality for two reasons. First, there is an obvious tree-structured symbolic algorithm that solves the task: compute whether the first pair is equal, compute whether the second is equal, then compute whether those two outputs are equal. We will encode this algorithm as a causal model. Second, equality reasoning is ubiquitous and has served as a case study for broader questions about the representations underlying relational reasoning in biological organisms \citep{marcus:1999, alhama:2019, Geiger:Carstensen:Frank:Potts:2022}. 

 We provide a \href{https://github.com/stanfordnlp/pyvene/blob/main/tutorials/advanced_tutorials/DAS_Main_Introduction.ipynb}{\textcolor{blue}{companion jupyter notebook}} that walks through this example.

\paragraph{A Tree-Structured Algorithm for Hierarchical Equality}

We define a tree structured algorithm $\highmodel$ consisting of four `input' variables $\inputvars_{\highmodel} =$ $ \{\onevar_1,\onevar_2,\onevar_3,\onevar_4\}$ each with possible values $\values{\onevar_j} = \{\pentagon, \bigtriangleup, \square \normalsize\}$, two `intermediate' variables $\onevarr_1, \onevarr_2$ with values $\values{\onevarr_j} = \{\texttt{True}, \texttt{False}\}$, and one `output' variable $\outputvars_{\highmodel} = \{\onevarrr\}$ with values $\values{\onevarrr} = \{\texttt{True}, \texttt{False}\}$. 

The acyclic causal graph is depicted in Figure~\ref{fig:low}, where each $\mechanism{\onevar_i}$ (with no arguments) is a constant function to $\pentagon$ (which will be overwritten, per Remark~\ref{rem:acyc}), and $\mechanism{\onevarr_1},\mechanism{\onevarr_2},\mechanism{O}$ each compute equality over their respective domains, e.g., $\mechanism{\onevarr_1}(\oneval_1,\oneval_2) = \mathbbm{1}\big [\oneval_1=\oneval_2\big ]$. A total setting can be captured by a vector $[\oneval_1,\oneval_2,\oneval_3, \oneval_4,\onevall_1,\onevall_2,\onevalll]$ of values for each of the variables.

The default total setting that results from no intervention is $[\pentagon,\pentagon,\pentagon,\pentagon,\normalsize\texttt{True}, \texttt{True}, \texttt{True}]$ (see Figure~\ref{fig:lownointervention}). We can also ask what would have occurred had we intervened to fix $\onevar_3$, $\onevar_4$, and $\onevarr_1$ to be $\bigtriangleup$, $\square$, and $\texttt{False}$, for example. The result is $[\pentagon,\pentagon,\bigtriangleup,\square\normalsize, \texttt{False}, \texttt{False}, \texttt{True}]$ (See Figure~\ref{fig:lowintervention}). 

\begin{figure}
    \centering
    \resizebox{0.8\textwidth}{!}{
    \begin{tikzpicture}[thick,scale=0.6, every node/.style={scale=0.6}]
\def\scalex{3}
\def\scaley{1.95}
\def\opac{0.2}
\Large

\def\startx{0}
\def\starty{-0.5}

\foreach \xshift in {1.5,3.5,5.5,7.5}{
    \node[] () at (\xshift *\scalex + -0.45*\scalex + \startx,-0.309*\scaley+ \starty) {\huge \{};
    \node[] () at (\xshift *\scalex + 0.45*\scalex + \startx,-0.309*\scaley+ \starty) {\huge \}};
    \node[fill=orange!80, draw,regular polygon, regular polygon sides=3, minimum size=16pt,inner sep=0pt] () at (\xshift *\scalex + -0.3*\scalex+ \startx,-0.316*\scaley+ \starty) {};
    \node[fill=green!80, draw,regular polygon, regular polygon sides=4, minimum size=10pt] () at (\xshift *\scalex + 0*\scalex+ \startx,-0.3*\scaley+ \starty) {};
    \node[fill=purple!80, draw,regular polygon, regular polygon sides=5, minimum size=10pt] () at (\xshift *\scalex + 0.3*\scalex++ \startx,-0.3*\scaley+ \starty) {};
}

\node[fill=yellow!20, draw, rectangle, rounded corners=3pt, minimum width=40pt, minimum height=20pt](out) at (4 * \scalex,2.75 * \scaley) {\Large $\outneuron_{\mathsf{True}}$};
\node[fill=yellow!20, draw, rectangle, rounded corners=3pt, minimum width=40pt, minimum height=20pt](out2) at (5 * \scalex,2.75 * \scaley) {\Large $\outneuron_{\mathsf{False}}$};

\node[fill=green!20, draw, rectangle, rounded corners=3pt, minimum width=40pt, minimum height=20pt] (BERT01) at (1 * \scalex,0 * \scaley) {${\inneuron_{1}}$};
\node[fill=green!20, draw, rectangle, rounded corners=3pt, minimum width=40pt, minimum height=20pt] (BERT02) at (2 * \scalex ,0 * \scaley) {${\inneuron_{2}}$};

\node[fill=green!20, draw, rectangle, rounded corners=3pt, minimum width=40pt, minimum height=20pt] (BERT03) at (3 * \scalex,0 * \scaley) {${\inneuron_{3}}$};
\node[fill=green!20, draw, rectangle, rounded corners=3pt, minimum width=40pt, minimum height=20pt] (BERT04) at (4 * \scalex ,0 * \scaley) {${\inneuron_{4}}$};

\node[fill=green!20, draw, rectangle, rounded corners=3pt, minimum width=40pt, minimum height=20pt] (BERT05) at (5 * \scalex,0 * \scaley) {${\inneuron_{5}}$};
\node[fill=green!20, draw, rectangle, rounded corners=3pt, minimum width=40pt, minimum height=20pt] (BERT06) at (6 * \scalex ,0 * \scaley) {${\inneuron_{6}}$};

\node[fill=green!20, draw, rectangle, rounded corners=3pt, minimum width=40pt, minimum height=20pt] (BERT07) at (7 * \scalex,0 * \scaley) {${\inneuron_{7}}$};
\node[fill=green!20, draw, rectangle, rounded corners=3pt, minimum width=40pt, minimum height=20pt] (BERT08) at (8 * \scalex ,0 * \scaley) {${\inneuron_{8}}$};

\foreach \row in { 1,2}
\foreach \col in {1,2,3,4,5,6,7,8}{
\node[fill=blue!20, draw, rectangle, rounded corners=3pt, minimum width=40pt, minimum height=20pt] (BERT\row\col) at (\col * \scalex,\row * \scaley) {$\hiddenneuron_{(\row, \col)}$ };
}

\foreach \col in {1,2,3,4,5,6,7,8}{
\foreach \coll in {1,2,3,4,5,6,7,8}{
\draw[opacity=0.3, ->] (BERT0\col.north) to (BERT1\coll.south);
\draw[opacity=0.3, ->] (BERT1\col.north) to (BERT2\coll.south);
}
}

\foreach \col in {1,2,3,4,5,6,7,8}{
\draw[opacity=0.3, ->] (BERT2\col.north) to (out.south);
\draw[opacity=0.3, ->] (BERT2\col.north) to (out2.south);
}

\node[draw, ellipse,  minimum width=160, minimum height=40pt] (piI1) at ( 1.5 * \scalex+ \startx,0.25 * \scaley + \starty) {};
\node[draw, ellipse,  minimum width=160, minimum height=40pt] (piI2) at ( 3.5 * \scalex+ \startx,0.25 * \scaley + \starty) {};
\node[draw, ellipse,  minimum width=160, minimum height=40pt] (piI3) at ( 5.5 * \scalex+ \startx,0.25 * \scaley + \starty) {};
\node[draw, ellipse,  minimum width=160, minimum height=40pt] (piI4) at ( 7.5 * \scalex+ \startx,0.25 * \scaley + \starty) {};

\node[draw, ellipse,  minimum width=360, minimum height=50pt] (piV1) at ( 2.5 * \scalex+ \startx,1.25 * \scaley + \starty) {};
\node[draw, ellipse,  minimum width=360, minimum height=50pt] (piV2) at ( 6.5 * \scalex+ \startx,1.25 * \scaley + \starty) {};

\node[draw, ellipse,  minimum width=160, minimum height=50pt] (pio) at ( 4.5 * \scalex+ \startx,3 * \scaley + \starty) {};

\def\starty{7.5}
\def\startx{9}

\node[fill=green!20, draw, circle, minimum size=20pt] (A) at (0+ \startx,0+ \starty) {$\onevar_1$};

\node[fill=green!20, draw, circle, minimum size=20pt] (B) at (1*\scalex+ \startx,0+ \starty) {$\onevar_2$};

\node[fill=green!20, draw, circle, minimum size=20pt] (C) at (2*\scalex+ \startx,0+ \starty) {$\onevar_3$};

\node[fill=green!20, draw, circle, minimum size=20pt] (D) at (3*\scalex+ \startx,0+ \starty) {$\onevar_4$};

\node[fill=blue!20, draw, ellipse, minimum size=20pt] (E) at (0.5*\scalex+ \startx,0.5*\scaley+ \starty) {$\onevarr_1$};

\node[fill=blue!20, draw, ellipse, minimum size=20pt] (F) at (2.5*\scalex + \startx,0.5*\scaley+ \starty) {$\onevarr_2$};

\node[fill=yellow!20, draw, ellipse, minimum size=20pt] (G) at (1.5*\scalex + \startx,1*\scaley+ \starty) {$\onevarrr$};

\draw [->,opacity=\opac] (A) -- (E);
\draw [->,opacity=\opac] (B) -- (E);
\draw [->,opacity=\opac] (C) -- (F);
\draw [->,opacity=\opac] (D) -- (F);
\draw [->,opacity=\opac] (E) -- (G);
\draw [->,opacity=\opac] (F) -- (G);

\draw[ -, dotted] (pio.north) to (G.south);

\draw[-, dotted] (piI1.west) to[in=180, out=180] (A.west);
\draw[-, dotted] (piI2.north) to[in=-135, out=90] (B.south);
\draw[-, dotted] (piI3.north) to[in=-45, out=90] (C.south);
\draw[-, dotted] (piI4.east) to[in=0, out=0] (D.east);

\draw [-, dotted] (piV1.west) to[in=180, out=180] (E.west);
\draw [-, dotted] (piV2.east) to[in=0, out=0] (F.east);
\end{tikzpicture}
    \vspace{10pt}
    }
    \[\footnotesize
\begin{tabular}{ccc}
 $\inneuronset_{\small\pentagon} = [0.012, -0.301]$ & $\inneuronset_{\large \square} = [-0.812, 0.456]$ & $\inneuronset_{\large \bigtriangleup} = [0.682, 0.333]$\\
\end{tabular}\]
\resizebox{0.8\textwidth}{!}{
$W_1 = \begin{bmatrix*}[r]
1 & 0 & \-1 & 0 & 0 & 0 & 0 & 0\\
0 & 1 & 0 & \-1 & 0 & 0 & 0 & 0\\
\-1 & 0 & 1 & 0 & 0 & 0 & 0 & 0\\
0 & \-1 & 0 & 1 & 0 & 0 & 0 & 0\\
0 & 0 & 0 & 0 & 1 & 0 & \-1 & 0\\
0 & 0 & 0 & 0 & 0 & 1 & 0 & \-1 \\
0 & 0 & 0 & 0 & \-1 & 0 & 1 & 0 \\
0 & 0 & 0 & 0 & 0 & \-1 & 0 & 1 \\
\end{bmatrix*}$ \hfill $ W_2 = \begin{bmatrix*}[r]
1 & \-1 & 0 & 1 & 0 & 0 & 0 & 0\\
1 & \-1 & 0 & 1 & 0 & 0 & 0 & 0\\
1 & \-1 & 0 & 1 & 0 & 0 & 0 & 0\\
1 & \-1 & 0 & 1 & 0 & 0 & 0 & 0\\
\-1 & 1 & 1 & 0 & 0 & 0 & 0 & 0\\
\-1 & 1 & 1 & 0 & 0 & 0 & 0 & 0\\
\-1 & 1 & 1 & 0 & 0 & 0 & 0 & 0 \\
\-1 & 1 & 1 & 0 & 0 & 0 & 0 & 0 \\
\end{bmatrix*}$ \hfill $W_3 = \begin{bmatrix*}[r]
1 & 0\\
1 & 0\\
1-\epsilon & 0\\
1-\epsilon & 0\\
0 & 0\\
0 & 0\\
0 & 0\\
0 & 0\\
\end{bmatrix*}$
}

\[W_3 \mathsf{ReLU}(W_2\mathsf{ReLU}(W_1[\mathbf{a}, \mathbf{b}, \mathbf{c}, \mathbf{d}])) = [||\mathbf{a}-\mathbf{b}| - |\mathbf{c}-\mathbf{d}|| -(1-\epsilon)|\mathbf{a}-\mathbf{b}|-(1-\epsilon)|\mathbf{c}-\mathbf{d}|, 0]\]
    \caption{A fully-connected feed-forward neural network that labels inputs for the hierarchical equality task. The weights of the network are handcrafted to implement the tree-structured solution to the task.}
    \label{fig:hand}
\end{figure}
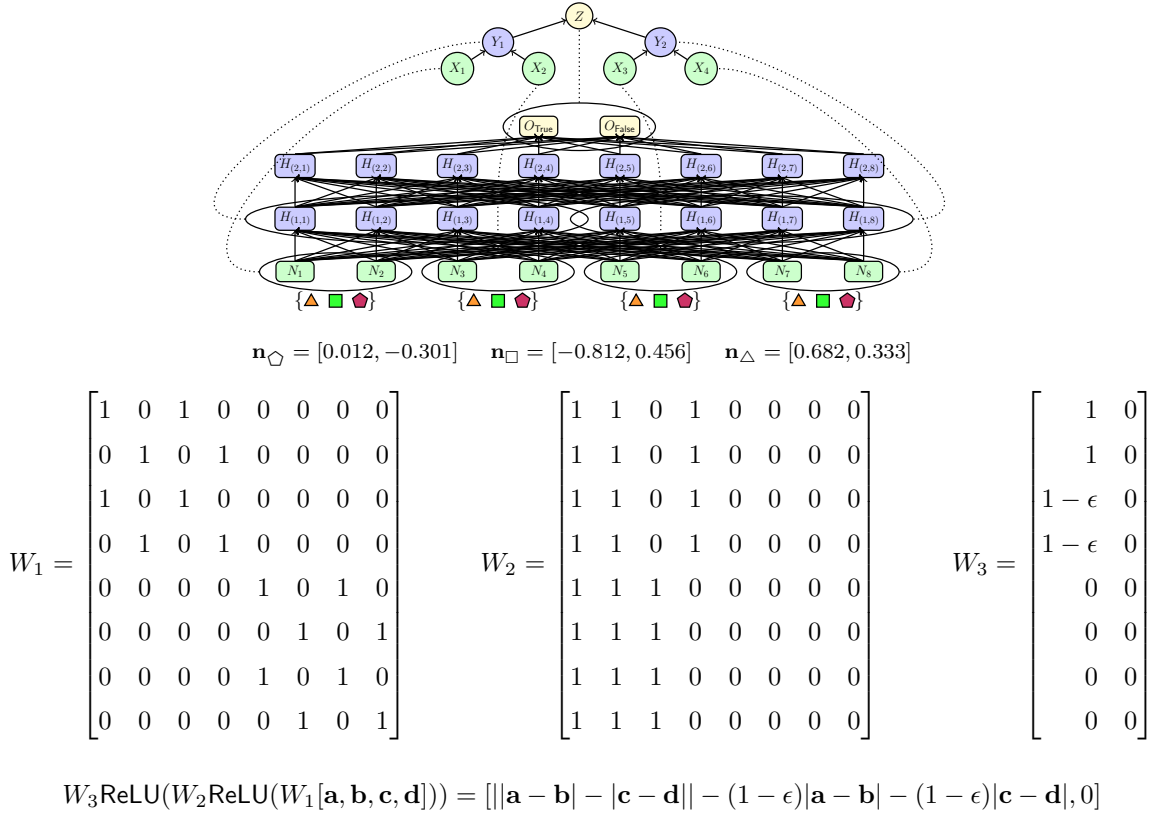

\paragraph{A Handcrafted Fully Connected Neural Network for Hierarchical Equality}
Define a neural network $\lowmodel$ consisting of eight `input' neurons $\outputvars_{\highmodel} = \{\inneuron_1, \dots, \inneuron_8\}$, twenty-four `intermediate' neurons $\hiddenneuron_{(i,j)}$ for $1 \leq i \leq 3$ and $1 \leq i \leq 8$, and two `output' neurons $\outputvars_{\highmodel} = \{\outneuron_{\mathsf{True}},\outneuron_{\mathsf{False}}\}$. The values for each of these variables are the real numbers. We depict the causal graph in Figure~\ref{fig:hand}.

    \begin{figure}[tp]
    
  \hspace{1em}\resizebox{0.97\textwidth}{!}{ 
                      \input{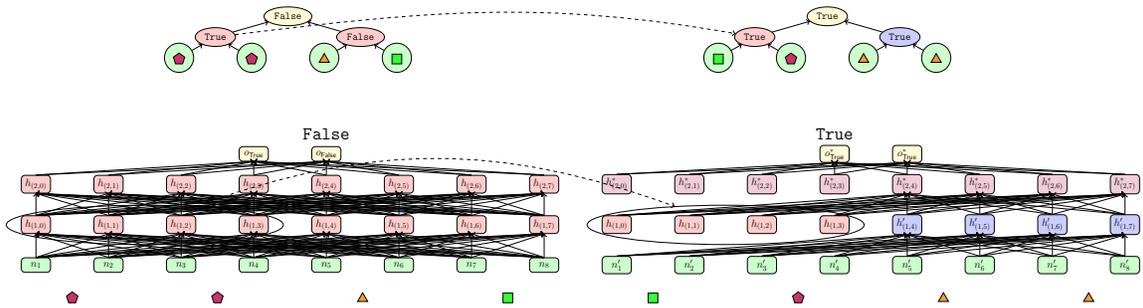}
                      }
    \caption{The result of aligned interchange intervention on the low-level fully-connected neural network and a high-level tree structured algorithm under the alignment in Figure~\ref{fig:hand}. Observe the equivalent counterfactual behavior across the two levels.}
    \label{fig:interchange}
    \end{figure}

Let $\inneurons$, $\hiddenneurons_1$, $\hiddenneurons_2$, $\hiddenneurons_3$ be the sets of variables for the first four layers, respectively. We define $\mechanism{\inneuron_k}$ (with no arguments) to be a constant function to $0$, for $1 \leq k \leq 8$. The intermediate and output neurons are determined by the network weights $\weights_{1},\weights_2 \in \mathbb{R}^{8\times8}$, and ${\weights}_3 \in \mathbb{R}^{8\times2}$.  For $1 \leq j \leq 8$, we define 
\begin{center}
\begin{tabular}{c c}
$\mechanism{\hiddenneuron_{(1,j)}}(\inneuronset) = \mathsf{ReLU}((\inneuronset \weights_1)_j)$ & $\mechanism{\hiddenneuron_{(2,j)}}(\hiddenneuronset_1) = \mathsf{ReLU}((\hiddenneuronset_1\weights_2)_j)$ \\
$\mechanism{\outneuron_{\mathsf{True}}}(\hiddenneuronset_2) = \mathsf{ReLU}((\hiddenneuronset_3{W}_3)_1)$ & $\mechanism{\outneuron_{\mathsf{False}}}(\hiddenneuronset_2) = \mathsf{ReLU}((\hiddenneuronset_2\weights_3)_2)$\\
\end{tabular}
\end{center}

The four shapes that are the input for the hierarchical equality task are represented in $\inneuronset_{\pentagon}, \inneuronset_{\square}, \inneuronset_{\bigtriangleup} \in \mathbb{R}^2$ by a pair of neurons with randomized activation values. The network outputs $\mathsf{True}$ if the value of the output logit $\outneuron_{\mathsf{True}}$ is larger than the value of $\outneuron_{\mathsf{False}}$, and $\mathsf{False}$ otherwise. We can simulate a network operating on the input $(\square, \pentagon, \square, \bigtriangleup)$ by performing an intervention setting $(\inneuron_1, \inneuron_2)$ and $(\inneuron_5,\inneuron_6)$ to $\inneuronset_{\square}$, $(\inneuron_3, \inneuron_4)$ to $\inneuronset_{\pentagon}$, and $(\inneuron_7,\inneuron_8)$ to $\inneuronset_{\bigtriangleup}$. 

In Figure~\ref{fig:hand}, we define the weights of the network $\lowmodel$, which have been handcrafted to implement the tree-structured algorithm $\highmodel$. 
\begin{figure}[tp]
\hspace{3em}
\input{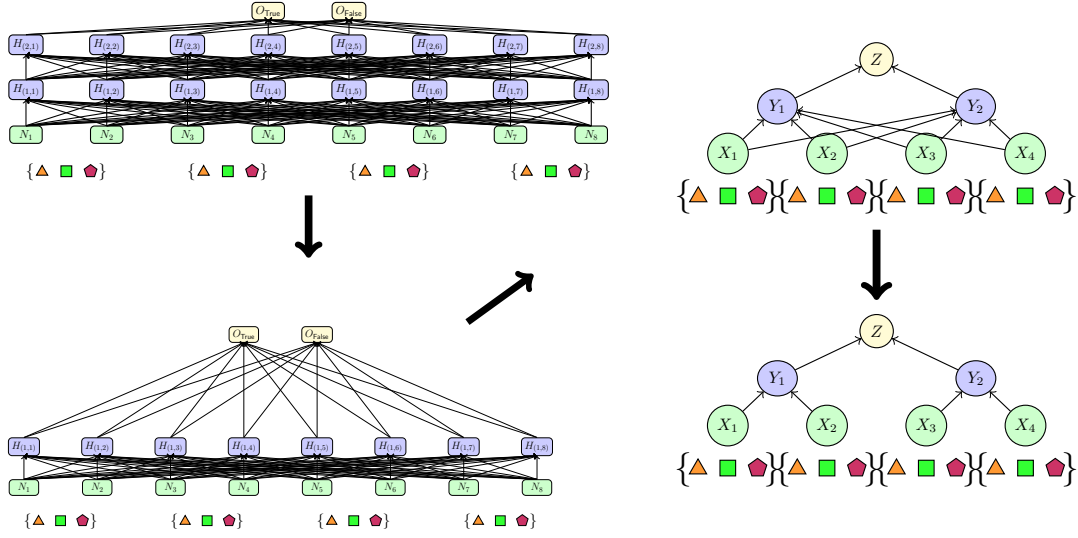}

\caption{An illustration of a fully-connected neural network being transformed into a tree structured algorithm by (1) marginalizing away neurons aligned with no high-level variable, (2) merging sets of variables aligned with high-level variables, and (3) merging the continuous values of neural activity into the symbolic values of the algorithm.}
\label{fig:constructive}
\end{figure}

\paragraph{An Alignment Between the Algorithm and the Neural Network}
The network  $\lowmodel$ was explicitly constructed to be abstracted by the algorithm $\highmodel$ under the alignment written formally below and depicted visually in Figure~\ref{fig:hand}.

\begin{center}
\begin{tabular}{c c c }
$\cellpart_{\onevarrr} = \{\outneuron_{\mathsf{True}}, \outneuron_{\mathsf{False}}\}$ & $\cellpart_{\onevar_k} = \{\inneuron_{2k-1}, \inneuron_{2k}\}$ & $\cellpart_{\onevarr_1} = \{\hiddenneuron_{(1,j)}: 1\leq j\leq4\}$ \\ 
 $\cellpart_{\onevarr_2} = \{\hiddenneuron_{(1,j)}: 5 \leq j \leq 8\}$ & \multicolumn{2}{c}{$\cellpart_{\bot} = \allvars \setminus (\cellpart_{\onevarrr} \cup \cellpart_{\onevarr_1} \cup \cellpart_{\onevarr_2} \cup \cellpart_{\onevar_1} \cup \cellpart_{\onevar_2} \cup \cellpart_{\onevar_3} \cup \cellpart_{\onevar_4})$}\\
\end{tabular}\newline

                      \resizebox{\textwidth}{!}{ 
\begin{tabular}{c c  }
$\cellfunc_{\onevarrr}(\outneuronval_{\mathsf{True}},\outneuronval_{\mathsf{False}} ) = \begin{cases} 
      \mathsf{True}  & \outneuronval_{\mathsf{True}} >\outneuronval_{\mathsf{False}} \\
      \mathsf{False} & \text{otherwise} \\
   \end{cases}$ & $\cellfunc_{\onevar_k}(\inneuronval_{2k-1}, \inneuronval_{2k}) = \begin{cases}
    {\square}& (\inneuronval_{2k-1}, \inneuronval_{2k}) = \inneuronset_{\square} \\
    \vspace{-2pt}
    {\pentagon}& (\inneuronval_{2k-1}, \inneuronval_{2k}) = \inneuronset_{\pentagon} \\
    {\bigtriangleup}& (\inneuronval_{2k-1}, \inneuronval_{2k}) = \inneuronset_{\bigtriangleup} \\
    \vspace{-1pt}
    \textsf{Undefined} & \text{otherwise}\\
   \end{cases}$ \\
\end{tabular}
}
\end{center}
We follow Definition~\ref{def:interchangealign} to define $\cellfunc_{\onevarr_1}$ and $\cellfunc_{\onevarr_2}$ on all interchange interventions. For each input  $ \inneuronset \in \{\inneuronset_{\pentagon}, \inneuronset_{\square}, \inneuronset_{\bigtriangleup} \}^{4}$, let $\pset = \cellfunc_{\inputvars_{\highmodel}}(\inneuronset)$ and $\{\hiddenneuronval_{(1,j)}: 1 \leq j \leq 8\} = \project{\sols(\lowmodel_{\inneuronset})}{\hiddenneurons}$. Then define 
\[\cellfunc_{\onevarr_1}(\hiddenneuronval_{(1,1)},\hiddenneuronval_{(1,2)}, \hiddenneuronval_{(1,3)},\hiddenneuronval_{(1,4)} ) = \project{\sols(\highmodel_{\pset})} {\onevarr_1}\]
\[\cellfunc_{\onevarr_2}(\hiddenneuronval_{(1,5)},\hiddenneuronval_{(1,6)}, \hiddenneuronval_{(1,7)},\hiddenneuronval_{(1,8)} ) = \project{\sols(\highmodel_{\pset})}{ \onevarr_2}\]
Otherwise leave $\cellfunc_{\onevarr_1}$ and $\cellfunc_{\onevarr_2}$ undefined.

Consider an intervention $\intpset$ in the domain of $\intmap^{\cellfunc}$. We have a fixed alignment for the input and output neurons, where $\intpset$ can have output values from the real numbers and input values from $\{\inneuronset_{\pentagon}, \inneuronset_{\square}, \inneuronset_{\bigtriangleup} \}^{4}$. The intermediate neurons are assigned high-level alignment by stipulation; $\mathbf{i}$ can only have intermediate variables that are realized on some input intervention $\lowmodel_{\inneuronset}$ for $\inneuronset \in \{\inneuronset_{\pentagon}, \inneuronset_{\square}, \inneuronset_{\bigtriangleup} \}^{4}$ (i.e., interchange interventions). Constructive abstraction will hold only if these stipulative alignments to intermediate variables do not violate the causal laws of $\lowmodel$.

\paragraph{The Algorithm Abstracts the Neural Network}\label{ex:abstract} 

Following Def.~\ref{def:interchangealign}, the domain of $\intmap^{\cellfunc}$ is restricted to $3^4$ input interventions, $(3^4)^2$ single-source hard interchange interventions for high-level interventions fixing either $\onevarr_1$ or $\onevarr_2$, and $(3^4)^3$ double-source hard interchange interventions for high-level interventions fixing both $\onevarr_1$ and $\onevarr_2$. 

This low-level neural network was hand crafted to be abstracted by the high-level algorithm under the alignment $\langle \cellpart, \cellfunc \rangle$. This means that for all $\intpset \in \dom(\intmap^{\cellfunc})$ we have
\begin{eqnarray}\setmap^{\cellfunc}\big(\sols\big(\lowmodel_{\intpset}\big)\big)  & = & \sols\big(\highmodel_{{\intmap^{\cellfunc}}(\intpset)}\big)\label{eq:networkalg}\end{eqnarray} 
In the 
\href{https://github.com/stanfordnlp/pyvene/blob/main/tutorials/advanced_tutorials/DAS_Main_Introduction.ipynb}{\textcolor{blue}{companion jupyter notebook}}, we provide code that verifies this is indeed the case. In Figure~\ref{fig:interchange}, we depict an aligned interchange intervention performed on $\lowmodel$ and $\highmodel$ with the base input $(\pentagon, \pentagon, \bigtriangleup, \square \normalsize)$ and a single source input $(\square, \pentagon, \bigtriangleup, \bigtriangleup)$. The central insight is that the network and algorithm have the same counterfactual behavior.

\paragraph{Decomposing the Alignment}\label{ex:alignment} 
Our decomposition of the alignment object in Section~\ref{sec:constructive} provides a new lens through which to view this result. The network $\lowmodel$ can be transformed into the algorithm $\highmodel$ through a marginalization, variable merge, and value merge. We visually depict the algorithm $\highmodel$ being constructed from the network $\lowmodel$ in Figure~\ref{fig:constructive}.

\paragraph{A Fully Connected Neural Network Trained on Hierarchical Equality}
Instead of handcrafting weights, we can also train the neural network $\lowmodel$ on the hierarchical equality task. Looking at the network weights provides no insight into whether or not it implements the algorithm $\highmodel$. A core result of \cite{Geiger-etal:2023:DAS} demonstrates that it is possible to learn a bijective translation $\setmap$ of the neural model $\lowmodel$ such that the algorithm $\highmodel$ is a constructive abstraction of the transformed model $\setmap(\lowmodel)$. This bijective translation is in the form of an orthogonal matrix that rotates a hidden vector into a new coordinate system. The method is Distributed Alignment Search which is covered in Section~\ref{sec:DAS} and the result is replicated in our \href{https://github.com/stanfordnlp/pyvene/blob/main/tutorials/advanced_tutorials/DAS_Main_Introduction.ipynb}{\textcolor{blue}{companion jupyter notebook}}.

\subsection{Example: Causal Abstraction with Cycles and Infinite Variables}\label{sec:future}
Causal abstraction is a highly expressive, general purpose framework. However, our example in Section~\ref{sec:example} involved only finite and acyclic models. To demonstrate the framework's expressive capacity, we will define a causal model with infinitely many variables with infinite value ranges that implements the bubble sort algorithm on lists of arbitrary length. Then, we show this acyclic model can be abstracted into a cyclic process with an equilibrium state. 

\paragraph{A Causal Model for Bubble Sort}\label{sec:example2}
Bubble sort is an iterative algorithm. On each iteration, the first two members of the sequence are compared and swapped if the left element is larger than the right element; then the second and third member of the resulting list are compared and possibly swapped, and so on until the end of the list is reached. This process is repeated until no more swaps are needed. 

Define the causal model $\model$ to have the following (countably) infinite variables and values 
\[\begin{tabular}{cc c}
     $\allvars = \{\onevar_i^j,\onevarr_i^j,\onevarrr^j_i: i,j \in \{1,2,3,4,5,\dots\}\}$& $\values{\onevar^j_i} = \values{\onevarr^j_i} = \{1,2,3,4,5,\dots\} \cup \{\bot\}$\\ 
 \multicolumn{2}{c}{$\values{\onevarrr_i^j} = \{\mathsf{True}, \mathsf{False}, \bot\}$}\\    
\end{tabular}\]

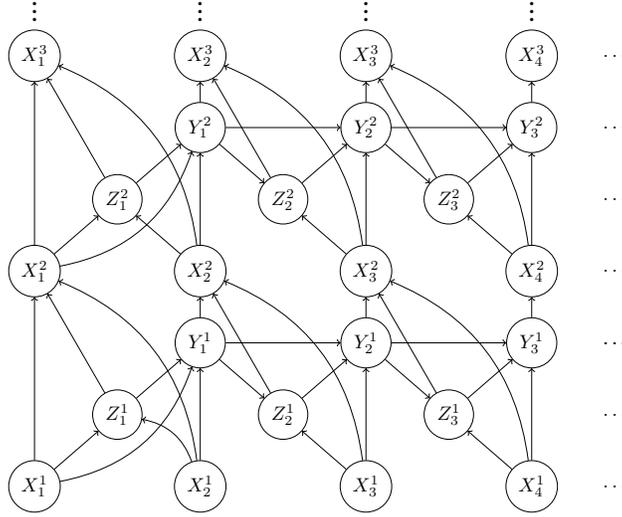
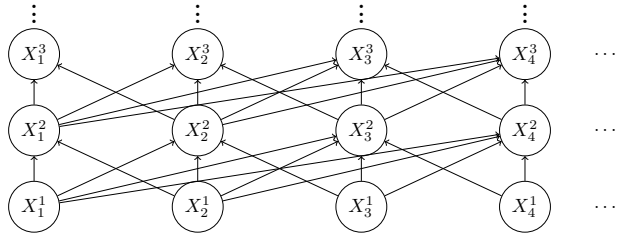
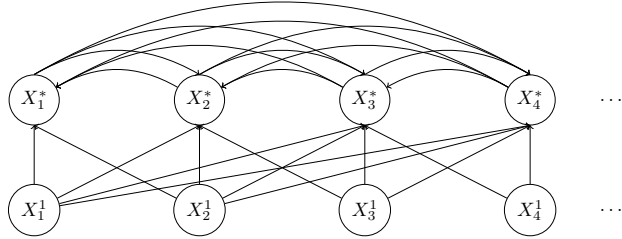
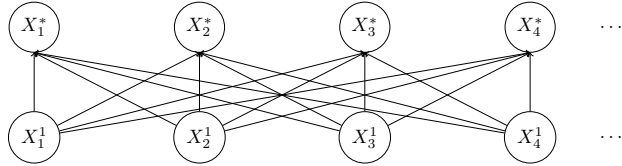
\begin{figure}
\begin{subfigure}[b]{\textwidth}
\centering
\resizebox{0.55\textwidth}{!}{
\begin{tikzpicture}

\def\xshift{3}
\def\yshift{1.3}

\node[circle,draw] (X11) at (0,0) {$\onevar^1_1$};
\node[circle,draw] (X12) at (\xshift, 0) {$\onevar^1_2$};
\node[circle,draw] (X13) at (\xshift*2, 0) {$\onevar^1_3$};
\node[circle,draw] (X14) at (\xshift*3, 0) {$\onevar^1_4$};
\node[] (X15) at (\xshift*3.5, 0) {$\huge\hdots$};

\node[circle,draw] (X21) at (0,\yshift*3) {$\onevar^2_1$};
\node[circle,draw] (X22) at (\xshift, \yshift*3) {$\onevar^2_2$};
\node[circle,draw] (X23) at (\xshift*2, \yshift*3) {$\onevar^2_3$};
\node[circle,draw] (X24) at (\xshift*3, \yshift*3) {$\onevar^2_4$};
\node[] (X25) at (\xshift*3.5, \yshift*3) {$\huge\hdots$};

\node[circle,draw] (X31) at (0,\yshift*6) {$\onevar^3_1$};
\node[circle,draw] (X32) at (\xshift, \yshift*6) {$\onevar^3_2$};
\node[circle,draw] (X33) at (\xshift*2, \yshift*6) {$\onevar^3_3$};
\node[circle,draw] (X34) at (\xshift*3, \yshift*6) {$\onevar^3_4$};
\node[] (X35) at (\xshift*3.5, \yshift*6) {$\huge\hdots$};

\node[] (d21) at (0,\yshift*6.7) {$\huge\vdots$};
\node[] (d22) at (\xshift, \yshift*6.7) {$\huge\vdots$};
\node[] (d23) at (\xshift*2, \yshift*6.7) {$\huge\vdots$};
\node[] (d24) at (\xshift*3, \yshift*6.7) {$\huge\vdots$};

\node[] (X25) at (\xshift*3.5, \yshift*2) {$\huge\hdots$};

\node[circle,draw] (B11) at (\xshift/2,\yshift) {$\onevarrr^1_1$};
\node[circle,draw] (B12) at (\xshift*3/2,\yshift) {$\onevarrr^1_2$};
\node[circle,draw] (B13) at (\xshift*5/2,\yshift) {$\onevarrr^1_3$};
\node[] () at (\xshift*7/2,\yshift) {$\huge\hdots$};

\node[circle,draw] (B21) at (\xshift/2,\yshift*4) {$\onevarrr^2_1$};
\node[circle,draw] (B22) at (\xshift*3/2,\yshift*4) {$\onevarrr^2_2$};
\node[circle,draw] (B23) at (\xshift*5/2,\yshift*4) {$\onevarrr^2_3$};
\node[] () at (\xshift*7/2,\yshift*4) {$\huge\hdots$};

\node[circle,draw] (Y11) at (\xshift,2*\yshift) {$\onevarr^1_1$};
\node[circle,draw] (Y12) at (\xshift*2,2*\yshift) {$\onevarr^1_2$};
\node[circle,draw] (Y13) at (\xshift*3,2*\yshift) {$\onevarr^1_3$};
\node[] () at (\xshift*7/2,\yshift*2) {$\huge\hdots$};

\node[circle,draw] (Y21) at (\xshift,5*\yshift) {$\onevarr^2_1$};
\node[circle,draw] (Y22) at (\xshift*2,5*\yshift) {$\onevarr^2_2$};
\node[circle,draw] (Y23) at (\xshift*3,5*\yshift) {$\onevarr^2_3$};
\node[] () at (\xshift*7/2,\yshift*5) {$\huge\hdots$};

\draw[->] (X11) to (B11);
\draw[->, bend right] (X12) to (B11);

\draw[->] (X11) to (X21);
\draw[->, bend right] (X12) to (X21);
\draw[->] (B11) to (X21);

\draw[->, bend right] (X11) to (Y11);
\draw[->] (X12) to (Y11);
\draw[->] (B11) to (Y11);

\draw[->] (Y11) to (B12);
\draw[->] (X13) to (B12);

\draw[->] (Y11) to (X22);
\draw[->] (B12) to (X22);
\draw[->, bend right] (X13) to (X22);

\draw[->] (Y11) to (Y12);
\draw[->] (X13) to (Y12);
\draw[->] (B12) to (Y12);

\draw[->] (Y12) to (Y13);
\draw[->] (X14) to (Y13);
\draw[->] (B13) to (Y13);

\draw[->] (Y12) to (B13);
\draw[->] (X14) to (B13);

\draw[->] (Y12) to (X23);
\draw[->] (B13) to (X23);
\draw[->, bend right] (X14) to (X23);

\draw[->] (Y13) to (X24);

\draw[->] (X21) to (B21);
\draw[->] (X22) to (B21);

\draw[->] (X21) to (X31);
\draw[->, bend right] (X22) to (X31);
\draw[->] (B21) to (X31);

\draw[->, bend right] (X21) to (Y21);
\draw[->] (X22) to (Y21);
\draw[->] (B21) to (Y21);

\draw[->] (Y21) to (B22);
\draw[->] (X23) to (B22);

\draw[->] (Y21) to (X32);
\draw[->] (B22) to (X32);
\draw[->, bend right] (X23) to (X32);

\draw[->] (Y21) to (Y22);
\draw[->] (X23) to (Y22);
\draw[->] (B22) to (Y22);

\draw[->] (Y22) to (Y23);
\draw[->] (X24) to (Y23);
\draw[->] (B23) to (Y23);

\draw[->] (Y22) to (B23);
\draw[->] (X24) to (B23);

\draw[->] (Y22) to (X33);
\draw[->] (B23) to (X33);
\draw[->, bend right] (X24) to (X33);

\draw[->] (Y23) to (X34);

\end{tikzpicture}}
\caption{A causal model that represents the bubble sort algorithm.}
\label{fig:bub1}
\end{subfigure}

\begin{subfigure}[b]{\textwidth}
\centering
\resizebox{0.55\textwidth}{!}{
\begin{tikzpicture}
\def\xshift{3}
\def\yshift{1.4}

\node[circle,draw] (X11) at (0,0) {$\onevar^1_1$};
\node[circle,draw] (X12) at (\xshift, 0) {$\onevar^1_2$};
\node[circle,draw] (X13) at (\xshift*2, 0) {$\onevar^1_3$};
\node[circle,draw] (X14) at (\xshift*3, 0) {$\onevar^1_4$};
\node[] (X15) at (\xshift*3.5, 0) {$\huge\dots$};

\node[circle,draw] (X21) at (0,\yshift*1) {$\onevar^2_1$};
\node[circle,draw] (X22) at (\xshift, \yshift*1) {$\onevar^2_2$};
\node[circle,draw] (X23) at (\xshift*2, \yshift*1) {$\onevar^2_3$};
\node[circle,draw] (X24) at (\xshift*3, \yshift*1) {$\onevar^2_4$};
\node[] (X25) at (\xshift*3.5, \yshift*1) {$\huge\dots$};

\node[circle,draw] (X31) at (0,\yshift*2) {$\onevar^3_1$};
\node[circle,draw] (X32) at (\xshift, \yshift*2) {$\onevar^3_2$};
\node[circle,draw] (X33) at (\xshift*2, \yshift*2) {$\onevar^3_3$};
\node[circle,draw] (X34) at (\xshift*3, \yshift*2) {$\onevar^3_4$};
\node[] (X35) at (\xshift*3.5, \yshift*2) {$\huge\dots$};

\node[] (d21) at (0,\yshift*2.6) {$\huge\vdots$};
\node[] (d22) at (\xshift, \yshift*2.6) {$\huge\vdots$};
\node[] (d23) at (\xshift*2, \yshift*2.6) {$\huge\vdots$};
\node[] (d24) at (\xshift*3, \yshift*2.6) {$\huge\vdots$};

\draw[->] (X11) to (X24);
\draw[->] (X12) to (X24);
\draw[->] (X13) to (X24);
\draw[->] (X14) to (X24);

\draw[->] (X21) to (X34);
\draw[->] (X22) to (X34);
\draw[->] (X23) to (X34);
\draw[->] (X24) to (X34);

\draw[->] (X11) to (X23);
\draw[->] (X12) to (X23);
\draw[->] (X13) to (X23);
\draw[->] (X14) to (X23);

\draw[->] (X21) to (X33);
\draw[->] (X22) to (X33);
\draw[->] (X23) to (X33);
\draw[->] (X24) to (X33);

\draw[->] (X11) to (X22);
\draw[->] (X12) to (X22);
\draw[->] (X13) to (X22);

\draw[->] (X21) to (X32);
\draw[->] (X22) to (X32);
\draw[->] (X23) to (X32);

\draw[->] (X11) to (X21);
\draw[->] (X12) to (X21);

\draw[->] (X21) to (X31);
\draw[->] (X22) to (X31);
\end{tikzpicture}
}
\caption{The causal model from Figure~\ref{fig:bub1} with the variables $\onevarr^i_j$ and $\onevarrr^i_j$ marginalized.}
\label{fig:bub2}
\end{subfigure}
     
\begin{subfigure}[b]{\textwidth}
\centering
\resizebox{0.55\textwidth}{!}{
\begin{tikzpicture}
\def\xshift{3}
\def\yshift{1.4}

\node[circle,draw] (X11) at (0,0) {$\onevar^*_1$};
\node[circle,draw] (X12) at (\xshift, 0) {$\onevar^*_2$};
\node[circle,draw] (X13) at (\xshift*2, 0) {$\onevar^*_3$};
\node[circle,draw] (X14) at (\xshift*3, 0) {$\onevar^*_4$};
\node[] (X15) at (\xshift*3.5, 0) {$\huge\dots$};

\node[circle,draw] (X01) at (0,-2) {$\onevar^1_1$};
\node[circle,draw] (X02) at (\xshift, -2) {$\onevar_2^1$};
\node[circle,draw] (X03) at (\xshift*2,-2) {$\onevar_3^1$};
\node[circle,draw] (X04) at (\xshift*3,-2) {$\onevar_4^1$};
\node[] (X05) at (\xshift*3.5, -2) {$\huge\dots$};

\draw[->, bend left] (X11.north) to (X12.north);
\draw[->, bend left] (X11.north) to (X13.north);
\draw[->, bend left] (X11.north) to (X14.north);

\draw[->, bend right] (X12) to (X11);
\draw[->, bend left] (X12.north) to (X13.north);
\draw[->, bend left] (X12.north) to (X14.north);

\draw[->, bend right] (X13) to (X12);
\draw[->, bend left] (X13.north) to (X14.north);
\draw[->, bend right] (X13) to (X11);

\draw[->, bend right] (X14) to (X13);
\draw[->, bend right] (X14) to (X12);
\draw[->, bend right] (X14) to (X11);

\draw[->] (X01) to (X11.south);
\draw[->] (X01) to (X12.south);
\draw[->] (X01) to (X13.south);
\draw[->] (X01) to (X14.south);

\draw[->] (X02) to (X11.south);
\draw[->] (X02) to (X12.south);
\draw[->] (X02) to (X13.south);
\draw[->] (X02) to (X14.south);

\draw[->] (X03) to (X12.south);
\draw[->] (X03) to (X13.south);
\draw[->] (X03) to (X14.south);

\draw[->] (X04) to (X13.south);
\draw[->] (X04) to (X14.south);

\end{tikzpicture}}
\caption{The causal model from Figure~\ref{fig:bub2} with the variables $\onevar^2_i, \onevar^3_i, \dots$ merged for all $i>0$. The values of these new variables contain the full history of the algorithm, e.g., the value of $\onevar^*_1$ a sequence containing the first element in the list after each bubbling iteration.}
\label{fig:bub3}
\end{subfigure}
\begin{subfigure}[b]{\textwidth}
\centering
\resizebox{0.55\textwidth}{!}{
\begin{tikzpicture}
\def\xshift{3}
\def\yshift{1.4}

\node[circle,draw] (X11) at (0,0) {$\onevar^*_1$};
\node[circle,draw] (X12) at (\xshift, 0) {$\onevar^*_2$};
\node[circle,draw] (X13) at (\xshift*2, 0) {$\onevar^*_3$};
\node[circle,draw] (X14) at (\xshift*3, 0) {$\onevar^*_4$};
\node[] (X15) at (\xshift*3.5, 0) {$\huge\dots$};

\node[circle,draw] (X01) at (0,-2) {$\onevar^1_1$};
\node[circle,draw] (X02) at (\xshift, -2) {$\onevar_2^1$};
\node[circle,draw] (X03) at (\xshift*2,-2) {$\onevar_3^1$};
\node[circle,draw] (X04) at (\xshift*3,-2) {$\onevar_4^1$};
\node[] (X05) at (\xshift*3.5, -2) {$\huge\dots$};

\draw[->] (X01) to (X11.south);
\draw[->] (X01) to (X12.south);
\draw[->] (X01) to (X13.south);
\draw[->] (X01) to (X14.south);

\draw[->] (X02) to (X11.south);
\draw[->] (X02) to (X12.south);
\draw[->] (X02) to (X13.south);
\draw[->] (X02) to (X14.south);

\draw[->] (X03) to (X11.south);
\draw[->] (X03) to (X12.south);
\draw[->] (X03) to (X13.south);
\draw[->] (X03) to (X14.south);

\draw[->] (X04) to (X11.south);
\draw[->] (X04) to (X12.south);
\draw[->] (X04) to (X13.south);
\draw[->] (X04) to (X14.south);

\end{tikzpicture}} 
\caption{The causal model from Figure~\ref{fig:bub3} with the values of each variable $\onevar^*_j$ merged for all $j>0$, e.g., the value of $X^*_1$ is the first element in the sorted list.}
\label{fig:bub4}
\end{subfigure}
    \centering
    \caption{Abstractions of the bubble sort causal model. }
    \label{fig:bub}
\end{figure}

The causal structure of $\model$ is depicted in Figure~\ref{fig:bub1}. The $\bot$ value will indicate that a variable is not being used in a computation, much like a blank square on a Turing machine. The (countably) infinite sequence of variables $\onevar^1_1, \onevar^1_2, \dots$ contains the unsorted input sequence, where an input sequence of length $k$ is represented by setting $\onevar_1^1, \dots, \onevar_k^1$ to encode the sequence and doing nothing to the infinitely many remaining input variables $\onevar_j$ for $j>k$.

For a given row $j$ of variables, the variables $\onevarrr^j_i$ store the truth-valued output of the comparison of two elements, the variables $\onevarr^j_i$ contain the values being `bubbled up' through the sequence, and the variables $\onevar^j_i$ are partially sorted lists resulting from $j-1$ passes through the algorithm. When there are rows $j$ and $j+1$ such that $\onevar^j_i$ and $\onevar^{j-1}_i$ take on the same value for all $i$, the output of the computation is the sorted sequence found in both of these rows. 

We define the mechanisms as follows. The input variables $\onevar_i^1$ have constant functions to $\bot$. The variable $\onevarrr_1^1$ is $\bot$ if either of $\onevar^1_1$ or $\onevar^1_2$ is $\bot$, $\mathsf{True}$ if the value of $\onevar_1^1$ is greater than $\onevar^1_2$, and $\mathsf{False}$ otherwise. The variable $\onevarr_1^1$ is $\bot$ if $\onevarrr^1_1$ is $\bot$, $\onevar^1_1$ if $\onevarrr_1^1$ is $\mathsf{True}$, and $\onevar^1_2$ if $\onevarrr_1^1$ is $\mathsf{False}$. The remaining intermediate variables can be uniformly defined:
\[\begin{tabular}{c@{\hspace{0.5em}}c}
    $\mechanism{\onevar_i^j}(\onevall^{j-1}_{i-1}, \onevalll^{j-1}_i, \oneval^{j-1}_{i+1}) = \begin{cases}
     \oneval^{j-1}_{i+1} & \onevalll^{j-1}_i = \mathsf{True}\\
     \onevall^{j-1}_{i-1} & \onevalll^{j-1}_i = \mathsf{False}\\
     \bot & \onevalll^{j-1}_i = \bot\\
     \end{cases}$ & 
     $\mechanism{\onevarr_i^j}(\onevall^{j}_{i-1}, \onevalll^{j}_i, \oneval^{j}_{i+1}) = \begin{cases}
     \oneval^{j}_{i+1} & \onevalll^{j}_i = \mathsf{True}\\
     \onevall^{j}_{i-1} & \onevalll^{j}_i = \mathsf{False}\\
     \bot & \onevalll^{j}_i = \bot\\
     \end{cases} $\\
\end{tabular}\]

\[\mechanism{\onevarrr^j_i}(\onevall^j_i, \oneval^j_{i+1}) = \begin{cases} \onevall^j_i < \oneval^j_{i+1} & \onevall^j_i  \not = \bot \text{ and } \oneval^j_{i+1} \not = \bot \\
     \bot & \text{otherwise}
     \end{cases}\]

This causal model is countably infinite, supporting both sequences of arbitrary length and an arbitrary number of sorting iterations. 

\paragraph{Abstracting Bubble Sort}
Suppose we aren't concerned with how, exactly, each iterative pass of the bubble sort algorithm is implemented. Then we can marginalize away the variables $\varrrs = \{\onevarrr_i^j\}$ and $\varrs = \{\onevarr_i^j\}$ and reason about the resulting model instead (Figure~\ref{fig:bub2}). Define the mechanisms of this model recursively with base case $\mechanism{\onevar_1^j}(\oneval_1^{j-1}, \oneval_2^{j-1}) =\mathsf{Min}(\oneval_1^{j-1}, \oneval_2^{j-1} ) $ for $j>1$ and recursive case
\[\mechanism{\onevar_i^j}(\oneval^{j-1}_{1}, \oneval^{j-1}_{2},\dots, \oneval^{j-1}_{i+1}) = \mathsf{Min}(\oneval^{j-1}_{i+1}, \mathsf{Max}(\oneval^{j-1}_{i}, \mechanism{\onevar_{i-1}^j}(\oneval^{j-1}_{1}, \oneval^{j-1}_{2},\dots, \oneval^{j-1}_{i})))\] 

Suppose instead that our only concern is whether the input sequence is sorted. We can further abstract the causal model using variable merge with the partition $\cellpart_{\onevar^*_i} = \{\onevar^j_i: j \in \{2,3,\dots\}\}$ for each $i \in \{1, 2, \dots\}$. The result is a model (Figure~\ref{fig:bub3}) where each variable $\onevar^*_i$ takes on the value of an infinite sequence. There are causal connections to and from $\onevar^*_i$ and $\onevar^*_j$ for any $i \not = j$, because the infinite sequences stored in each variable must jointly be a valid run of the bubble sort algorithm. This is a cyclic causal process with an equilibrium point.

Next, we can value-merge with a family of functions $\delta$ where the input variable functions $\delta_{\onevar^1_i}$ are identity functions and the other functions $\delta_{\onevar^*_i}$ output the constant value to which an eventually-constant infinite sequence converges. The mechanisms for the resulting model (Figure~\ref{fig:bub4}) simply map unsorted input sequences to sorted output sequences. 
\newpage

\section{ A Common Language for Mechanistic Interpretability}\label{sec:commonlanguage}
The central claim of this paper is that causal abstraction provides a theoretical foundation for mechanistic interpretability. Equipped with a general theory of causal abstraction, we will provide mathematically precise definitions for a handful of core mechanistic interpretability concepts and show that a wide range of methods can be viewed as special cases of causal abstraction analysis. A tabular summary of this section can be found in Table \ref{tab:methods}.

\begin{table}[hbtp]
\footnotesize
\caption{Interpretability Methods}
\label{tab:methods}
\begin{tabular}{p{0.23\textwidth}p{0.7\textwidth}}
\toprule
Behavioral Methods \newline (Section~\ref{sec:behave}) & \begin{itemize}[leftmargin=*,nosep]
    \item \textbf{Feature attribution} \citep{Zeiler2014,Ribeiro2016,Lundberg:2017}
    \item \textbf{Integrated gradients} \citep{sundararajan2017axiomatic}
    \item \textbf{Effects of real-world concepts on models} \citep{Goyal,CausalLM,abraham-etal-2022-cebab, wu-etal-2022-cpm}
\end{itemize} \\
\midrule
 Patching\newline Activations with \newline Interchange\newline Interventions \newline (Section~\ref{sec:abstractionwithinterchange}) & \begin{itemize}[leftmargin=*,nosep]
    \item \textbf{Interchange interventions} \citep{geiger-etal:2020:blackbox,vig2020causal,geiger2021causal,Li2021,chan2022causal,wang2023interpretability,Lieberum:2023,huang-etal-2023-rigorously,Hase:2023, Cunningham:2023,Davies:2023,tigges2023linear,feng2024how,patchscope, Todd2024}
    \item \textbf{Path patching} \citep{goldowskydill2023localizing,wang2023interpretability,Hanna:2023, prakash2024finetuning}
    \item \textbf{Causal mediation analysis} \citep{vig2020causal,finlayson-etal-2021-causal,Meng:2022,Stolfo23,mueller2024questrightmediatorhistory}
\end{itemize} \\
\midrule
Ablation-Based Analysis \newline (Section~\ref{sec:ablation}) & \begin{itemize}[leftmargin=*,nosep]
    \item \textbf{Concept erasure} \citep{Ravfogel:2020,linear_adversarial_concept_Eraser, kernelized_concept_Eraser,Ravfogel2023,Elazar-etal-2020, Lovering2022, LEACE}
    \item \textbf{Sub-circuit analysis} \citep{sixteen_heads_better_than_one,lowcomplexity_probing, Csordas2021, cammarata2020thread,olsson2022context,chan2022causal,Lepori2023a,Lepori2023b,wang2023interpretability, automated_discovery, Nanda2023}
    \item \textbf{Causal scrubbing} \citep{chan2022causal}
\end{itemize} \\
\midrule
Modular Feature\newline Learning \newline(Section~\ref{sec:localization}) & 
\begin{itemize}[leftmargin=*,nosep]
    \item \textbf{Principal Component Analysis} \citep{bolukbasi2016man,chormai2022disentangled, marks2023geometry,tigges2023linear}
    \item \textbf{Sparse autoencoders} \citep{bricken2023monosemanticity,Cunningham:2023, huben2024sparse, marks2024sparse}
    \item \textbf{Differential Binary Masking} \citep{Cao2020, de2021sparse, Csordas2021,Davies:2023, prakash2024finetuning, huang2024ravel}
    \item \textbf{Probing} \citep{peters-etal-2018-dissecting,tenney-etal-2019-bert,Hupkes:2018}
    \item \textbf{Difference of means} \citep{tigges2023linear, marks2023geometry}
    \item \textbf{Distributed Alignment Search} \citep{Geiger-etal:2023:DAS,Wu:Geiger:2023:BDAS, tigges2023linear, arora2024causalgym, huang2024ravel, minder2024, Feng2024Prop, Diego2024, grant2025emergentsymbollikenumbervariables}
\end{itemize} \\
\midrule
Activation Steering \newline (Section~\ref{sec:activationsteering}) & \citep{Giulianelli:2018, Bau19a,Soulos:2020,Besserve20,subramani2022extracting,turner2023activation,zou2023representation,vogel2024repeng,liInferenceTimeInterventionEliciting2024, wu2024advancing, REFT}\\
\midrule
Training Models to be\newline Interpretable \newline (Section~\ref{sec:trainingmodels})& \citep{geiger-etal-2021-iit,wu-etal-2021-distill,wu-etal-2022-cpm,elhage2022solu,Hewitt2023,bottleneck2,bottleneck3,huang-etal-2023-inducing, tamkin2024codebook,gomez2024, zur-etal-2024-updating,KAN}\\
\bottomrule
\end{tabular}
\end{table}

\subsection{Polysemantic Neurons, the Linear Representation Hypothesis, and Modular Features via Intervention Algebras}
A vexed question when analyzing black box AI is how to decompose a deep learning system into constituent parts. Should the units of analysis be real-valued activations, directions in the activation space of vectors, or entire model components?  Localizing an abstract concept to a component of a black box AI would be much easier if neurons were a sufficient unit of analysis. However, it has long been known that artificial (and biological) neural networks have polysemantic neurons that participate in the representation of multiple high-level concepts \citep{Smolensky1988a, PDP1, PDP2, Thorpe89}. Therefore, individual neural activations are insufficient as units of analysis in interpretability, a fact that has been recognized in the recent literature %
\citep{Harradon:2018,cammarata2020thread,olah2020zoom, goh2021multimodal, elhage2021mathematical,bolukbasi:2021, Geiger-etal:2023:DAS,Gurnee:2023,huang-etal-2023-rigorously}.

Perhaps the simplest case of polysemantic neurons is  where some rotation can be applied to the neural activations such that the dimensions in the new coordinate system are monosemantic \citep{Smolensky1988a, elhage2021mathematical,Scherlis:2022,Geiger-etal:2023:DAS}. Indeed, the \textit{linear representation hypothesis} \citep{mikolov-etal-2013-linguistic,elhage2022superposition,Nanda2023, Park:2023, Jiang:2024} states that linear representations will be sufficient for analyzing the complex non-linear building blocks of deep learning models. We are concerned that this is too restrictive. The ideal theoretical framework won't bake in an assumption like the linear representation hypothesis, but rather support any and all decompositions of a deep learning system into \textit{modular features} that each have separate mechanisms from one another. We should have the flexibility to choose the units of analysis, free of restrictive assumptions that may rule out meaningful structures. Whether a particular decomposition of a deep learning system into modular features is useful for mechanistic interpretability should be understood as an empirical hypothesis that can be falsified through experimentation.

Our theory of causal abstraction supports a flexible, yet precise conception of modular features via intervention algebras (Section~\ref{sec:intalgebras}). An intervention algebra formalizes the notion of a set of separable components with distinct mechanisms, satisfying the fundamental algebraic properties of commutativity and left-annihilativity (see \ref{keyproperty1} and \ref{keyproperty2} in Definition~\ref{keyproperties}). Individual activations, orthogonal directions in vector space, and model components (e.g. attention heads) are all separable components with distinct mechanisms in this sense. %
A bijective translation (Section~\ref{sec:bijtrans}) gives access to such features while preserving the overall mechanistic structure of the model. We propose to define modular features as any set of variables that form an intervention algebra accessed by a bijective translation. 

If the linear representation hypothesis is correct, then rotation matrices should be sufficient bijective translations for mechanistic interpretability. If not, there will be cases where non-linear bijective translations will be needed to discover modular features that are not linearly accessible, e.g., the `onion' representations found by \cite{csordas2024recurrent} in simple recurrent neural networks. Our conception of modular features enables us to remain agnostic to the exact units of analysis that will prove essential.

\subsection{Graded Faithfulness via Approximate Abstraction}
Informally, faithfulness has been defined as the degree to which an explanation accurately represents the `true reasoning process behind a model’s behavior' \citep{notnotAttention, Jacovi2020, Lyu, FaithfulnessCompare}. Crucially, faithfulness should be a graded notion \citep{Jacovi2020}, but precisely which metric of faithfulness is correct will depend on the situation. It could be that for safety reasons there are some domains of inputs for which we need a perfectly faithful interpretation of a black box AI, while for others it matters less. Ideally, we can defer the exact details to be filled in based on the use case. %
This would allow us to provide a variety of graded faithfulness metrics that facilitates apples-to-apples comparisons between existing (and future) mechanistic interpretability methods.

Approximate transformation (Section~\ref{section:approximate}) provides the needed flexible notion of graded faithfulness. The similarity metric between high-level and low-level states, the probability distribution over evaluated interventions, and the summary statistic used to aggregate individual similarity scores are all points of variation that enable our notion of approximate transformation to be adapted to a given situation. Interchange intervention accuracy \citep{geiger-etal-2021-iit, Geiger-etal:2023:DAS, Wu:Geiger:2023:BDAS}, probability or logit difference \citep{Meng:2022,chan2022causal,wang2023interpretability,zhang2024towards}, and KL-divergence all can be understood via approximate transformation.

\subsection{Behavioral Evaluations as Abstraction by a Two Variable Chains}\label{sec:behave}
The behavior of an AI model is simply the function from inputs to outputs that the model implements. Behavior is trivial to characterize in causal terms; any input--output behavior can be represented by a model with input variables directly connected to output variables. 

\subsubsection{LIME: Behavioral Fidelity as Approximate Abstraction}
Feature attribution methods ascribe scores to input features that capture the `impact' of a feature on model behavior. Gradient-based feature attribution methods \citep{Zeiler2014,springerberg2014,Shrikumar16,Binder16, Lundberg:2017, kim2018interpretability,Narendra:2018, lundberg2019consistent,schrouff2022best} measure causal properties when they satisfy some basic axioms \citep{sundararajan2017axiomatic}. In particular, \cite{geiger2021causal} provide a natural causal interpretation of the integrated gradients method and \cite{chattopadhyay19a} argue for a direct measurement of a feature's individual causal effect.

Among the most popular feature attribution methods is LIME \citep{Ribeiro2016}, which learns an interpretable model that locally approximates an uninterpretable model. LIME defines an explanation to be faithful to the degree that the interpretable model agrees with local input--output behavior. While not conceived as a causal explanation method, when we interpret priming a model with an input as an intervention, it becomes obvious that two models having the same local input--output behavior is a matter of causality. 

Crucially, however, the interpretable model lacks any connection to the internal causal dynamics of the uninterpretable model. In fact, it is presented as a benefit that LIME is a model-agnostic method that provides the same explanations for models with identical behaviors, but different internal structures. Without further grounding in causal abstraction, methods like LIME do not tell us anything meaningful about the abstract causal structure between input and output. 

\begin{definition}[LIME Fidelity of Interpretable Model]

Let $\lowmodel$ and $\highmodel$ be models with identical input and output spaces. Let $\mathsf{Distance}(\cdot, \cdot)$ compute some measure of distance between outputs. Given an input $\pset \in \values{\inputvars_{\highmodel}}$, let $\Delta_{\pset} \subseteq \values{\inputvars_{\highmodel}}$ be a finite neighborhood of inputs close to $\pset$. The LIME fidelity of using $\highmodel$ to interpret $\lowmodel$ on the input $\pset$ is given as:
\[\mathsf{LIME}(\highmodel, \lowmodel, \Delta_{\pset}) = \frac{1}{|\Delta_{\pset}|}\sum_{\pset' \in \Delta_{\pset}}\mathsf{Distance}( \project{\sols(\lowmodel_{\pset'})}{ \outputvars_{\lowmodel}},\project{\sols(\highmodel_{\pset'})}{ \outputvars_{\highmodel}})\]
\end{definition}

The uninterpretable model $\lowmodel$ is an AI model with fully-connected causal structure:  
   
     \centerline{\resizebox{0.4\textwidth}{!}
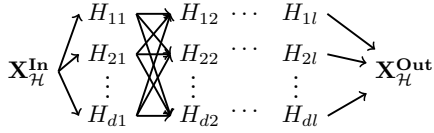
{
        \begin{tikzpicture}[node distance = 2.0cm]
     \tikzstyle{nnline}=[->, thick,  opacity=1.0]
      \tikzstyle{cpmline}=[->, thick,  opacity=1.0]        
      \tikzstyle{nnNode}=[]
      \tikzstyle{cpmNode}=[]
      \tikzstyle{lossline}=[->, thick, dashed]
      \node[] (In1) {$\inputvars_{\lowmodel}$};
      \node[nnNode, above right = 0.25cm and 0.3cm of In1] (n11) {$\hiddenneuron_{11}$};
      \node[nnNode, below= 0.05cm of n11] (n21) {$\hiddenneuron_{21}$};
      \node[nnNode, below= 0.3cm of n11] (n31) {$\large \vdots$};
      \node[nnNode, below= 0.95cm of n11] (nm1) {$\hiddenneuron_{d1}$};
      
      \node[nnNode, right = 0.5cm of n11] (n12) {$\hiddenneuron_{12}$};
      \node[nnNode, below right = 0.05cm and 0.5cm of n11] (n22) {$\hiddenneuron_{22}$};
      \node[nnNode, below right= 0.3cm and 0.65cm  of n11] (n32) {$\large \vdots$};
      \node[nnNode, below right= 0.95cm and 0.5cm  of n11] (nm2) {$\hiddenneuron_{d2}$};
      
      \node[nnNode, right = 1.25cm of n11] (n13) {$\large \hdots$};
      \node[nnNode, below right = 0.1cm and 1.25cm of n11] (n23) {$\large \hdots$};
      \node[nnNode, below right = 1cm and 1.25cm of n11] (n23) {$\large \hdots$};
      
      \node[nnNode, right = 2cm of n11] (n14) {$\hiddenneuron_{1l}$};
      \node[nnNode, below right = 0.05cm and 2cm of n11] (n24) {$\hiddenneuron_{2l}$};
      \node[nnNode, below right= 0.3cm and 2.15cm  of n11] (n34) {$\large \vdots$};
      \node[nnNode, below right= 0.95cm and 2cm  of n11] (nm4) {$\hiddenneuron_{dl}$};
      
      \node[nnNode, right = 4.5cm of In1] (Out1) {$\outputvars_{\lowmodel}$};
      
      \draw[nnline] (In1.east) -- (n11.west);
      \draw[nnline] (In1.east) -- (n21.west);
      \draw[nnline] (In1.east) -- (nm1.west);
      
      \draw[nnline] (n11.east) -- (n12.west);
      \draw[nnline] (n11.east) -- (n22.west);
      \draw[nnline] (n11.east) -- (nm2.west);
      
      \draw[nnline] (n21.east) -- (n12.west);
      \draw[nnline] (n21.east) -- (n22.west);
      \draw[nnline] (n21.east) -- (nm2.west);
      
      \draw[nnline] (nm1.east) -- (n12.west);
      \draw[nnline] (nm1.east) -- (n22.west);
      \draw[nnline] (nm1.east) -- (nm2.west);
      
      \draw[nnline] (n14.east) -- (Out1);
      \draw[nnline] (n24.east) -- (Out1);
      \draw[nnline] (nm4.east) -- (Out1);
        \end{tikzpicture}
  }}
  
The interpretable model $\highmodel$ will often also have rich internal structure---such as a decision tree model---which one could naturally interpret as causal. For instance: 

     \centerline{\resizebox{0.4\textwidth}{!}
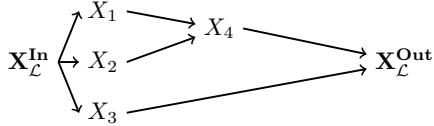
{
        \begin{tikzpicture}[node distance = 2.0cm]
     \tikzstyle{nnline}=[->, thick,  opacity=1.0]
      \tikzstyle{cpmline}=[->, thick,  opacity=1.0]        
      \tikzstyle{nnNode}=[]
      \tikzstyle{cpmNode}=[]
      \tikzstyle{lossline}=[->, thick, dashed]
      \node[] (In1) {$\inputvars_{\highmodel}$};
      \node[nnNode, above right = 0.15cm and 0.3cm of In1] (a1) {$\onevar_{1}$};
      \node[nnNode, right= 0.3cm of In1] (a2) {$\onevar_{2}$};
      \node[nnNode, below right = 0.15cm and 0.3cm of In1] (a3) {$\onevar_{3}$};
      
      \node[nnNode, above right = -0.1cm and 2cm of In1] (a4) {$\onevar_{4}$};
      
      \node[nnNode, right = 4.5cm of In1] (Out1) {$\outputvars_{\highmodel}$};
      
      \draw[nnline] (In1.east) -- (a1.west);
      \draw[nnline] (In1.east) -- (a2.west);
      \draw[nnline] (In1.east) -- (a3.west);
      
      \draw[nnline] (a1.east) -- (a4);
      \draw[nnline] (a2.east) -- (a4);
      
      \draw[nnline] (a4.east) -- (Out1);
      \draw[nnline] (a3.east) -- (Out1);
      
        \end{tikzpicture}
  }}

However, LIME only seeks to find a correspondence between the input--output behaviors of the interpretable and uninterpretable models. Therefore, representing both $\lowmodel$ and $\highmodel$ as a causal models connecting inputs to outputs is sufficient to describe the fidelity measure in LIME.
To shape approximate transformation to mirror the LIME fidelity metric, define 
\begin{enumerate}
    \item  The similarity between a low-level and high-level total state to be \[\simfunc(\tset_{\lowmodel}, \tset_{\highmodel}) = \mathsf{Distance}(\project{\tset_{\lowmodel}}{\outputvars_{\lowmodel}}, \project{\tset_{\highmodel}}{\outputvars_{\highmodel}})\]
 \item The probability distribution $\distribution$ to assign equal probability mass to input interventions in $\Delta_{\pset}$ and zero mass to all other interventions. 
 \item The statistic $\statistic$ to compute the expected value of a random variable. 
\end{enumerate}
 The LIME fidelity metric is the approximate transformation metric (Def. \ref{def:approx}) with $\tau$ and $\omega$ as identity functions:
\[\mathsf{LIME}(\highmodel,\lowmodel, \Delta_{\pset}) = 
 \mathbb{S}_{\intal\sim \distribution}[\simfunc\big( \setmap(\sols(\lowmodel_{\intal}) ), \sols(\highmodel_{\intmap(\intal)} )\big) ]\]
 
    \centerline{ \resizebox{0.4\textwidth}{!}
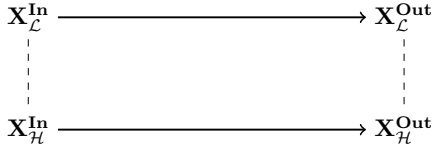
{
        \begin{tikzpicture}[node distance = 2.0cm]
     \tikzstyle{nnline}=[->, thick,  opacity=1.0]
      \tikzstyle{cpmline}=[-, dashed,  opacity=1.0]        
      \tikzstyle{nnNode}=[]
      \tikzstyle{cpmNode}=[]
      \tikzstyle{lossline}=[->, thick, dashed]
      \node[] (In1) {$\inputvars_{\lowmodel}$};
      
      \node[nnNode, right = 4.5cm of In1] (Out1) {$\outputvars_{\lowmodel}$};
      
      \node[below = 1cm of In1] (In2) {$\inputvars_{\highmodel}$};
      
      \node[nnNode, right = 4.5cm of In2] (Out2) {$\outputvars_{\highmodel}$};
      
      \draw[nnline] (In1.east) -- (Out1.west);
      \draw[nnline] (In2.east) -- (Out2.west);
      \draw[cpmline] (In1.south) -- (In2.north);
      \draw[cpmline] (Out1.south) -- (Out2.north);
        \end{tikzpicture}
  }}

\subsubsection{Single Source Interchange Interventions from Integrated Gradients}

Integrated gradients \citep{sundararajan2017axiomatic} computes the impact of neurons on model predictions. Following \cite{geiger2021causal}, we can easily translate the original integrated gradients equation into our causal model formalism.

\begin{definition}[Integrated Gradients]
Given a neural network as a causal model $\model$, we define the integrated gradient value of the $i$th neuron of an hidden vector $\varrs$ when the network is provided $\pset$ as an input as follows, where $\psett'$ is the so-called baseline value of $\varrs$:
\[\mathsf{IG}_i(\psett, \psett') = (\onevall_i - \onevall'_i) \cdot \int_{\delta = 0}^1 \frac{\partial \project{\model_{\pset\cup (\delta\psett+(1-\delta)\psett')}}{\outputvars}}{\partial \onevall_i} d\delta\]
The completeness axiom of the integrated gradients method is formulated as follows:
\[\sum_{i=1}^{|\varrs|} \mathsf{IG}_i(\psett, \psett') = \project{\model_{\pset \cup \psett}}{\outputvars} - \project{\model_{\pset \cup \psett'}}{ \outputvars}\]
\end{definition}

Integrated gradients was not initially conceived as a method for the causal analysis of neural networks. Therefore, it is perhaps surprising that integrated gradients can be used to compute interchange interventions. This hinges on a strategic use of the `baseline' value of integrated gradients; typically, the baseline value is set to be the zero vector, but here we set it to an interchange intervention. 
\begin{remark}[Integrated Gradients Can Compute Interventions]

The following is an immediate consequence of the completeness axiom
\[\project{\model_{\pset \cup \psett'}}{ \outputvars} = \project{\model_{\pset}}{\outputvars} - \sum_{i=1}^{|\varrs|} \mathsf{IG}_i\big(\project{\model_{\pset}}{ \varrs}, \psett'\big) \]
\end{remark}
In principle, we could perform causal abstraction analysis using the integrated gradients method and taking 
\[\psett' = \intinv(\model, \langle \pset' \rangle, \langle \varrs\rangle)\]
However, computing integrals is an inefficient way to compute interchange interventions.

\subsubsection{Estimating the Causal Effect of Real-World Concepts}

The ultimate downstream goal of explainable AI is to provide explanations with intuitive concepts \citep{Goyal, CausalLM, Elazar:2022, abraham-etal-2022-cebab} that are easily understood by human decision makers and guide their actions \citep{Karimi2021, Karimi2023,pmlr-v177-beckers22a}. These concepts can be abstract and mathematical, such as truth-valued propositional content, natural numbers, or real valued quantities like height or weight; they can also be grounded and concrete, such as the breed of a dog, the education level of a job applicant, or the pitch of a singer's voice. A basic question is how to estimate the effect of real-world concepts on the behavior of AI models.

The explainable AI benchmark CEBaB \citep{abraham-etal-2022-cebab} evaluates methods on their ability to estimate the causal effects of the quality of food, service, ambiance, and noise in a real-world dining experience on the prediction of a sentiment classifier, given a restaurant review as input data. Using CEBaB as an illustrative example, we represent the real-world data generating process and the neural network with a single causal model $\model_{\text{CEBaB}}$.\footnote{The models in \cite{abraham-etal-2022-cebab} are \textit{probabilistic} models, but we simplify to the deterministic case.}
The real-world concepts $C_{\text{service}}$, $C_{\text{noise}}$, $C_{\text{food}}$, and $C_{\text{ambiance}}$ can take on three values $+$, $-$, and $\textsf{Unknown}$, the input data $\inputvars$ takes on the value of a restaurant review text, the prediction output $\outputvars$ takes on the value of a five star rating, and hidden vectors $\hiddenneuron_{ij}$ can take on real number values. 

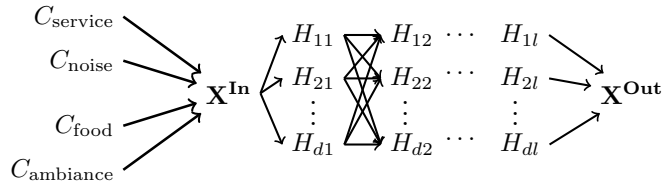
     \centerline{\resizebox{0.6\textwidth}{!}{
        \begin{tikzpicture}[node distance = 2.0cm]
     \tikzstyle{nnline}=[->, thick,  opacity=1.0]
      \tikzstyle{cpmline}=[->, thick,  opacity=1.0]        
      \tikzstyle{nnNode}=[]
      \tikzstyle{cpmNode}=[]
      \tikzstyle{lossline}=[->, thick, dashed]
      \node[] (In1) {$\inputvars$};
      \node[nnNode, above right = 0.25cm and 0.3cm of In1] (n11) {$\hiddenneuron_{11}$};
      \node[nnNode, below= 0.05cm of n11] (n21) {$\hiddenneuron_{21}$};
      \node[nnNode, below= 0.3cm of n11] (n31) {$\large \vdots$};
      \node[nnNode, below= 0.95cm of n11] (nm1) {$\hiddenneuron_{d1}$};
      
      \node[nnNode, right = 0.5cm of n11] (n12) {$\hiddenneuron_{12}$};
      \node[nnNode, below right = 0.05cm and 0.5cm of n11] (n22) {$\hiddenneuron_{22}$};
      \node[nnNode, below right= 0.3cm and 0.65cm  of n11] (n32) {$\large \vdots$};
      \node[nnNode, below right= 0.95cm and 0.5cm  of n11] (nm2) {$\hiddenneuron_{d2}$};
      
      \node[nnNode, right = 1.25cm of n11] (n13) {$\large \hdots$};
      \node[nnNode, below right = 0.1cm and 1.25cm of n11] (n23) {$\large \hdots$};
      \node[nnNode, below right = 1cm and 1.25cm of n11] (n23) {$\large \hdots$};
      
      \node[nnNode, right = 2cm of n11] (n14) {$\hiddenneuron_{1l}$};
      \node[nnNode, below right = 0.05cm and 2cm of n11] (n24) {$\hiddenneuron_{2l}$};
      \node[nnNode, below right= 0.3cm and 2.15cm  of n11] (n34) {$\large \vdots$};
      \node[nnNode, below right= 0.95cm and 2cm  of n11] (nm4) {$\hiddenneuron_{dl}$};
      
      \node[nnNode, right = 4.5cm of In1] (Out1) {$\outputvars$};
      
      \draw[nnline] (In1.east) -- (n11.west);
      \draw[nnline] (In1.east) -- (n21.west);
      \draw[nnline] (In1.east) -- (nm1.west);
      
      \draw[nnline] (n11.east) -- (n12.west);
      \draw[nnline] (n11.east) -- (n22.west);
      \draw[nnline] (n11.east) -- (nm2.west);
      
      \draw[nnline] (n21.east) -- (n12.west);
      \draw[nnline] (n21.east) -- (n22.west);
      \draw[nnline] (n21.east) -- (nm2.west);
      
      \draw[nnline] (nm1.east) -- (n12.west);
      \draw[nnline] (nm1.east) -- (n22.west);
      \draw[nnline] (nm1.east) -- (nm2.west);
      
      \draw[nnline] (n14.east) -- (Out1);
      \draw[nnline] (n24.east) -- (Out1);
      \draw[nnline] (nm4.east) -- (Out1);
            
            \node[above left =-0.1cm and 1cm of In1] (C2) {$C_{\text{noise}}$};
            \node[below left =-0.1cm and 1cm of In1] (C1) {$C_{\text{food}}$};
            \node[above left =0.5cm and 1cm of In1] (C3) {$C_{\text{service}}$};
            \node[below left =0.5cm and 1cm of In1] (Ck) {$C_{\text{ambiance}}$};

            \draw[->, line width = 1] (C1.east) -- (In1);
            \draw[->, line width = 1] (C2.east) -- (In1);
            \draw[->, line width = 1] (C3.east) -- (In1);
            \draw[->, line width = 1] (Ck.east) -- (In1);
        \end{tikzpicture}
  }}
  
If we are interested in the causal effect of food quality on model output, then we can marginalize away every variable other than the real-world concept $C_{\text{food}}$ and the neural network output $\outputvars$ to get a causal model with two variables. This marginalized causal model is a high-level abstraction of $\model_{\text{CEBaB}}$ that contains a single causal mechanism describing how food quality in a dining experience affects the neural network output. 

     \centerline{\resizebox{0.5\textwidth}{!}{
        \begin{tikzpicture}[node distance = 2.0cm]
     \tikzstyle{nnline}=[->, thick,  opacity=1.0]
      \tikzstyle{cpmline}=[->, thick,  opacity=1.0]        
      \tikzstyle{nnNode}=[]
      \tikzstyle{cpmNode}=[]
      \tikzstyle{lossline}=[->, thick, dashed]

      \node[nnNode, right = 4.5cm of In1] (Out1) {$\outputvars$};

            \node[above left =-0.1cm and 1cm of In1] (C2) {};
            \node[below left =-0.1cm and 1cm of In1] (C1) {$C_{\text{food}}$};
            \node[above left =0.5cm and 1cm of In1] (C3) {};
            \node[below left =0.5cm and 1cm of In1] (Ck) {};

            \draw[->, line width = 1] (C1.east) -- (Out1);

        \end{tikzpicture}
  }}

\subsection{Patching Activations with Interchange Interventions}\label{sec:abstractionwithinterchange}
There is a diverse body of mechanistic interpretability literature in which interchange interventions (see Section~\ref{sec:interchange}) are used to analyze neural networks (see Table~\ref{tab:methods}). However, the terminology used within this literature is often inconsistent and can lead to confusion regarding the precise techniques being employed. What is called an `activation patch' is typically equivalent to an interchange intervention on a neural network, but the term is sometimes used to describe a variety of other intervention techniques. \cite{wang2023interpretability} use `activation patching' to mean (recursive) interchange interventions, while \cite{automated_discovery}, \citet{zhang2024towards}, and \cite{heimersheim2024use} include ablation interventions under this heading (see Section~\ref{sec:ablation}), and \cite{patchscope} include arbitrary transformations that are more akin to activation steering (see Section~\ref{sec:activationsteering}). We propose to use `activation patch' to refer broadly to interventions on hidden vectors in neural networks, while using `interchange intervention' to pick out a specific type of intervention on causal models, which may be neural networks in some cases.

\subsubsection{Causal Mediation as Abstraction by a Three-Variable Chain} \label{sec:mediation}
\cite{vig2020causal},  \cite{finlayson-etal-2021-causal}, \cite{Meng:2022}, and \cite{Stolfo23} apply the popular causal inference framework of mediation analysis \citep{mediation1, mediation2} to understand how internal model components of neural networks mediate the causal effect of inputs on outputs. It is straightforward to show that mediation analysis is a special case of causal abstraction analysis. Mediation analysis is compatible with both ablation interventions (see Section~\ref{sec:ablation}) and interchange interventions. In this section, we present mediation analysis with interchange interventions.

Suppose that changing the value of variables $\vars$ from $\pset$ to $\pset'$ has an effect on a second set of variables~$\varrs$. Causal mediation analysis determines how this causal effect is mediated by a third set of intermediate variables $\varrrs$. The fundamental notions involved in mediation are total, direct, and indirect effects, which can be defined with interchange interventions.

\begin{definition}[Total, Direct, and Indirect Effects]

Consider a causal model $\model$ with disjoint sets of variables $\vars, \varrs, \varrrs \subset \allvars$ such that addition and subtraction are well-defined on values of $\varrs$. The \textit{total causal effect} of changing the values of $\vars$ from $\pset$ to $\mathbf{x'}$ on $\varrs$ is 
\[\mathsf{TotalEffect}(\model,\pset, \pset', \varrs) = \project{\sols(\model_{\pset'})}{ \varrs} - \project{\sols(\model_{\pset})}{ \varrs}\]
The \textit{direct causal effect} of changing $\vars$ from $\pset$ to $\pset'$ on $\varrs$ around mediator $\varrrs$ is 
\[\mathsf{DirectEffect}(\model,\pset, \pset', \varrs, \varrrs) = \project{\sols(\model_{\pset' \cup \mathsf{IntInv}(\model, \langle \pset \rangle, \langle \varrrs \rangle)})}{ \varrs} - \project{\sols(\model_{\pset})}{ \varrs}\]
The \textit{indirect causal effect} of changing $\vars$ from $\pset$ to $\pset'$ on $\varrs$ through mediator $\varrrs$ is 
\[\mathsf{IndirectEffect}(\model,\pset, \pset', \varrs, \varrrs) = \project{\sols(\model_{\pset \cup \mathsf{IntInv}(\model, \langle \pset' \rangle, \langle \varrrs \rangle)})}{\varrs} - \project{\sols(\model_{\pset})}{ \varrs}\]
\end{definition}

This method has been applied to the analysis of neural networks to characterize how the causal effect of inputs on outputs is mediated by (parts of) hidden vectors, with the goal identifying complete mediators. This is equivalent to a simple causal abstraction analysis.

\begin{remark}
Consider a neural network $\lowmodel$ with inputs $\inputvars$, outputs $\outputvars$, and hidden vector $\hiddenneurons$. 
Define $\cellpart_\onevar = \inputvars$, $\cellpart_\onevarr = \outputvars$, $\cellpart_{\onevarrr} = \hiddenneurons$, and $\cellpart_{\bot} = \allvars \setminus (\inputvars \cup \hiddenneurons \cup \outputvars)$. Apply variable merge and marginalization to $\lowmodel$ with $\cellpart$ to obtain a high-level model $\highmodel$ that is an abstraction of $\lowmodel$ with $\tau$ and $\omega$ as identity functions. The following are equivalent:
\begin{enumerate}
    \item The hidden vector $\hiddenneurons$ completely mediate the causal effect of inputs on outputs:
\[\mathsf{IndirectEffect}(\lowmodel,\pset, \pset', \outputvars, \hiddenneurons) = \mathsf{TotalEffect}(\model,\pset, \pset', \outputvars)\]
\item $\highmodel$ has a structure such that $\onevar$ is not a child of $\onevarr$:

\centerline{
\begin{tikzpicture}
\node[] (x) at (0,0) {$\onevar$} ;
\node[] (y) at (1.5,0) {$\onevarrr$} ;
\node[] (z) at (3,0) {$\onevarr$} ;
\draw[->] (x) -- (y);
\draw[->] (y) -- (z);
\end{tikzpicture}}
\end{enumerate}
\end{remark}

\subsubsection{Path Patching as Recursive Interchange Interventions}

Path patching \citep{wang2023interpretability, goldowskydill2023localizing, Hanna:2023, zhang2024towards, prakash2024finetuning} is a type of interchange intervention analysis %
that targets the connections between variables rather than variables themselves. To perform a path patch, we use a recursive interchange intervention on a model $\model$ processing a base input $\base$ that simulates `sender' variables $\hiddenneurons$ taking on intervened values from a source input $\source$, restricting the effect of this intervention to receiver variables $\mathbf{R}$ while freezing variables $\mathbf{F}$. 

Each receiver variable takes on a value determined by the $\source$ input for the sender variables $\hiddenneurons$ while fixing $\mathbf{F}$ to be the value determined by the $\base$ input. For receiver variable $R \in \mathbf{R}$, define an interchange intervention: 
\[\intpset_{R} = \intinv(\model, \langle \source, \base \rangle, \langle\hiddenneurons, \mathbf{F} \setminus \{R\}\rangle)\]

The path patch is a recursive interchange intervention resulting from the receiver variables $\{R_1, \dots, R_k\} = \mathbf{R}$ taking on the value determined by each of the basic interchange interventions:
\[\intpsett = \rintinv(\model, \langle \base, \dots, \base \rangle, \langle R_1,\dots, R_2 \rangle, \langle \intpset_{R_1}, \dots, \intpset_{R_k}\rangle)\]
The intervened model $\model_{\base \cup \intpsett}$ has a patched path according to the definition of \cite{wang2023interpretability}. Simpler path patching experiments will not freeze any variables, meaning $\mathbf{F} = \varnothing$. We show a visualization below, where $\hiddenneurons = \{H_{11}\}$ is the sender neuron, $\mathbf{R} = \{H_{32}\}$ is the receiver neuron, and  $\mathbf{F} = \{H_{22}\}$ is a neuron we would like to keep frozen (meaning it has no effect in computing the value of the patched receiver node).

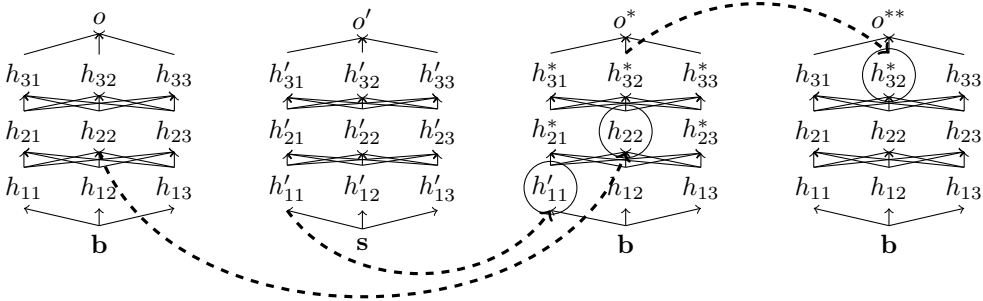
\centerline{
\begin{tikzpicture}
\tikzstyle{nnline}=[->, opacity=0.5]
\tikzstyle{intline}=[->, very thick, dashed]

\node[] (b) at (0, 0) {$\base$} ;
\node[] (h11) at (-1, 0.75) {$\hiddenneuronval_{11}$} ;
\node[] (h12) at (0, 0.75) {$\hiddenneuronval_{12}$} ;
\node[] (h13) at (1, 0.75) {$\hiddenneuronval_{13}$} ;
\node[] (h21) at (-1, 1.5) {$\hiddenneuronval_{21}$} ;
\node[] (f_s) at (0, 1.5) {$\hiddenneuronval_{22}$} ;
\node[] (h23) at (1, 1.5) {$\hiddenneuronval_{23}$} ;
\node[] (h31) at (-1, 2.25) {$\hiddenneuronval_{31}$} ;
\node[] (h32) at (0, 2.25) {$\hiddenneuronval_{32}$} ;
\node[] (h33) at (1, 2.25) {$\hiddenneuronval_{33}$} ;
\node[] (out) at (0, 3) {$\outneuronval$} ;

\draw[nnline] (b.north) -- (h11.south);
\draw[nnline] (b.north) -- (h12.south);
\draw[nnline] (b.north) -- (h13.south);

\draw[nnline] (h11.north) -- (h21.south);
\draw[nnline] (h11.north) -- (f_s.south);
\draw[nnline] (h11.north) -- (h23.south);
\draw[nnline] (h12.north) -- (h21.south);
\draw[nnline] (h12.north) -- (f_s.south);
\draw[nnline] (h12.north) -- (h23.south);
\draw[nnline] (h13.north) -- (h21.south);
\draw[nnline] (h13.north) -- (f_s.south);
\draw[nnline] (h13.north) -- (h23.south);

\draw[nnline] (h21.north) -- (h31.south);
\draw[nnline] (h21.north) -- (h32.south);
\draw[nnline] (h21.north) -- (h33.south);
\draw[nnline] (f_s.north) -- (h31.south);
\draw[nnline] (f_s.north) -- (h32.south);
\draw[nnline] (f_s.north) -- (h33.south);
\draw[nnline] (h23.north) -- (h31.south);
\draw[nnline] (h23.north) -- (h32.south);
\draw[nnline] (h23.north) -- (h33.south);

\draw[nnline] (h31.north) -- (out.south);
\draw[nnline] (h32.north) -- (out.south);
\draw[nnline] (h33.north) -- (out.south);

\node[] (s) at (3.5, 0) {$\source$} ;
\node[] (h_s) at (2.5, 0.75) {$\hiddenneuronval_{11}'$} ;
\node[] (h12) at (3.5, 0.75) {$\hiddenneuronval_{12}'$} ;
\node[] (h13) at (4.5, 0.75) {$\hiddenneuronval_{13}'$} ;
\node[] (h21) at (2.5, 1.5) {$\hiddenneuronval_{21}'$} ;
\node[] (h22) at (3.5, 1.5) {$\hiddenneuronval_{22}'$} ;
\node[] (h23) at (4.5, 1.5) {$\hiddenneuronval_{23}'$} ;
\node[] (h31) at (2.5, 2.25) {$\hiddenneuronval_{31}'$} ;
\node[] (h32) at (3.5, 2.25) {$\hiddenneuronval_{32}'$} ;
\node[] (h33) at (4.5, 2.25) {$\hiddenneuronval_{33}'$} ;
\node[] (out) at (3.5, 3) {$\outneuronval'$} ;

\draw[nnline] (s.north) -- (h_s.south);
\draw[nnline] (s.north) -- (h12.south);
\draw[nnline] (s.north) -- (h13.south);

\draw[nnline] (h_s.north) -- (h21.south);
\draw[nnline] (h_s.north) -- (h22.south);
\draw[nnline] (h_s.north) -- (h23.south);
\draw[nnline] (h12.north) -- (h21.south);
\draw[nnline] (h12.north) -- (h22.south);
\draw[nnline] (h12.north) -- (h23.south);
\draw[nnline] (h13.north) -- (h21.south);
\draw[nnline] (h13.north) -- (h22.south);
\draw[nnline] (h13.north) -- (h23.south);

\draw[nnline] (h21.north) -- (h31.south);
\draw[nnline] (h21.north) -- (h32.south);
\draw[nnline] (h21.north) -- (h33.south);
\draw[nnline] (h22.north) -- (h31.south);
\draw[nnline] (h22.north) -- (h32.south);
\draw[nnline] (h22.north) -- (h33.south);
\draw[nnline] (h23.north) -- (h31.south);
\draw[nnline] (h23.north) -- (h32.south);
\draw[nnline] (h23.north) -- (h33.south);

\draw[nnline] (h31.north) -- (out.south);
\draw[nnline] (h32.north) -- (out.south);
\draw[nnline] (h33.north) -- (out.south);

\node[] (b) at (7, 0) {$\base$} ;
\node[] (h_t) at (6, 0.75) {$\hiddenneuronval_{11}'$} ;
\node[] (h12) at (7, 0.75) {$\hiddenneuronval_{12}$} ;
\node[] (h13) at (8, 0.75) {$\hiddenneuronval_{13}$} ;
\node[] (h21) at (6, 1.5) {$\hiddenneuronval_{21}^*$} ;
\node[] (f_t) at (7, 1.5) {$\hiddenneuronval_{22}$} ;
\node[] (h23) at (8, 1.5) {$\hiddenneuronval_{23}^*$} ;
\node[] (h31) at (6, 2.25) {$\hiddenneuronval_{31}^*$} ;
\node[] (r_s) at (7, 2.25) {$\hiddenneuronval_{32}^*$} ;
\node[] (h33) at (8, 2.25) {$\hiddenneuronval_{33}^*$} ;
\node[] (out) at (7, 3) {$\outneuronval^*$} ;

\draw[nnline] (b.north) -- (h_t.south);
\draw[nnline] (b.north) -- (h12.south);
\draw[nnline] (b.north) -- (h13.south);

\draw[nnline] (h_t.north) -- (h21.south);
\draw[nnline] (h_t.north) -- (f_t.south);
\draw[nnline] (h_t.north) -- (h23.south);
\draw[nnline] (h12.north) -- (h21.south);
\draw[nnline] (h12.north) -- (f_t.south);
\draw[nnline] (h12.north) -- (h23.south);
\draw[nnline] (h13.north) -- (h21.south);
\draw[nnline] (h13.north) -- (f_t.south);
\draw[nnline] (h13.north) -- (h23.south);

\draw[nnline] (h21.north) -- (h31.south);
\draw[nnline] (h21.north) -- (r_s.south);
\draw[nnline] (h21.north) -- (h33.south);
\draw[nnline] (f_t.north) -- (h31.south);
\draw[nnline] (f_t.north) -- (r_s.south);
\draw[nnline] (f_t.north) -- (h33.south);
\draw[nnline] (h23.north) -- (h31.south);
\draw[nnline] (h23.north) -- (r_s.south);
\draw[nnline] (h23.north) -- (h33.south);

\draw[nnline] (h31.north) -- (out.south);
\draw[nnline] (r_s.north) -- (out.south);
\draw[nnline] (h33.north) -- (out.south);

\node[draw, ellipse, minimum width=20pt, minimum height=20pt] at (6, 0.75) {};
\node[draw, ellipse, minimum width=20pt, minimum height=20pt] at (7, 1.5) {};

\node[] (b) at (10.5, 0) {$\base$} ;
\node[] (h11) at (9.5, 0.75) {$\hiddenneuronval_{11}$} ;
\node[] (h12) at (10.5, 0.75) {$\hiddenneuronval_{12}$} ;
\node[] (h13) at (11.5, 0.75) {$\hiddenneuronval_{13}$} ;
\node[] (h21) at (9.5, 1.5) {$\hiddenneuronval_{21}$} ;
\node[] (h22) at (10.5, 1.5) {$\hiddenneuronval_{22}$} ;
\node[] (h23) at (11.5, 1.5) {$\hiddenneuronval_{23}$} ;
\node[] (h31) at (9.5, 2.25) {$\hiddenneuronval_{31}$} ;
\node[] (r_t) at (10.5, 2.25) {$\hiddenneuronval_{32}^*$} ;
\node[] (h33) at (11.5, 2.25) {$\hiddenneuronval_{33}$} ;
\node[] (out) at (10.5, 3) {$\outneuronval^{**}$} ;

\draw[nnline] (b.north) -- (h11.south);
\draw[nnline] (b.north) -- (h12.south);
\draw[nnline] (b.north) -- (h13.south);

\draw[nnline] (h11.north) -- (h21.south);
\draw[nnline] (h11.north) -- (h22.south);
\draw[nnline] (h11.north) -- (h23.south);
\draw[nnline] (h12.north) -- (h21.south);
\draw[nnline] (h12.north) -- (h22.south);
\draw[nnline] (h12.north) -- (h23.south);
\draw[nnline] (h13.north) -- (h21.south);
\draw[nnline] (h13.north) -- (h22.south);
\draw[nnline] (h13.north) -- (h23.south);

\draw[nnline] (h21.north) -- (h31.south);
\draw[nnline] (h21.north) -- (r_t.south);
\draw[nnline] (h21.north) -- (h33.south);
\draw[nnline] (h22.north) -- (h31.south);
\draw[nnline] (h22.north) -- (r_t.south);
\draw[nnline] (h22.north) -- (h33.south);
\draw[nnline] (h23.north) -- (h31.south);
\draw[nnline] (h23.north) -- (r_t.south);
\draw[nnline] (h23.north) -- (h33.south);

\draw[nnline] (h31.north) -- (out.south);
\draw[nnline] (r_t.north) -- (out.south);
\draw[nnline] (h33.north) -- (out.south);

\node[draw, ellipse, minimum width=20pt, minimum height=20pt] at (10.5, 2.25) {};

\draw[intline] (f_s.south) to[out=-70, in=-110] (f_t.south);

\draw[intline] (h_s.south) to[out=-55, in=-125] (h_t.south);

\draw[intline] (r_s.north) to[out=40, in=140] (r_t.north);

\end{tikzpicture}
}

\subsection{Ablation as Abstraction by a Three Variable Collider}\label{sec:ablation}
Neuroscientific lesion studies involve damage to a region of the brain in order to determine its function; if the lesion results in a behavioral deficit, then the brain region is assumed to be involved in the production of that behavior. In mechanistic interpretability, such interventions are known as \textit{ablations}. Common ablations include replacing neural hidden vectors with zero activations \citep{cammarata2020thread,olsson2022context, geva2023dissecting} or with mean activations over a set of input data \citep{wang2023interpretability}, adding random noise to the activations (causal tracing; \citealt{Meng:2022, Meng23}), and replacing activations with the values from a different input (resample ablations; \citealt{chan2022causal}). To capture ablation studies as a special case of causal abstraction analysis, we only need a high-level model with an input variable, an output variable, and a binary valued variable aligned with the variables targeted for ablation. 

\subsubsection{Concept Erasure}
Concept erasure is a common application of ablations in which an hidden vector $\hiddenneurons$ in a neural network $\lowmodel$ is ablated in order to remove information about a particular concept $C$ \citep{Ravfogel:2020,linear_adversarial_concept_Eraser, kernelized_concept_Eraser,Ravfogel2023,Elazar-etal-2020, Lovering2022, olsson2022context, LEACE}. To quantify the success of a concept erasure experiment, each concept $C$ is associated with some degraded behavioral capability encoded as a partial function $\mathcal{A}_C:\values{\inputvars_{\lowmodel}} \to \values{\outputvars_{\lowmodel}}$ (e.g., ablating the concept `dog' would be associated with the behavior of inaccurately captioning images with dogs in them). If performing an ablation on $\hiddenneurons$ to erase the concept $C$ leads $\lowmodel$ to have the degraded behavior $\mathcal{A}_C$ without changing other behaviors, then the ablation was successful.

We can model ablation on $\lowmodel$ as abstraction by a three-variable causal model. Define a high-level signature to be an input variable $\onevar$ taking on values from $\inputvars_{\lowmodel}$, output variable $\onevarr$ taking on values from $\outputvars_{\lowmodel}$, and a binary variable $\onevarrr$ that indicates whether the concept $C$ has been erased. The mechanism for $\onevar$ assigns an arbitrary default input, the mechanism for $\onevarrr$ assigns $0$, and the mechanism for $\onevarr$ produces the degraded behavior if $\onevarrr$ is $1$, and mimics $\lowmodel$ otherwise:
\[ \mechanism{\onevarr}(\oneval, \onevalll) = \begin{cases}
\mathcal{A}_{C}(\oneval) & \onevalll = 1 \text{ and } \oneval \in \dom(\mathcal{A}_C)\\
\project{\sols(\lowmodel_{\oneval})}{\outputvars_{\lowmodel}} & \text{Otherwise}\\
\end{cases}.\]

The map $\setmap$ from low-level settings to the high-level settings simply sets $\onevarrr$ to be $0$ exactly when the low-level input determines the value of the model component $\hiddenneurons$:
\[\setmap(\tset) = \begin{cases} 
\{\project{\tset}{\inputvars}, 0,\project{\tset}{\outputvars} \} &  \project{\tset}{ \hiddenneurons} = \project{\sols(\lowmodel_{\project{\tset}{\inputvars}})}{\hiddenneurons} \\ 
\{\project{\tset}{\inputvars}, 1,\project{\tset}{\outputvars} \} & \text{Otherwise} \\
\end{cases}.\]

The function $\intmap$ is defined on low-level input interventions and the interventional $\intal$ that is an ablation on $\hiddenneurons$ (e.g., setting activations to zero or a mean value, or projecting activations onto a linear subspace whose complement is thought to encode the concept $C$). Low-level input interventions are mapped by $\intmap$ to identical high-level input interventions, while $\intal$ is mapped by $\intmap$ to the high-level intervention setting $\onevarrr$ to $1$.

The high-level causal model $\highmodel$ is an exact transformation of the low-level neural model $\lowmodel$ under $(\setmap, \intmap)$ exactly when the ablation removing the concept $C$ results in degraded behavior defined by $\mathcal{A}_{C}$. We show a visualization below, where $\hiddenneurons = \{\hiddenneuron_{12}, \hiddenneuron_{22} \}$.

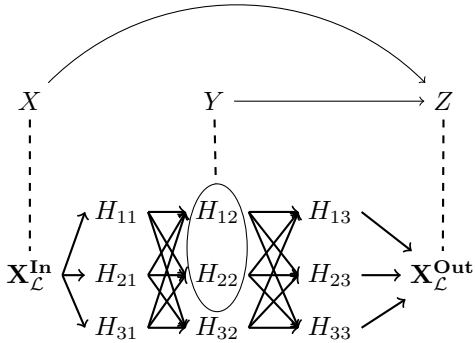
\centerline{
\begin{tikzpicture}
\node[] (x) at (0,0) {$\onevar$} ;
\node[] (y) at (2.45,0) {$\onevarrr$} ;
\node[] (z) at (5.5,0) {$\onevarr$} ;
\draw[->] (x) to[out=45, in=135] (z);
\draw[->] (y) -- (z);
     \tikzstyle{nnline}=[->, thick,  opacity=1.0]
      \tikzstyle{cpmline}=[->, thick,  opacity=1.0]        
      \tikzstyle{nnNode}=[]
      \tikzstyle{cpmNode}=[]
      \tikzstyle{lossline}=[thick, dashed]
      \node[below = 1.75cm of x] (In1) {$\inputvars_{\lowmodel}$};
      \node[nnNode, above right = 0.25cm and 0.3cm of In1] (n11) {$\hiddenneuron_{11}$};
      \node[nnNode, below= 0.3cm of n11] (n21) {$\hiddenneuron_{21}$};
      \node[nnNode, below= 1cm of n11] (nm1) {$\hiddenneuron_{31}$};
      
      \node[nnNode, right = 0.5cm of n11] (n12) {$\hiddenneuron_{12}$};
      
      \node[draw, ellipse, minimum width=0.8cm, minimum height=1.7cm, below right = -0.4cm and 0.63cm of n11] (pi) {};
      
      \node[nnNode, below right = 0.3cm and 0.5cm of n11] (n22) {$\hiddenneuron_{22}$};
      \node[nnNode, below right= 1cm and 0.5cm  of n11] (nm2) {$\hiddenneuron_{32}$};
      
      \node[nnNode, right = 2cm of n11] (n14) {$\hiddenneuron_{13}$};
      \node[nnNode, below right = 0.3cm and 2cm of n11] (n24) {$\hiddenneuron_{23}$};
      \node[nnNode, below right= 1cm and 2cm  of n11] (nm4) {$\hiddenneuron_{33}$};
      
      \node[nnNode, right = 4.5cm of In1] (Out1) {$\outputvars_{\lowmodel}$};
      
      \draw[nnline] (In1.east) -- (n11.west);
      \draw[nnline] (In1.east) -- (n21.west);
      \draw[nnline] (In1.east) -- (nm1.west);
      
      \draw[nnline] (n11.east) -- (n12.west);
      \draw[nnline] (n11.east) -- (n22.west);
      \draw[nnline] (n11.east) -- (nm2.west);
      
      \draw[nnline] (n21.east) -- (n12.west);
      \draw[nnline] (n21.east) -- (n22.west);
      \draw[nnline] (n21.east) -- (nm2.west);
      
      \draw[nnline] (nm1.east) -- (n12.west);
      \draw[nnline] (nm1.east) -- (n22.west);
      \draw[nnline] (nm1.east) -- (nm2.west);
      
      \draw[nnline] (n12.east) -- (n14.west);
      \draw[nnline] (n12.east) -- (n24.west);
      \draw[nnline] (n12.east) -- (nm4.west);
      
      \draw[nnline] (n22.east) -- (n14.west);
      \draw[nnline] (n22.east) -- (n24.west);
      \draw[nnline] (n22.east) -- (nm4.west);
      
      \draw[nnline] (nm2.east) -- (n14.west);
      \draw[nnline] (nm2.east) -- (n24.west);
      \draw[nnline] (nm2.east) -- (nm4.west);
      
      \draw[nnline] (n14.east) -- (Out1);
      \draw[nnline] (n24.east) -- (Out1);
      \draw[nnline] (nm4.east) -- (Out1);
    
    \draw[lossline] (x) -- (In1);
    \draw[lossline] (y) -- (pi);
    \draw[lossline] (z) -- (Out1);
    
        \end{tikzpicture}
  }

Observe that the high-level model $\highmodel$ does not have a variable encoding the concept $C$ and the values it might take on. Ablation studies attempt to determine \textit{whether} a concept is used by a model; they do not characterize \textit{how} that concept is used.

\subsubsection{Sub-Circuit Analysis}\label{sec:subcircuit}
Sub-circuit analysis \citep{sixteen_heads_better_than_one,lowcomplexity_probing, Csordas2021, cammarata2020thread,olsson2022context,chan2022causal,Lepori2023a,Lepori2023b,wang2023interpretability, automated_discovery} aims to identify a minimal circuit $\mathcal{C} \subseteq \allvars \times \allvars$ between components in a model $\lowmodel$ that is sufficient to perform particular behavior that we represent with a partial function $\mathcal{B}:\values{\inputvars_{\lowmodel}} \to \values{\outputvars_{\lowmodel}}$. This claim is cashed out in terms of ablations, specifically, the behavior $\mathcal{B}$ should remain intact when all connections between model components $\overline{\mathcal{C}} = \allvars \times \allvars \setminus \mathcal{C}$ are ablated. Define $\hiddenneurons = \{\hiddenneuron: \exists \hiddenneuronn(\hiddenneuronn, \hiddenneuron) \in \overline{\mathcal{C}}\}$ as the model components with incoming severed connections, $\hiddenneuronss = \{\hiddenneuronn: \exists \hiddenneuron (\hiddenneuronn, \hiddenneuron) \in \overline{\mathcal{C}}\}$ as the model components with outgoing severed connections, and let $\hiddenneuronsett$ be the ablation values.

We can model sub-circuit analysis on $\lowmodel$ as abstraction by a three-variable causal model. Define a high-level signature to be an input variable $\onevar$ taking on values from $\inputvars_{\lowmodel}$, output variable $\onevarr$ taking on values from $\outputvars_{\lowmodel} \cup \{\bot\}$, and a binary variable $\onevarrr$ that indicates whether the connections $\overline{\mathcal{C}}$ have been severed. The mechanism for $\onevar$ assigns an arbitrary default input, the mechanism for $\onevarrr$ assigns $0$, and the mechanism for $\onevarr$ mimics the behavior of $\lowmodel$ when $\onevarrr$ is $0$ and preserves only the behavior $\mathcal{B}$ when $\onevarrr$ is $1$:
\[ \mechanism{\onevarr}(\oneval, \onevalll) = \begin{cases}
\project{\lowmodel_{\oneval}}{\outputvars_{\lowmodel}} & \onevalll = 0\\
\mathcal{B}(\oneval) & \onevalll = 1 \text{ and } \oneval \in \dom(\mathcal{B})\\
\bot& \onevalll = 1 \text{ and } \oneval \not \in \dom(\mathcal{B})\\
\end{cases}\]

The map $\setmap$ from low-level settings to  high-level settings simply sets $\onevarrr$ to be $0$ exactly when the low-level input determines the value of the model components $\hiddenneurons$:
\[\setmap(\tset) = \begin{cases} 
\{\project{\tset}{\inputvars}, 0,\project{\tset}{\outputvars} \} &  \project{\tset}{\hiddenneurons} = \project{\sols(\lowmodel_{\project{\tset}{\inputvars}})}{ \hiddenneurons} \\ 
\{\project{\tset}{\inputvars}, 1,\project{\tset}{\outputvars} \} & \text{Otherwise} \\
\end{cases}\]

The function $\intmap$  is defined on low-level input interventions in $\dom(\mathcal{B})$ and the interventional $\intal$ that fixes the connections in $\overline{\mathcal{C}}$ to an ablated value and leaves the connections in $\mathcal{C}$ untouched:
\[\intal\langle\mechanism{\hiddenneurons}\rangle_{\hiddenneuron} =\tset \mapsto \mechanism{\hiddenneuron}\Big( 
\project{\tset}{\{\hiddenneuronn: (\hiddenneuronn, \hiddenneuron) \not \in \overline{\mathcal{C}}\}} \cup \bigcup_{\hiddenneuronn \in \{\hiddenneuronn: (\hiddenneuronn, \hiddenneuron) \in \overline{\mathcal{C}}\}}\project{\hiddenneuronsett}{\hiddenneuronn}  \Big)\]
Low-level input interventions are mapped by $\intmap$ to identical high-level input interventions and $\intal$ is mapped by $\intmap$ to the high-level intervention setting $\onevarrr$ to $1$.

The high-level causal model $\highmodel$ is an exact transformation of the low-level neural model $\lowmodel$ under $(\setmap, \intmap)$ exactly when the subcircuit $\mathcal{C}$ preserves the behavior $\mathcal{B}$. We show a visualization below, where:
\[\overline{\mathcal{C}} = \big(\inputvars \times \{\hiddenneuron_{31}\}\big) \cup \{(\hiddenneuron_{31}, \hiddenneuron_{12}),(\hiddenneuron_{31}, \hiddenneuron_{22}),(\hiddenneuron_{31}, \hiddenneuron_{32}), (\hiddenneuron_{22}, \hiddenneuron_{23}),(\hiddenneuron_{12}, \hiddenneuron_{23})\}\]

\centerline{
\begin{tikzpicture}
\node[] (x) at (0,0) {$\onevar$} ;
\node[] (y) at (2.45,0) {$\onevarrr$} ;
\node[] (z) at (5.5,0) {$\onevarr$} ;
\draw[->] (x) to[out=45, in=135] (z);
\draw[->] (y) -- (z);
     \tikzstyle{nnline}=[->, thick,  opacity=1.0]
     \tikzstyle{phantomnnline}=[->, thick,  opacity=0.1]
      \tikzstyle{cpmline}=[->, thick,  opacity=1.0]        
      \tikzstyle{nnNode}=[]
      \tikzstyle{cpmNode}=[]
      \tikzstyle{lossline}=[thick, dashed]
      \node[below = 1.75cm of x] (In1) {$\inputvars_{\lowmodel}$};
      \node[nnNode, above right = 0.25cm and 0.3cm of In1] (n11) {$\hiddenneuron_{11}$};
      \node[nnNode, below= 0.3cm of n11] (n21) {$\hiddenneuron_{21}$};
      \node[nnNode, below= 0.92cm of n11, draw, circle, inner sep = 0cm] (nm1) {$\hiddenneuron_{31}$};
      
      \node[nnNode, right = 0.5cm of n11] (n12) {$\hiddenneuron_{12}$};

      \node[nnNode, below right = 0.3cm and 0.5cm of n11] (n22) {$\hiddenneuron_{22}$};
      \node[nnNode, below right= 1cm and 0.5cm  of n11] (nm2) {$\hiddenneuron_{32}$};
      
      \node[nnNode, right = 2cm of n11] (n14) {$\hiddenneuron_{13}$};
      \node[nnNode, below right = 0.3cm and 2.18cm of n11, draw, circle, inner sep = 0cm] (n24) {$\hiddenneuron_{23}$};
      \node[nnNode, below right= 1cm and 2cm  of n11] (nm4) {$\hiddenneuron_{33}$};
      
      \node[nnNode, right = 4.5cm of In1] (Out1) {$\outputvars_{\lowmodel}$};
      
      \node[draw, ellipse, minimum width=0.8cm, minimum height=1.7cm, below right = -0.4cm and 0.63cm of n11] (pi1) {};
      
      \draw[nnline] (In1.east) -- (n11.west);
      \draw[nnline] (In1.east) -- (n21.west);
      \draw[phantomnnline] (In1.east) -- (nm1.west);
      
      \draw[nnline] (n11.east) -- (n12.west);
      \draw[phantomnnline] (n11.east) -- (n22.west);
      \draw[nnline] (n11.east) -- (nm2.west);
      
      \draw[phantomnnline] (n21.east) -- (n12.west);
      \draw[nnline] (n21.east) -- (n22.west);
      \draw[nnline] (n21.east) -- (nm2.west);
      
      \draw[phantomnnline] (nm1.east) -- (n12.west);
      \draw[phantomnnline] (nm1.east) -- (n22.west);
      \draw[phantomnnline] (nm1.east) -- (nm2.west);
      
      \draw[nnline] (n12.east) -- (n14.west);
      \draw[phantomnnline] (n12.east) -- (n24.west);
      \draw[nnline] (n12.east) -- (nm4.west);
      
      \draw[nnline] (n22.east) -- (n14.west);
      \draw[phantomnnline] (n22.east) -- (n24.west);
      \draw[nnline] (n22.east) -- (nm4.west);
      
      \draw[nnline] (nm2.east) -- (n14.west);
      \draw[nnline] (nm2.east) -- (n24.west);
      \draw[nnline] (nm2.east) -- (nm4.west);
      
      \draw[nnline] (n14.east) -- (Out1);
      \draw[nnline] (n24.east) -- (Out1);
      \draw[nnline] (nm4.east) -- (Out1);
    
    \draw[lossline] (x) -- (In1);
    \draw[lossline] (y) -- (pi);
    \draw[lossline] (y) -- (nm1);
    \draw[lossline] (y) -- (n24);
    \draw[lossline] (z) -- (Out1);
    
        \end{tikzpicture}
  }

\subsubsection{Causal Scrubbing}\label{sec:scrub}

Causal scrubbing \citep{chan2022causal} is an ablation method that proposes to determine whether a circuit $\mathcal{C}$ is sufficient for a behavior $\mathcal{B}$. It doesn't fit into our general paradigm of sub-circuit analysis for two reasons. First, the minimal circuit $\mathcal{C}$ is determined by a high-level causal model $\highmodel$ with identical input and output spaces as $\lowmodel$ and a surjective partial function $\delta: \allvars_{\lowmodel} \to \allvars_{\highmodel}$ assigning each low-level variable a high-level variable. Specifically, the low-level minimal circuit is the high-level causal graph pulled back into the low-level $\mathcal{C} = \{(\hiddenneuronn, \hiddenneuron): \delta(\hiddenneuronn) \prec \delta(\hiddenneuron)\}$. Second, the connections in the minimal circuit $\mathcal{C}$ are intervened upon in addition to connections in  $\overline{\mathcal{C}}$. This means that every single connection in the network is being intervened upon. 

Given a base input $\base$, causal scrubbing recursively intervenes on every connection in the network. The connections in $\overline{\mathcal{C}}$ are replaced using randomly sampled source inputs, and the connections $(\hiddenneuronn, \hiddenneuron) \in \mathcal{C}$ are replaced using randomly sampled source inputs that set $\delta(\hiddenneuron)$ to the same value in $\highmodel$. \cite{chan2022causal} call these interchange interventions---where the base and source input agree on a high-level variable---\textit{resampling ablations}. The exact intervention value for each targeted connection is determined by a recursive interchange intervention performed by calling the algorithm $\mathsf{Scrub}(\base,\outputvars_{\lowmodel})$ defined below.

    \begin{algorithm}[H]
    \small
    \caption{$\mathsf{Scrub}(\base, \hiddenneurons)$}
    $\hiddenneuronset \leftarrow \{\}$
    
        \For{$\hiddenneuron \in \hiddenneurons$}{
    \If{$\hiddenneuron \in \inputvars_{\lowmodel}$}{
    $\hiddenneuronset \leftarrow \hiddenneuronset \cup \project{\base}{\hiddenneuron}$
    
    \textbf{continue}
    } 

    $\hiddenneuronsett \leftarrow \{\}$
    
    \For{$\hiddenneuronn \in \{\hiddenneuronn: (\hiddenneuronn, \hiddenneuron) \not \in \mathcal{C} \}$}{
    $\source \sim \values{\inputvars_{\lowmodel}}$
    
    $\hiddenneuronsett \leftarrow \hiddenneuronsett \cup \mathsf{Scrub}(\source, \{\hiddenneuronn\})$
    }
    
    \If{$\hiddenneuron \in \dom(\delta)$}{
    
    $\source \sim \{\source \in \values{\inputvars_{\lowmodel}}: \project{\sols(\highmodel_{\source})}{\delta(H)} = \project{\sols(\highmodel_{\base})}{\delta(H)}\}$
    
    $\hiddenneuronsett \leftarrow \hiddenneuronsett \cup \mathsf{Scrub}(\source, \{\hiddenneuronn:  (\hiddenneuronn, \hiddenneuron) \in \mathcal{C} \})$
    }
    $\hiddenneuronset \leftarrow \hiddenneuronset \cup \project{\sols(\lowmodel_{\hiddenneuronsett})}{\hiddenneuron}$
    }
    \Return $\hiddenneuronset$
    \end{algorithm}
    
While causal scrubbing makes use of a high-level model $\mathcal{H}$, the only use of this model in the algorithm $\mathsf{Scrub}$ is to sample source inputs $\source$ that assign the same value to a variable as a base input $\base$. No interventions are performed on the high-level model and so no correspondences between high-level and low-level interventions are established. This means that we can still model causal scrubbing as abstraction by the same three-variable causal model we defined in Section~\ref{sec:subcircuit}, which we will name $\mathcal{H}^*$. The map $\setmap$ from low-level settings to the high-level settings simply sets $\onevarr$ to be $0$ exactly when the low-level input determines the value of the total setting
\[\setmap(\tset) = \begin{cases} 
\{\project{\tset}{\inputvars}, 0,\project{\tset}{\outputvars} \} &   \tset = \sols(\lowmodel_{\project{\tset}{\inputvars}}) \\ 
\{\project{\tset}{\inputvars}, 1,\project{\tset}{\outputvars} \} & \text{Otherwise} \\
\end{cases}\]

The function $\intmap$ is defined to map low-level input interventions to identical high-level input interventions and map any interventional $\intal$ resulting from a call to $\mathsf{Scrub}$ to the high-level intervention setting $Z$ to $1$.

\cite{chan2022causal} propose to measure the faithfulness of $\mathcal{H}$ by appeal to the proportion of performance maintained by $\mathcal{L}$ under interventionals determined by $\mathsf{Scrub}$. This is equivalent to the approximate transformation metric for high-level causal model $\highmodel^*$ and low-level causal model $\lowmodel$ under $(\setmap, \intmap)$ (Definition~\ref{def:approx}) with $\distribution$ as a random distribution over inputs and interventionals from $\mathsf{Scrub}$, $\simfunc$ as a function outputting $1$ only if the low-level and high-level outputs are equal, and $\statistic$ as expected value.

\subsection{Modular Feature Learning as Bijective Transformation}\label{sec:localization}

A core task of mechanistic interpretability is disentangling a vector of activations into a set of modular features that correspond to human-intelligible concepts. We construe modular feature learning as constructing a bijective translation (Def.~\ref{def:translation}). Some methods use a high-level causal model as a source of supervision in order to construct modular features that localize the concepts encoded in the high-level intermediate variables. Other methods are entirely unsupervised and produce modular features that must be further analyzed to determine the concepts they might encode. Formalizing modular feature learning as bijective transformation provides a unified framework to evaluate commonly used mechanistic interpretability methods through distributed interchange interventions (see \citealt{huang2024ravel} for a mechanistic interpretability benchmark based in this framework).

\subsubsection{Unsupervised Methods}

\paragraph{Principal Component Analysis}

Principal Component Analysis (PCA) is a technique that represents high-dimensional data in a lower-dimensional space while maximally preserving information in the original data. PCA has been used to identify subspaces related to human-interpretable concepts, e.g., gender \citep{bolukbasi2016man}, sentiment \citep{tigges2023linear}, truthfulness \citep{marks2023geometry}, and visual concepts involved in object classification \citep{chormai2022disentangled}.

The modular features produced by PCA are simply the principal components. Given a model $\model$ and $k$ inputs, to featurize an $n$-dimensional hidden vector $\hiddenneurons$ we first create a $k \times n$ matrix with a column $\project{\sols(\model_{\inputval})}{\hiddenneurons}$ for each input $\inputval$. Then we use PCA to compute an $n \times n$ matrix $\mathbf{P}$ whose rows are principal components. The bijective translation $\setmap: \allvars \to \allvars$ is defined as 
\[\setmap(\tset) = \project{\tset}{\allvars \setminus \hiddenneurons} \cup \mathbf{P}^{T} \project{\tset}{\hiddenneurons}.\]

\paragraph{Sparse Autoencoders}

Sparse autoencoders are tools for learning to translate an $n$-dimensional hidden vector $\hiddenneurons$ into a sparse, $k$-dimensional feature space with $k \gg n$ \citep{bricken2023monosemanticity,Cunningham:2023,huben2024sparse, marks2024sparse}. However, instead of learning a single invertible function that translates activations into a new feature space, sparse autoencoders separately learn an encoder $f_{\text{enc}}$ and decoder $f_{\text{dec}}$, each typically parameterized by a single layer feed-forward network. These two functions are optimized to reconstruct the activations $\values{\hiddenneurons}$ while creating a sparse feature space, with a hyperparameter $\lambda$ balancing the two terms: 
\begin{align*}
\ell = \sum_{\inputval \in \inputvars} \Bigg( \left\Vert f_{\text{dec}}\bigg (f_{\text{enc}}(\project{\sols(\model_{\inputval})}{\hiddenneurons})\bigg) - \project{\sols(\model_{\inputval})}{\hiddenneurons}\right\Vert_{2} \\
+ \lambda \Vert f_{\text{enc}}(\project{\sols(\model_{\inputval})}{\hiddenneurons}) \Vert_{1} \Bigg)\\
\end{align*}
Given a sparse autoencoder that perfectly reconstructs the activations, i.e., one with a reconstruction loss of zero, we can view the encoder and decoder as the bijective translation
\[\setmap(\tset) = \project{\tset}{\allvars \setminus \hiddenneurons} \cup f_{\text{enc}}(\project{\tset}{\hiddenneurons});\;\;\setmap^{-1}(\tset) = \project{\tset}{\allvars \setminus \hiddenneurons} \cup f_{\text{dec}}(\project{\tset}{\hiddenneurons})\]
However, in practice, reconstruction loss is never zero and sparse autoencoders are approximate transformations of the underlying model.

\subsubsection{Aligning Low-Level Features with High-Level Variables}
Once a space of features has been learned, there is still the task of aligning features with high-level causal variables. Supervised modular feature learning techniques learn a feature space with an explicit alignment to high-level causal variables already in mind. However, in unsupervised modular feature learning an additional method is needed to align features with high-level causal variables.

\paragraph{Sparse Feature Selection}
A simple baseline method for aligning features with a high-level variable is to train a linear probe with a regularization term to select the features most correlated with the high-level variable \citep{huang2024ravel}.

\paragraph{Differential Binary Mask}

Differential binary masking selects a subset of features for a high-level variable by optimizing a binary mask with a training objective defined using interventions \citep{Cao2020, Csordas2021, de2021sparse, Davies:2023, prakash2024finetuning, huang2024ravel}. Differential binary masking has often been used to select individual neurons that play a particular causal role, but can just as easily be used to select features as long as the bijective translation is differentiable.

\subsubsection{Supervised Methods }

\paragraph{Probes}
Probing is the technique of using a supervised or unsupervised model to determine whether a concept is present in a hidden vector of a separate model. Probes are a popular tool for analyzing deep learning models, especially pretrained language models \citep{Hupkes:2018,conneau-etal-2018-cram, peters-etal-2018-dissecting,tenney-etal-2019-bert,clark-etal-2019-bert}.

Although probes are quite simple, they raise subtle methodological issues and our theoretical understanding of probes has greatly improved since their recent introduction into the field. From an information-theoretic point of view, we can observe that using arbitrarily powerful probes is equivalent to measuring the mutual information between the concept and the hidden vector \citep{hewitt-liang-2019-designing, Pimentel:2020}. If we restrict the class of probing models based on their complexity, we can measure how usable the information is \citep{Xu:2020, Hewitt:2021}.

Regardless of what probe models are used, successfully probing a hidden vector does not guarantee that it plays a causal role in model behavior \citep{ravichander2020probing,Elazar-etal-2020, geiger-etal:2020:blackbox,geiger2021causal}. However, a linear probe with weights $\weights$ trained to predict the value of a concept from an hidden vector $\hiddenneurons$ can be understood as learning a feature space for activations that can be analyzed with causal abstraction. Let $r_1, \dots, r_k$ be a set of orthonormal vectors that span the rowspace of $\weights$ and $u_{k+1}, \dots, u_{n}$ be a set of orthonormal vectors that span the nullspace of $\weights$. The bijective transformation is
\[\setmap(\tset) = \project{\tset}{\allvars \setminus \hiddenneurons} \cup [r_1 \dots r_k u_{k+1} \dots u_{n}]\project{\tset}{\hiddenneurons}\]
Since the probe is trained to capture information related to the concept $C$, the rowspace of the projection matrix $W$ (i.e., the first $k$ dimensions of the new feature space) might localize the concept $C$. If we have a high-level model that has a variable for the concept $C$, that variable should be aligned with the first $k$ features.

\paragraph{Distributed Alignment Search}\label{sec:DAS}
Distributed Alignment Search (DAS) finds linear subspaces of an $n$-dimensional hidden vector $\hiddenneurons$ in model $\lowmodel$ that align with high-level variables $\onevar_1, \dots, \onevar_k$ in model $\highmodel$ by optimizing an orthogonal matrix $\mathbf{Q} \in \mathbb{R}^{n \times n}$ with a loss objective defined using distributed interchange interventions (Section~\ref{sec:interchange}). This method has been used to analyze causal mechanisms in a variety of deep learning models \citep{Geiger-etal:2023:DAS,Wu:Geiger:2023:BDAS, tigges2023linear, arora2024causalgym, huang2024ravel, minder2024}. The bijective translation $\setmap: \allvars \to \allvars$ is defined
\[\setmap(\tset) = \project{\tset}{\allvars \setminus \hiddenneurons} \cup \mathbf{Q}^{T} \project{\tset}{\hiddenneurons}.\]

DAS optimizes $\mathbf{Q}$ so that $\varrs_1, \dots, \varrs_k$, disjoint subspaces of $\hiddenneurons$, are abstracted by high-level variables $\onevar_1, \dots, \onevar_k$ using the loss
\[
\ell = \sum_{\base, \source_1, \dots, \source_k \in \inputvars_{\lowmodel}}\mathsf{CE}(\lowmodel_{\base \cup \dintinv(\lowmodel, \tau, \langle \source_1,\dots, \source_k\rangle\langle\varrs_1, \dots, \varrs_k \rangle)}, \highmodel_{\intmap(\base) \cup \intinv(\highmodel, \langle \intmap(\source_1),\dots, \intmap(\source_k)\rangle\langle\onevar_1, \dots, \onevar_k \rangle)}),
\]
where $\intmap$ maps low-level inputs to high-level inputs and $\mathsf{CE}$ is cross-entropy loss. 

\subsection{Activation Steering as Causal Abstraction} \label{sec:activationsteering}
Causal explanation and manipulation are intrinsically linked \citep{Woodward2003}; if we understand which components in a deep learning model store high-level causal variables, we will be able to control the behavior of the deep learning model via intervention on those components.
Controlling model behavior via interventions was initially studied on recursive neural networks and generative adversarial networks \citep{Giulianelli:2018, Bau19a,Soulos:2020,Besserve20}. Recently, various works have focused on steering large language model generation through interventions. For instance, researchers have demonstrated that adding fixed steering vectors to the residual stream of transformer models can control the model's generation without training \citep{subramani2022extracting,turner2023activation,zou2023representation,vogel2024repeng,liInferenceTimeInterventionEliciting2024, wu2024advancing}. Additionally, parameter-efficient fine-tuning methods such as Adapter-tuning \citep{houlsbyParameterEfficientTransferLearning2019} can be viewed as interventions on model parameters. Between these two types of methods is \textit{representation fine-tuning} \citep{REFT}, where low-rank adapters are attached to a small number of hidden vectors in order to steer model behavior. 

While a successful interchange intervention analysis implies an ability to control the low-level model, the reverse does not hold. 
Activation steering has the power to bring the hidden vectors of a network off the distribution induced by the input data, potentially targeting hidden vectors that have the same value for every possible model input. An interchange intervention on such a vector will never have any impact on the model behavior. 

However, activation steering can still be represented in the framework of causal abstraction. For hidden vectors that are useful knobs to steer model generations in specific direction, we can simply define the map $\intmap$ from low-level interventionals to high-level interventionals on steering interventions rather than interchange interventions. The crucial point is that causal abstraction analysis that doesn't define $\intmap$ on interchange interventions will fail to uncover how the network reasons. It may nonetheless uncover how to control the network's reasoning process.

\subsection{Training AI Models to be Interpretable}\label{sec:trainingmodels}
Our treatment of interpretability has been largely through a scientific lens; an AI model being uninterpretable simply makes it an interesting object of study. However, when cracking open the black box is understood as normative goal that could have real positive societal impacts, a natural question is whether we can make our job any easier by creating models that are inherently more interpretable. This is an active area of research.

General purpose approaches attempt to design training procedure that produces more interpretable representations, such as using a SoLU function as a non-linearity \citep{elhage2022solu}, collapsing real-valued vector space into a discrete-valued space \citep{tamkin2024codebook}, learning a contextually weighted combination of vectors for different word meanings \citep{Hewitt2023}, or replacing MLPs with a new kind of network \citep{KAN}. More targeted approaches will construct architectural bottle-necks \citep{bottleneck1, bottleneck2, bottleneck3} or use interchange intervention based loss terms \citep{geiger-etal-2021-iit,wu-etal-2021-distill,wu-etal-2022-cpm, huang-etal-2023-inducing, zur-etal-2024-updating} in order to force a concept to be mediated by a particular vector or feature. While these are active avenues of exploration, there are, to date, no state-of-the-art models that have architectures or losses designed around interpretability.

\section{Conclusion}
We submit that causal abstraction provides a theoretical foundation for mechanistic interpretability that clarifies core concepts and lays useful groundwork for future development of methods that investigate algorithmic hypotheses about the internal reasoning of AI models.

\section*{Acknowledgements}
Thank you to the reviewers, Nora Belrose, and Frederik Hytting J\o rgensen for their feedback on earlier drafts of this paper. In particular, we would like to thank Sander Beckers for deep and thoughtful engagement as a reviewer, which improved the quality of this work over the course of the review process. This research is supported by a grant from Open Philanthropy.

\bibliography{refs}

\appendix

\end{document}